\documentclass{article}
\usepackage{pgfplots}
\PassOptionsToPackage{numbers,sort&compress}{natbib}

\usepackage[preprint]{neurips_2026}
\usepackage{makecell}
\usepackage[utf8]{inputenc}
\usepackage[T1]{fontenc}
\usepackage{hyperref}
\usepackage{url}
\usepackage{booktabs}
\usepackage{amsfonts}
\usepackage{nicefrac}
\usepackage{microtype}
\usepackage{enumitem}
\usepackage{xcolor}
\usepackage{amssymb}
\usepackage{graphicx}
\usepackage{twemojis}

\newcommand{\affmarkemoji}[1]{%
  \raisebox{0.50ex}{\twemoji[height=1.45ex]{#1}}%
}

\newcommand{\affinlineemoji}[1]{%
  \raisebox{-0.25ex}{\twemoji[height=1.9ex]{#1}}\,%
}

\newcommand{\instpoliba}{\affmarkemoji{oyster}}
\newcommand{\instsapienza}{\affmarkemoji{bank}}
\newcommand{\instedinburgh}{\affmarkemoji{unicorn}}
\newcommand{\instklagenfurt}{\affmarkemoji{dragon}}
\newcommand{\instminimlai}{\affmarkemoji{castle}}

\usepackage{tikz}
\usetikzlibrary{patterns}
\usepackage{multirow}
\usepackage{amsmath}
\usepackage{pgfplots}
\usepgfplotslibrary{groupplots}
\pgfplotsset{compat=1.18}
\usepackage[capitalize,noabbrev]{cleveref}
\usepackage{xspace}
\usepackage{amsthm}

\newcommand{\ie}{i.e.,\xspace}

\newcommand{\invalidtext}{\textsc{Invalid text}\xspace}
\newcommand{\invalidref}{\textsc{Invalid reference image}\xspace}
\newcommand{\invalidtarget}{\textsc{Invalid target image}\xspace}
\newcommand{\broadquery}{\textsc{Overly broad query}\xspace}
\newcommand{\valid}{\textsc{Valid}\xspace}
\newcommand{\circusname}{\textsc{CIRCUS}}
\newcommand{\circus}{\circusname\xspace}
\newcommand{\circussf}{\circusname$_{\textnormal{SF}}$\xspace}
\newcommand{\circusv}{\circusname$_{\textnormal{V}}$\xspace}
\newcommand{\circusexpanded}{Composed Image Retrieval Cleaned of Unimodal Shortcuts\xspace}


\title{
Do Composed Image Retrieval Benchmarks \\ Require Multimodal Composition?}

\author{
  \textbf{Matteo Attimonelli\instpoliba\instsapienza \qquad
  Alessandro De Bellis\instpoliba \qquad
  Aryo Pradipta Gema\instedinburgh \qquad
  Rohit Saxena\instedinburgh} \\
  \textbf{Monica Sekoyan\instedinburgh \qquad
  Wai-Chung Kwan\instedinburgh \qquad
  Claudio Pomo\instpoliba \qquad
  Alessandro Suglia\instedinburgh} \\
  \textbf{Dietmar Jannach\instklagenfurt \qquad
  Tommaso Di Noia\instpoliba \qquad
  Pasquale Minervini\instedinburgh\instminimlai} \\[0.7em]
  \affinlineemoji{oyster}Politecnico di Bari \quad
  \affinlineemoji{bank}Sapienza University of Rome \quad
  \affinlineemoji{unicorn}University of Edinburgh \\
  \affinlineemoji{dragon}University of Klagenfurt \quad
  \affinlineemoji{castle}Miniml.AI \\
  \texttt{matteo.attimonelli@poliba.it}
}

\begin{document}
\maketitle

\begin{abstract}
Composed Image Retrieval (CIR) is a multimodal retrieval task where a query consists of a reference image and a textual modification, and the goal is to retrieve a target image satisfying both. In principle, strong performance on CIR benchmarks is assumed to require \emph{multimodal composition}, i.e., combining complementary information from reference image and textual modification.
In this work, we show that this assumption does not always hold. Across four widely used CIR benchmarks and eleven Generalist Multimodal Embedding models, a large fraction of queries can be solved using a single modality (from 32.2\% to 83.6\%), revealing pervasive unimodal shortcuts. Thus, high CIR performance can arise from unimodal signals rather than true multimodal composition.
To better understand this issue, we perform a two-stage audit. First, we identify shortcut-solvable queries through cross-model analysis. Second, we conduct human validation on 4{,}741 shortcut-free queries, of which only 1{,}689 are well-formed, with common issues including ambiguous edits and mismatched targets.
Re-evaluating models on this validated subset reveals qualitatively different behaviour: queries can no longer be solved with a single modality, and successful retrieval requires combining both inputs. While accuracy decreases, reliance on multimodal information increases.
Overall, current CIR benchmarks conflate shortcut-solvable, noisy, and genuinely compositional queries, leading to an overestimation of model capability in multimodal composition. \\


\raisebox{-0.25\height}{\hspace{0.05cm}\includegraphics[width=0.42cm]{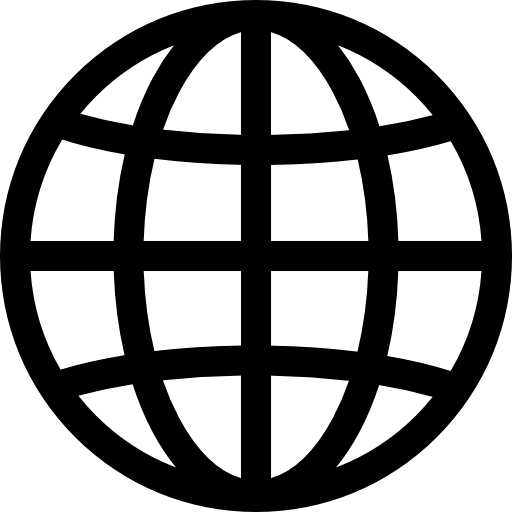}} \small \textbf{\mbox{Website:}} \href{https://matteoattimonelli.github.io/CIRCUS/}{https://matteoattimonelli.github.io/CIRCUS/}.

\end{abstract}

\section{Introduction}
Composed Image Retrieval (CIR) is a multimodal retrieval task in which a query consists of a reference image and a textual modification, and the goal is to retrieve a target image satisfying both.
This formulation is intended to test \emph{multimodal composition}, i.e., the ability to combine complementary information from the reference image and the textual modification into a representation that cannot be recovered from either modality alone.
Benchmarks such as CIRR~\citep{DBLP:conf/iccv/LiuROG21}, FashionIQ~\citep{DBLP:conf/cvpr/WuGGARGF21}, LaSCo~\citep{DBLP:conf/aaai/LevyBDL24}, and CIRCO~\citep{Baldrati_2023_ICCV} are commonly used to evaluate this ability.

A key challenge in multimodal learning is ensuring that models genuinely integrate information across modalities, rather than relying on a single dominant signal.
Prior work has shown that models often exploit shortcut signals: in visual question answering, many questions can be answered from text alone~\citep{DBLP:conf/cvpr/GoyalKSBP17}, and similar effects have been observed in multimodal and language benchmarks~\citep{DBLP:journals/natmi/GeirhosJMZBBW20}---showcasing weak visual understanding~\citep{asadi2026mirage}.
This raises the question of whether similar behaviour arises in CIR.
As a compositional multimodal retrieval task, CIR is expected to require both modalities. However, if a query can be solved using only text or image information, retrieval does not reflect true multimodal composition, but instead indicates reliance on \emph{unimodal shortcuts}~\citep{DBLP:conf/aaai/LevyBDL24}.
In such cases, benchmark performance may overestimate a model’s ability to combine modalities.

In this work, we systematically evaluate whether CIR benchmarks enforce multimodal composition.
For each query, we compare retrieval under three conditions: (i) the full multimodal input, (ii) text-only, where the reference image is suppressed, and (iii) image-only, where the textual edit is removed.
We perform this analysis using a heterogeneous pool of eleven \emph{Generalist Multimodal Embedding models}~\citep[GMEs;][]{DBLP:journals/corr/abs-2412-16855,DBLP:conf/iclr/LinLSLCP25,DBLP:conf/cvpr/LiuZCJ0YWX25,DBLP:journals/corr/abs-2601-03666,DBLP:journals/corr/abs-2601-04720}, which encode multimodal queries into a shared representation space.
As these models are trained on large-scale multimodal corpora and support multiple retrieval settings, they provide a suitable probe for identifying dataset-level shortcuts.

\begin{figure*}[t]
\centering
\includegraphics[width=0.86\textwidth]{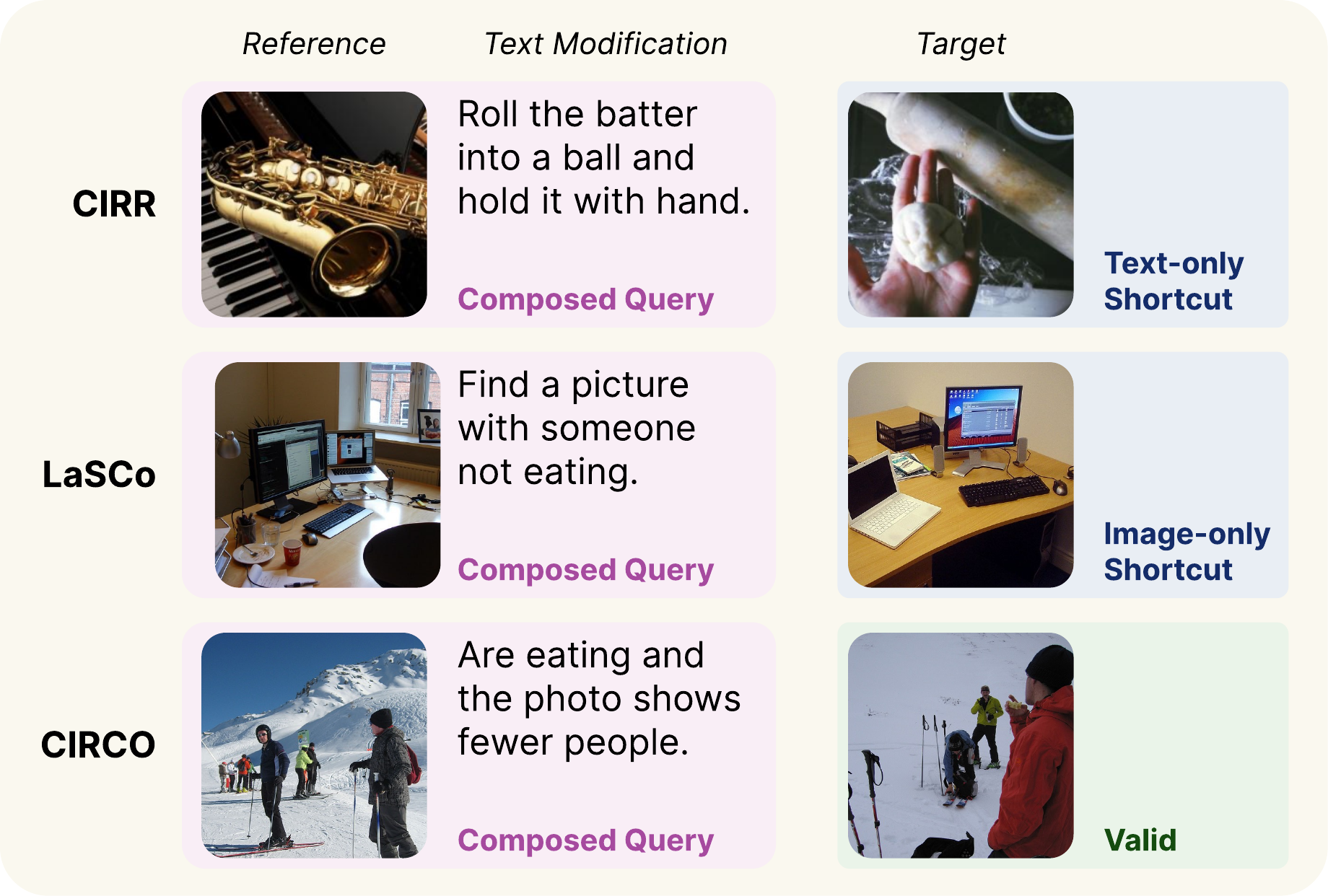}
\caption{Examples of text-only shortcuts (top), image-only shortcuts (middle), and valid queries (bottom) from three CIR benchmarks. These cases show that high retrieval accuracy does not necessarily imply multimodal composition.}
\label{fig:intro-shortcut-examples}
\end{figure*}

A query is said to admit a unimodal shortcut if any model can retrieve the target using a single modality.
\cref{fig:intro-shortcut-examples} illustrates this phenomenon---in text-only shortcuts, the textual modification alone is sufficient to identify the target; in image-only shortcuts, the reference image alone suffices, often due to uninformative edits.
This highlights that strong benchmark performance does not necessarily imply effective multimodal composition.
Our analysis reveals that unimodal shortcuts are pervasive across standard CIR benchmarks: 83.6\% of CIRR, 32.2\% of FashionIQ, 38.7\% of LaSCo, and 74.5\% of CIRCO queries are solvable using a single modality.
This shows that a large portion of benchmark performance does not require multimodal composition and can therefore be explained by unimodal shortcuts, undermining the intended role of CIR as a test of compositional reasoning.

Removing shortcut-solvable queries is necessary but not sufficient---among the remaining \emph{shortcut-free} queries, many are ambiguous, underspecified, or incorrectly annotated.
To quantify this effect, we conduct a human validation study on 4{,}741 such queries (\cref{sec:human-val}) and find that only 1{,}689 (35.6\%) form well-defined CIR instances (e.g., \cref{fig:intro-shortcut-examples}, bottom).
The dominant issue is overly broad queries, where multiple images satisfy the modification, making the target non-unique.
Therefore, we present \circus{} (\circusexpanded), a cleaned evaluation suite of CIR benchmarks, which provides a more reliable test of multimodal composition.

Re-evaluating models on \circus reveals a different behaviour: queries can no longer be solved using a single modality, and successful retrieval requires combining both inputs.
Performance drops sharply---for instance, Qwen3-VL-8B's Recall on CIRR falls from 70.1 to 24.1, indicating that much of the apparent benchmark performance came from unimodal shortcuts rather than genuine multimodal composition (\cref{subsec:shortcut_audit}).
\circus therefore provides a more faithful assessment of how well models compose image and text.

To summarise, our contributions are the following:
\begin{enumerate}[leftmargin=*]
    \item We quantify the prevalence of unimodal shortcuts in CIR benchmarks (\Cref{subsec:shortcut_audit}), showing that a large fraction of queries can be solved using only text or only image information.
    \item We show, through manual validation (\Cref{subsec:validity_audit}), that the four audited CIR benchmarks suffer from additional issues beyond shortcuts, including overly broad modifications and invalid or non-unique ground-truth targets, and construct \circus{}, retaining the well-formed CIR queries.
    \item We demonstrate that \circus{} provides a more diagnostic test of multimodal composition (\Cref{subsec:re-eval}): queries can only be solved by combining both modalities, with balanced reliance on each, whereas original benchmarks overestimate performance due to unimodal shortcuts.
\end{enumerate}
Overall, our results show that current CIR benchmarks conflate shortcut-solvable, noisy, and genuinely compositional queries, leading to an overestimation of model capabilities.
Consequently, reliance on these benchmarks may overestimate models’ multimodal capabilities, as performance can be driven by unimodal shortcuts rather than multimodal composition.

\section{Auditing CIR Benchmarks}
\subsection{Preliminaries}
\label{sec:preliminaries}

In CIR, a query $Q_i$ consists of a reference image $R_i$ and a textual modification $T_i$ describing a desired change. The goal is to retrieve, from a gallery $\mathcal{G}$, a target image that best matches the intent of $T_i$ applied to $R_i$. A model $p$ produces a ranking over the gallery for each query using a similarity function. We denote by $r_p(i)$ the rank of the ground-truth target for query $Q_i$, and say that the query is \emph{solved} at cutoff $K$ if $r_p(i) \leq K$.
\subsection{Unimodal Shortcut Audit} \label{sec:unimodal-ablation}

We ask whether CIR queries genuinely require combining both modalities, or whether a single modality is sufficient to retrieve the target, signifying the presence of a unimodal shortcut. To answer this, we evaluate each query under three setups for every evaluated GME, hereafter referred to interchangeably as a \emph{retriever}:

\begin{itemize}[leftmargin=*]
    \item \textbf{Multimodal}: the full query, combining the reference image and the editing text.
    \item \textbf{Text-only}: the editing text only; the reference image is replaced with a solid black image of the same dimensions, used as a masking placeholder for retrievers that require an image input~\citep{DBLP:journals/corr/abs-2505-18657}.
    \item \textbf{Image-only}: the reference image only; the editing text is removed.
\end{itemize}

Let $r_p^{\text{mm}}(i)$, $r_p^{\text{txt}}(i)$, and $r_p^{\text{img}}(i)$ denote the rank of the ground-truth target for query $i$ under multimodal, text-only, and image-only conditions for retriever $p$, respectively.
Since unimodal shortcut solvability may depend on the specific retriever, we aggregate across models to obtain a dataset-level view. In particular, we define the best achievable unimodal ranks:
\begin{equation} \label{eq:unimodal}
    r^{\text{txt}}_*(i) = \min_{p} r_p^{\text{txt}}(i), \qquad
    r^{\text{img}}_*(i) = \min_{p} r_p^{\text{img}}(i).
\end{equation}
This corresponds to asking the question: \emph{can any tested GME retrieve the target using a single modality?}
If so, the query admits a unimodal shortcut, regardless of which retriever exploits it. We then assign each query an aggregate label:
\begin{itemize}[leftmargin=*]
\item \textbf{Shortcut:} $r^{\text{txt}}_*(i) \leq K$ or $r^{\text{img}}_*(i) \leq K$.
\item \textbf{Composition-required:} $r^{\text{txt}}_*(i) > K$ and $r^{\text{img}}_*(i) > K$, and at least one retriever has $r_p^{\text{mm}}(i) \leq K$.
\item \textbf{Unresolved:} $r^{\text{txt}}_*(i) > K$ and $r^{\text{img}}_*(i) > K$, and no retriever achieves $r_p^{\text{mm}}(i) \leq K$.
\end{itemize}
The union of \emph{composition-required} and \emph{unresolved} defines the \textbf{\circussf} (\emph{shortcut-free}) set: queries for which no tested unimodal configuration succeeds. This removes shortcut-solvable cases by construction. However, it does not guarantee that the remaining queries are valid CIR instances. In particular, unresolved queries may be unsolved either because they are genuinely difficult or because they are ambiguous, ill-posed, or incorrectly annotated (e.g., nonsensical edits or mismatched targets). To disentangle genuine compositional difficulty from dataset noise, we therefore perform a second stage of analysis based on human validation.

\subsection{Human Validation Protocol}
\label{sec:human-val}

To assess the quality of the shortcut-free set, we perform a human validation study on a subset of its queries
following a shared protocol
(see Appendix~\ref{app:human-validation-materials} for more details on the annotation process).
For each instance, annotators are presented with the reference image, the textual edit, and the target image. In addition, they are shown an \emph{aggregate multimodal panel}, defined as the deduplicated union of the top-$K$ retrieval results across all retrievers under the multimodal condition.
This panel is provided as an auxiliary signal during annotation.
Given this information, annotators are asked to determine whether the triplet—reference image, textual edit, and target image—forms a valid CIR example. An instance is marked as \valid if it is coherent, specific, and visually verifiable. Otherwise, annotators identify one or more issues according to the following taxonomy:
\begin{itemize}[leftmargin=*]
    \item \textbf{\invalidtext}: the textual edit is garbled, incoherent, or unrelated to the reference image.
    \item \textbf{\invalidref}: the reference image is too cropped, blurry, or low-quality to support the query.
    \item \textbf{\invalidtarget}: the target is degraded or does not match the requested modification.
    \item \textbf{\broadquery}: the triplet is plausible, but the composed query admits many non-ground-truth images as valid matches, making the instance unsuitable as a precise retrieval benchmark.
\end{itemize}
The aggregate multimodal panel is used to support the identification of \broadquery cases.
It provides annotators with a set of candidate images that can be inspected to determine whether the composed query admits multiple plausible matches.
A query is labelled as \broadquery when at least $K$ distinct images plausibly satisfy the same composed query, indicating that the target is not uniquely identifiable.

Annotators are instructed to evaluate the \emph{data instance} and are not informed whether a query belongs to the composition-required or unresolved categories to avoid bias.
While multiple issue labels can be assigned, annotators follow a fixed decision order—text validity, reference image quality, target correctness, and finally query specificity—to ensure consistency across annotations.
All instances marked as \valid define the \textbf{\circusv} (\emph{validated}) evaluation set, which we use to assess whether CIR truly requires multimodal composition.
Audit scope and statistics are reported in~\cref{subsec:validity_audit}.
Inter-annotator agreement statistics are provided in Appendix~\ref{app:agreement}.

\section{Experimental Setup}

\paragraph{Models}
\label{sec:taxonomy}
We evaluate a diverse set of GMEs, which encode text, images, and mixed multimodal queries into a shared embedding space and perform retrieval via cosine similarity. This design enables a single model to support text-only, image-only, and composed retrieval without task-specific heads, and makes GMEs a natural probe for auditing benchmark properties: if a query is solvable using a single modality for any strong GME, this is more likely a property of the benchmark than of a specific model. Our retriever pool spans different model families, backbones, and training strategies. We include MM-Embed~\citep{DBLP:conf/iclr/LinLSLCP25}, GME-Qwen2VL~\citep{DBLP:journals/corr/abs-2412-16855}, LamRA~\citep{DBLP:conf/cvpr/LiuZCJ0YWX25}, E5-Omni~\citep{DBLP:journals/corr/abs-2601-03666}, VLM2Vec-V2~\citep{DBLP:journals/tmlr/MengJLSYFQTZCXXZCY26}, and Rzen-Embed~\citep{DBLP:journals/corr/abs-2510-27350}. We also include LamRA-Qwen2.5VL and the 2B/8B variants of Qwen3-VL-Embedding~\citep{li2026qwen3vlembeddingqwen3vlrerankerunifiedframework} to capture controlled architectural and scale variations, as well as two commercial systems, Gemini Embedding~2 and Voyage MM-3.5.
The goal is to assemble a diverse retriever pool so that shortcut detection reflects dataset properties rather than model-specific behaviour.
All experiments are run on a single NVIDIA H100 GPU.

\paragraph{Datasets}
We evaluate four composed image retrieval benchmarks that span diverse visual domains and query types, namely \textbf{CIRR}~\citep{DBLP:conf/iccv/LiuROG21}, \textbf{FashionIQ}~\citep{DBLP:conf/cvpr/WuGGARGF21}, \textbf{LaSCo}~\citep{DBLP:conf/aaai/LevyBDL24}, and \textbf{CIRCO}~\citep{Baldrati_2023_ICCV}. All benchmarks require the combination of a reference image and a textual modification to retrieve a target from a gallery. CIRR contains 4{,}170 test queries over natural images drawn from NLVR$^2$, where queries describe relative modifications to everyday scenes (e.g., ``Show three bottles of soft drink'') over a gallery of 21{,}552 images. FashionIQ focuses on fashion products, with 6{,}003 test queries specifying attribute or style changes (e.g., ``Is shiny and silver with shorter sleeves'') and a gallery of 18{,}853 images. LaSCo scales this setting to 30{,}031 queries on COCO images, with scene-level modifications (e.g., ``The boats are not docked'') over a gallery of 40{,}083 images. CIRCO differs in being open-domain and allowing multiple valid targets per query; since its test split is not publicly available, we evaluate on the validation set, which contains 220 queries and a gallery of 123{,}403 images.

\paragraph{Metrics}
We use Recall with a cutoff of 10 (Recall@10) as the primary retrieval metric and as the success criterion for the shortcut audit. A query is considered solved if at least one annotated relevant target appears within the top-10 retrieved results. Retrieval is performed by ranking gallery embeddings by cosine similarity to the query embedding. Throughout the paper, we follow each benchmark's native relevance definition: CIRR, FashionIQ, and LaSCo provide a single annotated target per query, whereas CIRCO provides multiple annotated positives. All recall and ranking metrics are therefore computed against the full annotated relevant set for the benchmark at hand, which keeps the same audit and ablation setup fair under both single-target and multi-positive labelling. For the shortcut audit, we compute Recall separately for multimodal, text-only, and image-only queries, and aggregate across retrievers using the best-rank criterion (\Cref{sec:unimodal-ablation}) to determine whether any GME can solve a query under each condition.
To analyse ranking behaviour beyond a fixed cutoff, we report full-catalogue nDCG, computed over the entire gallery ranking. Unlike Recall, which reflects success at a fixed threshold, nDCG captures ranking quality and is sensitive to improvements throughout the list. This makes it more suitable for comparing multimodal and unimodal retrieval and for quantifying modality reliance.
We further derive a normalised measure of multimodal contribution (Composition Gap) in~\cref{subsec:re-eval}, which accounts for differences in score scale across splits.

\section{Results}
\subsection{Shortcut Audit: How Many Queries Require Multimodal Composition?}
\label{subsec:shortcut_audit}

We begin by asking a simple question: \emph{how many CIR queries actually require combining image and text?}
\Cref{tab:aggregated} reports the result of our shortcut audit after aggregating across all retrievers. The main finding is clear: \emph{unimodal shortcuts are pervasive across all benchmarks}. A large fraction of queries can be solved (\ie the target is retrieved in the top-10) using only the text or only the image, without requiring multimodal composition.

\begin{table}[t]
\centering
\caption{Aggregated query categorisation across eleven retrievers at $K=10$.
Percentages are of the total query count for each dataset.}
\label{tab:aggregated}
\small
\begin{tabular}{l rrr rrr rrr rrr}
\toprule
 & \multicolumn{3}{c}{\textbf{CIRR}} & \multicolumn{3}{c}{\textbf{FashionIQ}}
 & \multicolumn{3}{c}{\textbf{LaSCo}} & \multicolumn{3}{c}{\textbf{CIRCO}} \\
 & \multicolumn{3}{c}{(4{,}170 queries)} & \multicolumn{3}{c}{(6{,}003 queries)}
 & \multicolumn{3}{c}{(30{,}031 queries)} & \multicolumn{3}{c}{(220 queries)} \\
\cmidrule(lr){2-4} \cmidrule(lr){5-7} \cmidrule(lr){8-10} \cmidrule(lr){11-13}
\textbf{Category} & $N$ & \multicolumn{1}{c}{\%} & & $N$ & \multicolumn{1}{c}{\%}
& & $N$ & \multicolumn{1}{c}{\%} & & $N$ & \multicolumn{1}{c}{\%} & \\
\midrule
\textbf{Shortcut}          & \textbf{3{,}485} & \textbf{83.6} && \textbf{1{,}934} & \textbf{32.2} && \textbf{11{,}613} & \textbf{38.7} && \textbf{164} & \textbf{74.5} \\
\quad Both conditions      &    871 & 20.9 &&     89 &  1.5 &&  2{,}084 &  6.9 &&     47 & 21.4 \\
\quad Text only            &  2{,}244 & 53.8 &&  1{,}552 & 25.9 &&  6{,}380 & 21.2 &&     54 & 24.5 \\
\quad Image only           &    370 &  8.9 &&    293 &  4.9 &&  3{,}149 & 10.5 &&     63 & 28.6 \\
\midrule
\textbf{Composition-required}  &    \textbf{271} &  \textbf{6.5} &&  \textbf{1{,}462} & \textbf{24.4} &&  \textbf{2{,}064} &  \textbf{6.9} &&     \textbf{53} & \textbf{24.1} \\
\textbf{Unresolved}            &    \textbf{414} &  \textbf{9.9} &&  \textbf{2{,}607} & \textbf{43.4} && \textbf{16{,}354} & \textbf{54.5} &&      \textbf{3} &  \textbf{1.4} \\
\bottomrule
\end{tabular}

\end{table}

This shortcut effect is particularly pronounced in CIRR, where \textbf{83.6\%} of queries admit a unimodal solution and only \textbf{6.5\%} require both modalities. FashionIQ is comparatively less affected, but still only \textbf{24.4\%} of queries are composition-required. LaSCo is dominated by unresolved cases (\textbf{54.5\%}), suggesting that much of its difficulty arises from general retrieval challenges rather than clean multimodal reasoning. CIRCO presents a different profile: although \textbf{74.5\%} of queries admit shortcuts, it retains a non-negligible composition-required subset (\textbf{24.1\%}).
These results show that \textbf{common CIR benchmarks do not reliably enforce multimodal composition}. In many cases, high retrieval performance can be achieved without combining modalities, undermining their intended role as tests of compositional reasoning.

Importantly, this phenomenon becomes apparent only after aggregating across retrievers. Individual models underestimate shortcut prevalence because different models exploit different shortcuts. For example, on CIRR, the strongest single retriever identifies 64.4\% of queries as unimodal-solvable, whereas the aggregated shortcut rate rises to 83.6\%. This confirms that shortcuts are a \emph{dataset-level property}, not a model-specific artifact.

\begin{table}[t]
\centering
\caption{Recall@10 (\%) per retriever under multimodal (MM), text-only (T), and image-only (I) conditions. Best per-dataset results in each condition are \textbf{bolded}; second-best results are underlined.}
\label{tab:per_provider}
\small
\resizebox{\textwidth}{!}{
\begin{tabular}{l ccc ccc ccc ccc}
\toprule
 & \multicolumn{3}{c}{\textbf{CIRR}} & \multicolumn{3}{c}{\textbf{FashionIQ}}
 & \multicolumn{3}{c}{\textbf{LaSCo}} & \multicolumn{3}{c}{\textbf{CIRCO}} \\
\cmidrule(lr){2-4} \cmidrule(lr){5-7} \cmidrule(lr){8-10} \cmidrule(lr){11-13}
\textbf{Retriever} & MM & T & I & MM & T & I & MM & T & I & MM & T & I \\
\midrule
E5-Omni             & 55.2 & 47.5 & 17.2 & 16.9 & 10.6 & 2.3 &  9.4 &  8.4 & 7.2 & 65.9 & 19.1 & 23.6 \\
GME-Qwen2VL         & 64.4 & \underline{51.9} & 16.2 & 31.1 & 13.5 & 2.3 & 13.2 & 11.6 & 7.3 & \underline{85.9} & \underline{25.9} & 23.6 \\
LamRA               & 65.1 & 49.7 & 16.6 & \textbf{33.2} & 10.7 & 1.9 & 12.3 &  9.1 & 6.9 & 81.4 & 17.3 & 24.1 \\
LamRA-Qwen2.5VL     & \underline{65.2} & 47.3 & 16.1 & 31.9 & 11.5 & 1.8 & 12.8 & 10.3 & 7.3 & 84.5 & 18.6 & 23.2 \\
MM-Embed            & 63.3 & 46.5 & 16.6 & 25.5 & 11.5 & \underline{2.6} & \textbf{18.2} &  9.9 & 7.2 & 82.7 & 23.2 & 24.6 \\
Qwen3-VL-2B         & 64.6 & 48.1 & 17.9 & 28.3 & 12.9 & 2.3 & 11.2 & 10.4 & 7.3 & 78.2 & 23.6 & 20.9 \\
Qwen3-VL-8B         & \textbf{70.1} & 46.6 & \textbf{18.2} & \underline{32.6} & \textbf{15.3} & 2.5 & 13.3 & \underline{12.1} & \textbf{7.7} & \textbf{86.4} & 20.9 & 25.9 \\
Rzen-Embed          & \textbf{70.1} & \textbf{56.7} & 16.6 & 30.8 & \underline{14.1} & 2.1 & 12.8 & \textbf{13.6} & 7.2 & \textbf{86.4} & \textbf{27.7} & 25.0 \\
VLM2Vec-V2          & 55.0 & 39.4 & 16.7 & 15.8 &  7.0 & 2.1 &  9.1 &  6.0 & \underline{7.5} & 38.6 & 10.9 & \underline{26.4} \\
Gemini Emb.\ 2      & 49.2 & 37.3 & \underline{18.1} & 24.9 & 13.8 & \textbf{2.9} & 12.4 &  9.3 & 7.0 & 74.1 & 23.6 & 23.6 \\
Voyage MM-3.5       & 54.4 & 45.5 & 15.8 & 19.4 &  9.7 & 2.2 & \underline{14.8} & 11.2 & 7.4 & 77.7 & 20.9 & \textbf{26.8} \\
\bottomrule
\end{tabular}
}
\end{table}

\Cref{tab:per_provider} provides a complementary per-retriever view and helps explain how these shortcuts arise in practice. Across models, unimodal retrieval is already highly competitive, especially for text. On CIRR, for example, the strongest text-only model (Rzen-Embed) reaches 56.7\% Recall@10, compared to 70.1\% in the multimodal setting, i.e., over 80\% of full performance.
Similar patterns hold across other strong models, indicating that much of the retrieval signal is already captured by the textual edit alone.
Notably, text-only retrieval is often much closer to multimodal performance than image-only retrieval, pointing to a strong reliance on textual cues. A non-trivial fraction of queries are also solvable under both text-only and image-only conditions, suggesting that the two modalities are redundant for those instances.
Consistently, image-only retrieval is the weakest on CIRR, FashionIQ, and LaSCo, reinforcing the dominance of textual signals. CIRCO is a notable exception, where image-only performance is substantially higher, in line with the more balanced shortcut distribution observed in \Cref{tab:aggregated}.

Overall, these trends are consistent across a diverse set of retrievers, including both open and closed models. While absolute performance varies, no model fundamentally breaks this pattern: even the strongest systems rely heavily on unimodal signals. This supports the view that shortcut exploitation is a structural property of current CIR benchmarks.
These findings are further reinforced by Appendix~\ref{app:uncertainty} and \ref{app:shortcut-robustness}, which show that the observed patterns are stable under sampling uncertainty, across retrieval cutoffs, and when recomputing the aggregate after removing each retriever in turn.
\emph{As a consequence, performance on these benchmarks can substantially overestimate a model’s ability to combine modalities, and reported gains may not always correspond to progress in CIR.}

\subsection{Validity Audit: How Many Shortcut-Free Queries Are Well-Formed?}
\label{subsec:validity_audit}

Removing unimodal shortcuts is necessary, but not sufficient to obtain a reliable evaluation set. The resulting \circussf{} queries (\emph{shortcut-free}) may still be ambiguous, underspecified, or incorrectly annotated. We therefore perform a human validation study to assess the quality of this residual set.

\begin{figure*}[t]
    \centering
    \includegraphics[width=\textwidth]{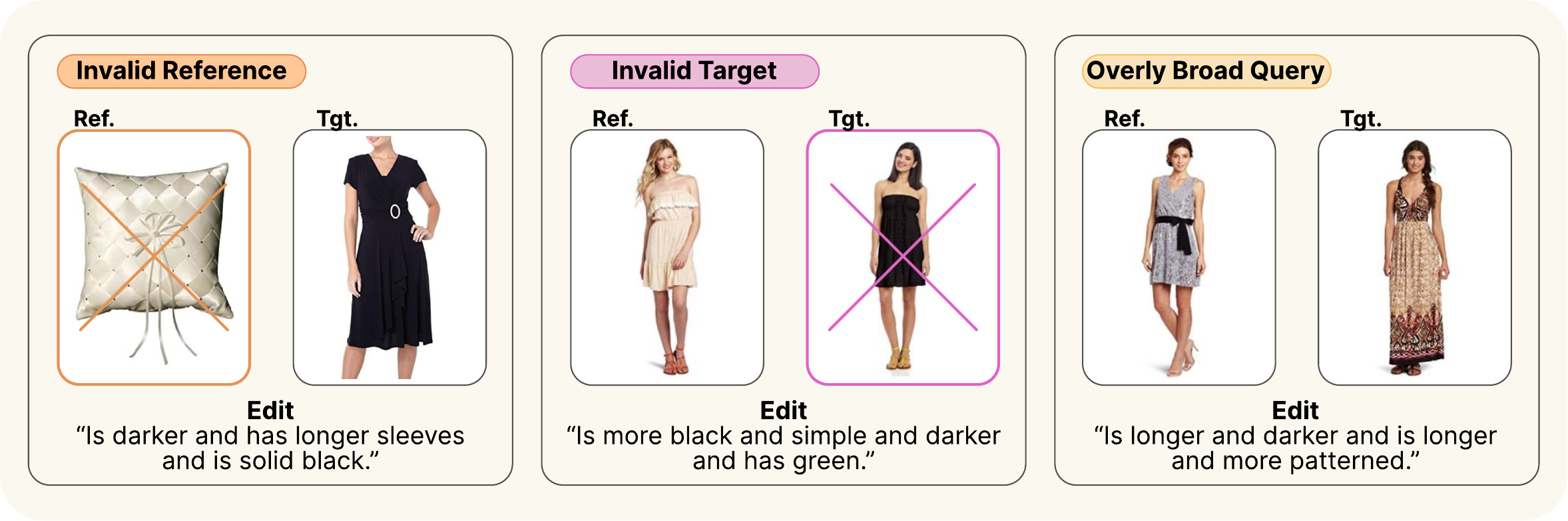}
    \caption{
    Representative failures from the audited \circussf{} set. Each panel shows the reference image, the textual edit, and the target. Examples illustrate three failure modes: unsupported references (\invalidref), mismatched targets (\invalidtarget), and underspecified queries with multiple plausible matches (\broadquery).
    }
    \label{fig:validity_issue_examples}
\end{figure*}

\Cref{tab:validity_audit} reports the results. For CIRR and CIRCO, we audit the full shortcut-free residue. For FashionIQ and LaSCo, where the shortcut-free residue is much larger, we audit stratified samples of 1{,}000 composition-required and 1{,}000 unresolved queries per dataset. In total, 4{,}741 instances are annotated.
Across all audited shortcut-free queries, only 1{,}689 (\textbf{35.6\%}) are marked as \valid. The composition-required subset is consistently cleaner than the unresolved subset, but neither is close to noise-free. For example, on CIRR, 54.2\% of composition-required queries are valid, compared to 37.7\% of unresolved ones; the same pattern appears on FashionIQ (36.8\% vs.\ 21.8\%) and LaSCo (45.2\% vs.\ 30.6\%). Thus, even queries that appear to require both modalities under our shortcut audit are often not well-formed CIR instances.
Appendix~\ref{app:uncertainty} provides interval estimates strengthening results under the stratified sampling design.
Representative audited failures are shown in \Cref{fig:validity_issue_examples}.

\begin{table*}[t]
\centering
\caption{Human validation of the shortcut-free residue.
For CIRR and CIRCO the full shortcut-free set was audited; For FashionIQ and LaSCo stratified samples of 1{,}000 composition-required and 1{,}000 unresolved queries were audited.
}
\label{tab:validity_audit}
\small
\setlength{\tabcolsep}{5pt}
\resizebox{\textwidth}{!}{
\begin{tabular}{l l rrr rrr r}
\toprule
& & \multicolumn{3}{c}{\textbf{Composition-required}} & \multicolumn{3}{c}{\textbf{Unresolved}} & \\

\cmidrule(lr){3-5} \cmidrule(lr){6-8}
\textbf{Dataset} & \textbf{Audit Scope} & \textbf{Audited} & \textbf{Valid} & \textbf{Valid \%} & \textbf{Audited} & \textbf{Valid} & \textbf{Valid \%} & \textbf{Total Valid} \\
\midrule
CIRR       & full shortcut-free  & 271  & 147 & 54.2 & 414 & 156 & 37.7 & 303 \\
FashionIQ  & stratified sample      & 1000 & 368 & 36.8 & 1000 & 218 & 21.8 & 586 \\
LaSCo      & stratified sample      & 1000 & 452 & 45.2 & 1000 & 306 & 30.6 & 758 \\
CIRCO      & full shortcut-free  & 53   & 39  & 73.6 & 3    & 3   & 100.0 & 42 \\
\midrule
\textbf{Total} & -- & \textbf{2{,}324} & \textbf{1{,}006} & \textbf{43.3} & \textbf{2{,}417} & \textbf{683} & \textbf{28.3} & \textbf{1{,}689} \\
\bottomrule
\end{tabular}}
\end{table*}

The dominant failure mode is \broadquery, accounting for the majority of invalid cases.
These instances are not necessarily incorrect: the reference image, edit, and target may all be plausible, but the composed query admits many non-ground-truth matches.
As a result, the annotation is effectively non-unique, and a retriever may return a semantically valid alternative and still be penalized.
This indicates that much of the residual difficulty arises from under-specification rather than compositional reasoning.
This pattern is consistent across datasets (Appendix~\ref{app:audit-error-distribution}), where \broadquery accounts for 60--85\% of invalid instances.

In contrast, \invalidtarget is more frequent among unresolved queries, reaching up to $\sim$30\% of invalid cases, suggesting that some unresolved examples fail due to target mismatches rather than model limitations. By comparison, \invalidtext and \invalidref are rare, indicating that low-quality inputs are not a major source of noise.

\emph{These results highlight that shortcut filtering addresses only part of the problem. While it removes queries solvable by a single modality, the remaining set still contains
ill-posed instances. As a result, the shortcut-free set cannot be used directly as a clean test of multimodal composition.}
We denote \circusv{} (\emph{validated}) as the subset of the shortcut-free subset whose instances are marked \valid. For CIRR and CIRCO, \circusv{} covers the full shortcut-free residue; for FashionIQ and LaSCo, \circusv{} is the validated subset of the audited shortcut-free sample.
\circusv{} provides a cleaner diagnostic split for evaluating whether retrieval models genuinely combine image and text, rather than relying on shortcuts or ambiguous instances.

\subsection{Re-evaluation of Retrievers on Filtered Subsets}\label{subsec:re-eval}

We re-evaluate each retriever on the original benchmark and on the two \circus{} views—\circussf{} and \circusv{}—using composed queries.
This analysis isolates two effects: how much reported performance relies on shortcut-solvable queries, and whether the remaining benchmark more faithfully requires multimodal composition.

\paragraph{Performance drops on filtered subsets}

\Cref{tab:subset_recall} reports multimodal Recall@10 on the original benchmark and on the two \circus{} views. Removing shortcuts leads to a substantial drop in performance. On CIRR, for example, Qwen3-VL-8B falls from 70.1 on the full benchmark to 17.2 on \circussf{}, before recovering to 24.1 on \circusv{}. The same pattern holds across FashionIQ and LaSCo: performance on the raw benchmark overestimates how well current GMEs handle queries that truly require multimodal composition, while human validation removes invalid and overly broad instances from the \circussf{} residue, yielding a cleaner and more interpretable evaluation.
Importantly, this drop should not be interpreted as a degradation of model capability, but as the removal of shortcut-solvable and ill-posed queries that confound the evaluation.
CIRCO is the main exception, as its \circussf{} residue is already small and relatively clean, so shortcut-free and validated scores remain close.
These results are supported by bootstrap intervals in Appendix~\ref{app:uncertainty}, indicating that the observed differences are not driven by a small number of borderline queries.

\begin{table*}[t]
\centering
\caption{Recall@10 (\%) on the original benchmark (\textbf{Full}), \circussf (SF), and the \circusv subset (V). Image-only and text-only results are omitted, as subsets contain only queries unsolved by any unimodal configuration within top-10. Best results are \textbf{bolded}, second-best are underlined.}
\label{tab:subset_recall}
\scriptsize
\setlength{\tabcolsep}{4pt}
\resizebox{\textwidth}{!}{
\begin{tabular}{l ccc ccc ccc ccc}
\toprule
\textbf{Provider} & \multicolumn{3}{c}{\textbf{CIRR}} & \multicolumn{3}{c}{\textbf{FIQ}} & \multicolumn{3}{c}{\textbf{LaSCo}} & \multicolumn{3}{c}{\textbf{CIRCO}} \\
\cmidrule(lr){2-4} \cmidrule(lr){5-7} \cmidrule(lr){8-10} \cmidrule(lr){11-13}
 & \textbf{Full} & \textbf{SF} & \textbf{V} & \textbf{Full} & \textbf{SF} & \textbf{V} & \textbf{Full} & \textbf{SF} & \textbf{V} & \textbf{Full} & \textbf{SF} & \textbf{V} \\
\midrule
E5-Omni & 55.2 & 11.8 & 15.2 & 16.9 & 6.9 & 12.5 & 9.4 & 0.5 & 2.3 & 65.9 & 39.3 & 38.1 \\
GME-Qwen2VL & 64.4 & 16.1 & 20.4 & 31.1 & 14.7 & 23.8 & 13.2 & 1.9 & 11.2 & \underline{85.9} & \underline{69.6} & \underline{66.7} \\
LamRA & 65.1 & \textbf{18.1} & \underline{22.6} & \textbf{33.2} & \textbf{17.5} & \textbf{31.1} & 12.3 & 1.3 & 7.5 & 81.4 & \underline{69.6} & \underline{66.7} \\
LamRA-Qwen2.5VL & \underline{65.2} & 15.2 & 19.8 & 31.9 & \underline{16.3} & \underline{29.7} & 12.8 & 1.3 & 7.0 & 84.5 & 67.9 & 64.3 \\
MM-Embed & 63.3 & 16.2 & 20.1 & 25.5 & 11.3 & 19.6 & \textbf{18.2} & \textbf{3.1} & \textbf{19.0} & 82.7 & 58.9 & 61.9 \\
Qwen3-VL-2B & 64.6 & 13.7 & 17.6 & 28.3 & 11.7 & 24.7 & 11.2 & 1.6 & 10.6 & 78.2 & 57.1 & 59.5 \\
Qwen3-VL-8B & \textbf{70.1} & 17.2 & \textbf{24.1} & \underline{32.6} & 14.7 & 28.5 & 13.3 & \underline{2.1} & \underline{13.5} & \textbf{86.4} & \textbf{71.4} & \textbf{71.4} \\
Rzen-Embed & \textbf{70.1} & \underline{17.4} & 22.3 & 30.8 & 14.9 & 26.2 & 12.8 & 1.7 & 10.7 & \textbf{86.4} & \textbf{71.4} & \underline{66.7} \\
VLM2Vec-V2 & 55.0 & 11.1 & 15.2 & 15.8 & 5.4 & 9.2 & 9.1 & 1.3 & 7.9 & 38.6 & 19.6 & 23.8 \\
Gemini Emb.\ 2 & 49.2 & 11.8 & 15.2 & 24.9 & 11.5 & 18.7 & 12.4 & \underline{2.1} & 10.4 & 74.1 & 62.5 & 61.9 \\
Voyage MM-3.5 & 54.4 & 10.2 & 14.2 & 19.4 & 7.4 & 14.7 & \underline{14.8} & 1.8 & 10.4 & 77.7 & 51.8 & 52.4 \\
\bottomrule
\end{tabular}
}
\end{table*}

\paragraph{Composition gap on the filtered benchmark}
To assess whether the remaining benchmark more genuinely requires multimodal reasoning, we use the normalised composition gap (\emph{CompGap}).
$\mathrm{CompGap} = 1 - \max(I,T)/\mathrm{MM}$, where $\mathrm{MM}$, $I$, and $T$ denote multimodal, image-only, and text-only full-catalogue nDCG, respectively.
Intuitively, $\mathrm{CompGap}$ quantifies the ranking quality gap between the multimodal retriever and the strongest unimodal baseline. Larger values, therefore, indicate a stronger dependence on combining image and text, while values close to zero indicate that one modality alone is sufficient.
We adopt this normalised measure because absolute nDCG values vary substantially across splits: the shortcut-free and validated subsets are inherently harder, leading to lower overall scores. As a result, raw differences (e.g., $\mathrm{MM}-\mathrm{T}$) are not directly comparable across settings. By normalising with respect to the multimodal score, $\mathrm{CompGap}$ enables a fair comparison of modality reliance across benchmarks of different difficulty.

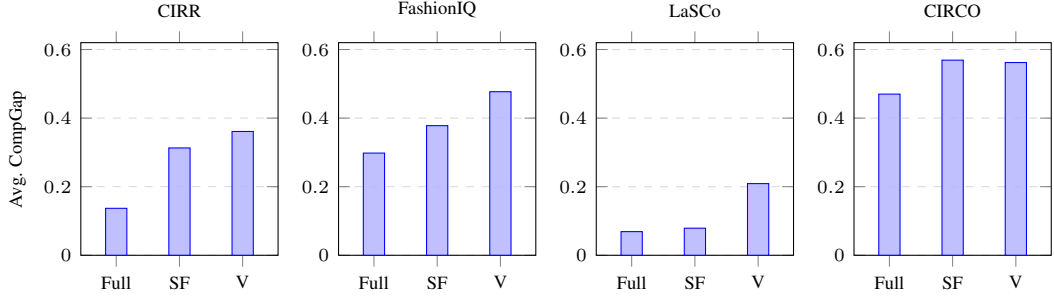
\begin{figure*}[t]
\centering
\begin{tikzpicture}
\begin{groupplot}[
  group style={group size=4 by 1, horizontal sep=0.8cm},
  width=0.3\textwidth,
  height=4.4cm,
  ymin=0.0,
  ymax=0.62,
  ybar,
  enlarge x limits=0.28,
  symbolic x coords={Full,SF,V},
  xtick=data,
  ymajorgrids,
  grid style={dashed,gray!30},
  tick label style={font=\scriptsize},
  title style={font=\scriptsize},
  ylabel style={font=\scriptsize},
  xlabel style={font=\scriptsize},
  every axis plot/.append style={draw=teal!70!black, fill=teal!55, fill opacity=0.82, /tikz/bar width=8pt},
]
\nextgroupplot[title={CIRR}, ylabel={Avg. CompGap}, xlabel={}]
\addplot coordinates {(Full,0.137) (SF,0.313) (V,0.361)};
\nextgroupplot[title={FashionIQ}, ylabel={}, xlabel={}]
\addplot coordinates {(Full,0.298) (SF,0.378) (V,0.477)};
\nextgroupplot[title={LaSCo}, ylabel={}, xlabel={}]
\addplot coordinates {(Full,0.069) (SF,0.079) (V,0.209)};
\nextgroupplot[title={CIRCO}, ylabel={}, xlabel={}]
\addplot coordinates {(Full,0.470) (SF,0.569) (V,0.562)};
\end{groupplot}
\end{tikzpicture}
\caption{Retriever-averaged normalised composition gap based on full-catalogue nDCG. Each panel corresponds to one dataset; the x-axis shows the original benchmark (\textbf{Full}), \circussf{} (\textbf{SF}), and \circusv{} (\textbf{V}). Higher values indicate stronger reliance on multimodal composition.}
\label{fig:avg_compgap_ndcg}
\end{figure*}

As shown in \Cref{fig:avg_compgap_ndcg}, \circusv{} exhibits a markedly higher retriever-averaged \emph{CompGap} than the original benchmark on CIRR (0.361 vs.\ 0.137), FashionIQ (0.477 vs.\ 0.298), and LaSCo (0.209 vs.\ 0.069). Thus, although \circusv{} is harder in absolute Recall@10, a larger fraction of its ranking quality depends on combining both modalities. CIRCO again behaves differently: its full benchmark already shows a high composition gap (0.470), and validation only slightly increases it (0.562), consistent with its relatively clean composition-required core.
Additional full-catalogue evaluation results are reported in Appendices~\ref{app:additional-eval-tables} and~\ref{app:compgap-mrr}, including nDCG@10 and composition-gap variants based on MRR, which follow the same trends as the main analysis. The full MRR tables are also reported for completeness.

\section{Related Work}
\paragraph{Composed Image Retrieval}

Composed Image Retrieval (CIR), introduced by~\citet{DBLP:conf/cvpr/Vo0S0L0H19}, retrieves a target image from a query consisting of a reference image and a textual modification. Standard benchmarks include CIRR~\citep{DBLP:conf/iccv/LiuROG21}, FashionIQ~\citep{DBLP:conf/cvpr/WuGGARGF21}, LaSCo~\citep{DBLP:conf/aaai/LevyBDL24}, and CIRCO~\citep{Baldrati_2023_ICCV}.
Supervised methods are usually trained on annotated data~\citep{DBLP:conf/wacv/AnwaarLK21,DBLP:conf/wacv/JandialBCCSK22}. To reduce annotation cost, subsequent work explores alternatives such as zero-shot pseudo-token approaches~\citep{DBLP:conf/cvpr/SaitoSHLKSG23,DBLP:conf/iccv/BaldratiBUB23,DBLP:conf/cvpr/GuCKKY24}, training-free composition via LLM-based caption rewriting~\citep{DBLP:conf/iclr/KarthikRMA24}, generative editing with diffusion models~\citep{DBLP:journals/tmlr/GuCKJKY24}, embedding interpolation~\citep{DBLP:conf/eccv/JangHSCL24}, and large-scale self-supervised pretraining~\citep{DBLP:conf/icml/0033LHLQC0C24}.
\citet{Huynh2025CoLLM} improve CIRR and FashionIQ by rewriting ambiguous queries and revalidating targets to increase specificity.
Our work is orthogonal: we audit four CIR benchmarks to assess whether their original instances genuinely require multimodal composition by measuring unimodal shortcut solvability across retrievers and validating the remaining shortcut-free residue.
Recent work has shifted toward \emph{Generalist Multimodal Embedding models} (GMEs), which encode text, images, and multimodal queries into a shared space and support diverse retrieval settings without task-specific heads~\citep{DBLP:journals/corr/abs-2412-16855,DBLP:conf/iclr/LinLSLCP25,DBLP:conf/cvpr/LiuZCJ0YWX25,DBLP:journals/corr/abs-2601-03666,DBLP:journals/tmlr/MengJLSYFQTZCXXZCY26,DBLP:journals/corr/abs-2510-27350,li2026qwen3vlembeddingqwen3vlrerankerunifiedframework}.
These models are commonly evaluated through unified benchmarks such as M-BEIR~\citep{DBLP:conf/eccv/WeiCCHZFRC24},
where CIR appears as one task among many, and their strong performance is often taken as evidence of multimodal composition.

\paragraph{Benchmark artefacts}
A long thread of work shows that benchmarks can be partly solved without using all available inputs.
In natural language inference, hypothesis-only models reach well-above-chance accuracy due to annotation artifacts~\citep{DBLP:conf/starsem/PoliakNHRD18,DBLP:conf/naacl/GururanganSLSBS18}, motivating hard subsets in which partial-input baselines fail~\citep{DBLP:conf/emnlp/ClarkYZ19}. In visual question answering, question-only baselines achieve high accuracy~\citep{DBLP:conf/cvpr/GoyalKSBP17,DBLP:conf/cvpr/AgrawalBPK18}, motivating debiased training~\citep{DBLP:conf/nips/RamakrishnanAL18,DBLP:conf/nips/CadeneDCPC19,DBLP:conf/cvpr/NiuTZL0W21} and counterexample-based diagnostics~\citep{DBLP:conf/iccv/DancetteCTC21}.
More broadly, multimodal models tend to rely on the most predictive modality~\citep{DBLP:conf/icml/WuJGS22}, often defaulting to language-dominated representations~\citep{DBLP:conf/emnlp/FrankBE21,DBLP:journals/corr/abs-2505-18657}; image--text matchers behave like bag-of-words~\citep{DBLP:conf/iclr/YuksekgonulBKJZ23}; and benchmarks such as Winoground~\citep{DBLP:conf/cvpr/ThrushJBSWKR22} reveal limited compositional ability. \citet{DBLP:conf/emnlp/HesselL20} and \citet{DBLP:journals/natmi/GeirhosJMZBBW20} document how readily such shortcuts arise.
\emph{Unimodal shortcuts} and \emph{annotation artefacts} thus describe two facets of the same phenomenon: shortcuts characterise model behaviour, while artefacts characterise the dataset patterns that enable it; both are diagnosed by the same tool—evaluation with one input ablated.
We bring this diagnostic to CIR, where it has not been systematically applied across benchmarks and GMEs, and identify shortcut solvability as a dataset-level property by aggregating across models.

\section{Conclusion}
We investigated whether current CIR benchmarks genuinely require multimodal composition through a two-stage audit.
A cross-model shortcut analysis shows that large portions of existing benchmarks are solvable using a single modality, while a human audit of 4{,}741 shortcut-free queries reveals that many remaining instances are ambiguous or ill-posed, with only 1{,}689 fully valid cases.
Re-evaluation on our filtered benchmark (\circus{} splits) shows that multimodal reasoning is most clearly required on the validated subset, where performance depends more strongly on combining both modalities.
Reliable evaluation, therefore, requires separating these factors, for example, by reporting results on shortcut-free and validated subsets rather than relying on a single benchmark score.
While our analysis depends on the evaluated retriever pool and the choice of cutoff, and human validation is partially estimated for larger datasets, the main conclusions remain stable across retriever subsets, cutoffs, and uncertainty estimates. Future work could extend this validation protocol to larger portions of existing benchmarks, enabling more comprehensive evaluation under the same principles.
Taken together, our results highlight \emph{a key limitation of current CIR benchmarks}: performance is often driven by unimodal shortcuts.
We argue that progress in CIR should be measured by the extent to which retrievers genuinely combine image and text. Although more challenging, \circus{} provides a more faithful evaluation of this ability.

\section*{Acknowledgements}
%
This work has been carried out while \textit{Matteo Attimonelli} was enrolled in the Italian National Doctorate on Artificial Intelligence run by Sapienza University of Rome in collaboration with \textit{Politecnico Di Bari}. 
APG was supported by the United Kingdom Research and Innovation (grant EP/S02431X/1), UKRI Centre for Doctoral Training in Biomedical AI at the University of Edinburgh, School of Informatics.
Rohit Saxena was supported by the Engineering and Physical Sciences Research Council (EPSRC) through the AI Hub in Generative Models (grant number EP/Y028805/1).
%
This work was supported by the Edinburgh International Data Facility (EIDF) and the Data-Driven Innovation Programme at the University of Edinburgh.
%
%
%
We acknowledge the CINECA award under the ISCRA initiative for the availability of high-performance computing resources and support. We acknowledge ISCRA for awarding this project access to the LEONARDO supercomputer, owned by the EuroHPC Joint Undertaking, hosted by CINECA (Italy).

\bibliography{references}
\bibliographystyle{unsrtnat}

\clearpage
\section*{NeurIPS Paper Checklist}

\begin{enumerate}

\item {\bf Claims}
    \item[] Question: Do the main claims made in the abstract and introduction accurately reflect the paper's contributions and scope?
    \item[] Answer: \answerYes{}
    \item[] Justification: The abstract and Introduction state the paper's three core claims: current CIR benchmarks contain unimodal shortcuts, shortcut-free queries still contain many invalid instances, and validated subsets better diagnose multimodal composition. These claims are supported by the shortcut audit, human validation study, and re-evaluation analyses in Sections 3--6.

\item {\bf Limitations}
    \item[] Question: Does the paper discuss the limitations of the work performed by the authors?
    \item[] Answer: \answerYes{}
    \item[] Justification: The Conclusion contains an explicit discussion of limitations. It covers dependence on the evaluated retriever pool, sensitivity to the retrieval cutoff and unimodal ablation design, and the sampling uncertainty introduced by partial human validation on FashionIQ and LaSCo, with supporting robustness and uncertainty analyses in Appendices~\ref{app:shortcut-robustness} and~\ref{app:uncertainty}.

\item {\bf Theory assumptions and proofs}
    \item[] Question: For each theoretical result, does the paper provide the full set of assumptions and a complete (and correct) proof?
    \item[] Answer: \answerNA{}
    \item[] Justification: The paper does not present new theoretical results or formal proofs; it is an empirical audit of composed image retrieval benchmarks.

    \item {\bf Experimental result reproducibility}
    \item[] Question: Does the paper fully disclose all the information needed to reproduce the main experimental results of the paper to the extent that it affects the main claims and/or conclusions of the paper (regardless of whether the code and data are provided or not)?
    \item[] Answer: \answerYes{}
    \item[] Justification: The paper specifies the evaluated retriever pool, datasets, shortcut criterion, human validation protocol, and ranking metrics used for the main results. The anonymized repository linked in the abstract additionally provides the filtered split definitions, validation artifacts, and evaluation scripts used to regenerate the reported tables.

\item {\bf Open access to data and code}
    \item[] Question: Does the paper provide open access to the data and code, with sufficient instructions to faithfully reproduce the main experimental results, as described in supplemental material?
    \item[] Answer: \answerYes{}
    \item[] Justification: The abstract links to an anonymized repository containing the evaluation code, filtered and validated split files, and instructions for reproducing the main analyses and manuscript tables.

\item {\bf Experimental setting/details}
    \item[] Question: Does the paper specify all the training and test details (e.g., data splits, hyperparameters, how they were chosen, type of optimizer) necessary to understand the results?
    \item[] Answer: \answerYes{}
    \item[] Justification: Section 4 describes the evaluated models, datasets, gallery sizes, retrieval protocol, and metrics. No model is trained in this work; the experiments evaluate fixed retrievers under multimodal, image-only, and text-only query conditions.

\item {\bf Experiment statistical significance}
    \item[] Question: Does the paper report error bars suitably and correctly defined or other appropriate information about the statistical significance of the experiments?
    \item[] Answer: \answerYes{}
    \item[] Justification: Appendix~\ref{app:uncertainty} reports 95\% confidence intervals for the filtered-set retrieval analysis using non-parametric query bootstrap, including multimodal retrieval metrics and paired multimodal-versus-unimodal comparisons.


\item {\bf Experiments compute resources}
    \item[] Question: For each experiment, does the paper provide sufficient information on the computer resources (type of compute workers, memory, time of execution) needed to reproduce the experiments?
    \item[] Answer: \answerYes{}
    \item[] Justification: Appendix~\ref{app:checklist-details} reports the hardware used (single H100 NVL GPU, AMD EPYC Genoa CPU, 256\,GB RAM), representative GPU memory footprints and representative per-dataset runtimes for the main open-weight experiments.

\item {\bf Code of ethics}
    \item[] Question: Does the research conducted in the paper conform, in every respect, with the NeurIPS Code of Ethics \url{https://neurips.cc/public/EthicsGuidelines}?
    \item[] Answer: \answerYes{}
    \item[] Justification: The work audits existing benchmarks and evaluated retrievers, preserves submission anonymity, and discloses the human validation component separately. We are not aware of any aspect of the study that deviates from the NeurIPS Code of Ethics.

\item {\bf Broader impacts}
    \item[] Question: Does the paper discuss both potential positive societal impacts and negative societal impacts of the work performed?
    \item[] Answer: \answerYes{}
    \item[] Justification: Appendix~\ref{app:checklist-details} discusses the broader impacts of the work, including improved evaluation reliability and transparency, and describes how releasing derived artifacts with full provenance supports reproducibility and responsible use.

\item {\bf Safeguards}
    \item[] Question: Does the paper describe safeguards that have been put in place for responsible release of data or models that have a high risk for misuse (e.g., pre-trained language models, image generators, or scraped datasets)?
    \item[] Answer: \answerNA{}
    \item[] Justification: The paper does not release a high-risk generative model or a newly scraped dataset. The released artifacts are evaluation code and filtered subsets derived from existing benchmarks.

\item {\bf Licenses for existing assets}
    \item[] Question: Are the creators or original owners of assets (e.g., code, data, models), used in the paper, properly credited and are the license and terms of use explicitly mentioned and properly respected?
    \item[] Answer: \answerYes{}
    \item[] Justification: Appendix~\ref{app:checklist-details} explicitly documents the benchmark assets used in the paper, their upstream licenses or terms of use, and our redistribution policy. The main text also cites the original benchmark and model papers.

\item {\bf New assets}
    \item[] Question: Are new assets introduced in the paper well documented and is the documentation provided alongside the assets?
    \item[] Answer: \answerYes{}
    \item[] Justification: The new assets are the validated split definitions, query identifiers, and annotation outputs introduced by the audit. Their construction is described in the Human Validation Protocol section, and the supplemental material documents the released files alongside the evaluation code.

\item {\bf Crowdsourcing and research with human subjects}
    \item[] Question: For crowdsourcing experiments and research with human subjects, does the paper include the full text of instructions given to participants and screenshots, if applicable, as well as details about compensation (if any)?
    \item[] Answer: \answerYes{}
    \item[] Justification: Appendix~\ref{app:human-validation-materials} describes the annotation interface, includes a screenshot, reproduces the main instructions used in the study, and clarifies that the work was performed internally by trained researchers rather than via a paid crowdsourcing platform, with no separate crowdsourcing compensation scheme.

\item {\bf Institutional review board (IRB) approvals or equivalent for research with human subjects}
    \item[] Question: Does the paper describe potential risks incurred by study participants, whether such risks were disclosed to the subjects, and whether Institutional Review Board (IRB) approvals (or an equivalent approval/review based on the requirements of your country or institution) were obtained?
    \item[] Answer: \answerNA{}
    \item[] Justification: Appendix~\ref{app:human-validation-materials} states that the study used public benchmark triples, required no sensitive personal data, was treated as minimal-risk internal research, and that no formal IRB or equivalent ethics approval was obtained for this annotation effort.

\item {\bf Declaration of LLM usage}
    \item[] Question: Does the paper describe the usage of LLMs if it is an important, original, or non-standard component of the core methods in this research? Note that if the LLM is used only for writing, editing, or formatting purposes and does \emph{not} impact the core methodology, scientific rigor, or originality of the research, declaration is not required.
    \item[] Answer: \answerNA{}
    \item[] Justification: LLMs, if used at all, were used only for writing or editing assistance and are not part of the methodology.

\end{enumerate}

\clearpage
\appendix

\section{Extended nDCG and MRR Results}
\label{app:additional-eval-tables}
This appendix reports the \circus{} evaluation requested in the final audit. \Cref{tab:final_ndcg,tab:final_mrr} report the corresponding nDCG and MRR analyses on the original full benchmark, on \circussf{} (SF), and on \circusv{} (V). Both metrics are computed on the full gallery ordering without any cutoff. \Cref{tab:final_ndcg_at10} adds a cutoff-based robustness view by recomputing the
compact nDCG table at $K=10$ with the same Full/SF/V layout.
Overall, these extended metrics support the same interpretation as Figure~3.
On CIRR, LaSCo, and CIRCO, absolute MM scores decrease from Full to SF/V, but the multimodal query retains a positive advantage over both unimodal variants.
FashionIQ is the main exception in absolute performance: validated scores recover close to Full, while $\Delta$MM-T increases, suggesting that validation removes noisy examples without weakening the modality-composition signal.
The MRR and nDCG@10 tables confirm that this pattern is not specific to full-catalogue nDCG, but also appears in early-rank and cutoff-based metrics.
\begin{table*}[t]
\centering
\caption{Compact full-catalogue nDCG (\%) for each retriever, computed without cutoff over the full gallery ranking. We report the original benchmark (\textbf{Full}) together with the \emph{shortcut-free} subset (\textbf{SF}) and the \emph{validated} subset (\textbf{V}). For each split, we show the multimodal query score (MM) and the three signed delta columns $\Delta$MM-I, $\Delta$MM-T, and $\Delta$I-T, where I and T denote image-only and text-only queries.}
\label{tab:final_ndcg}
\scriptsize
\setlength{\tabcolsep}{3pt}
\resizebox{\textwidth}{!}{
\begin{tabular}{l l cccc cccc cccc}
\toprule
 &  & \multicolumn{4}{c}{\textbf{Full}} & \multicolumn{4}{c}{\textbf{SF}} & \multicolumn{4}{c}{\textbf{V}} \\
\cmidrule(lr){3-6} \cmidrule(lr){7-10} \cmidrule(lr){11-14}
\textbf{Dataset} & \textbf{Retriever} & \textbf{MM} & \textbf{$\Delta$MM-I} & \textbf{$\Delta$MM-T} & \textbf{$\Delta$I-T} & \textbf{MM} & \textbf{$\Delta$MM-I} & \textbf{$\Delta$MM-T} & \textbf{$\Delta$I-T} & \textbf{MM} & \textbf{$\Delta$MM-I} & \textbf{$\Delta$MM-T} & \textbf{$\Delta$I-T} \\
\midrule
\multirow{12}{*}{\textbf{CIRR}} & E5-Omni & 32.4 & 14.2 & 0.4 & -13.8 & 18.4 & 5.6 & 4.9 & -0.7 & 20.0 & 6.5 & 6.4 & -0.1 \\
 & GME-Qwen2VL & 38.3 & 20.3 & 4.4 & -15.8 & 20.5 & 8.0 & 6.4 & -1.5 & 22.7 & 9.4 & 8.3 & -1.1 \\
 & LamRA & 38.8 & 20.6 & 6.0 & -14.6 & 21.4 & 8.3 & 7.1 & -1.2 & 23.4 & 9.7 & 9.2 & -0.5 \\
 & LamRA-Qwen2.5VL & 38.6 & 20.5 & 6.4 & -14.1 & 20.6 & 7.8 & 6.9 & -0.9 & 23.0 & 9.5 & 9.3 & -0.3 \\
 & MM-Embed & 37.5 & 19.2 & 5.9 & -13.3 & 20.7 & 7.7 & 7.3 & -0.4 & 22.7 & 9.0 & 9.2 & 0.2 \\
 & Qwen3-VL-2B & 38.9 & 20.3 & 6.5 & -13.9 & 19.8 & 6.8 & 6.4 & -0.4 & 21.7 & 7.9 & 8.1 & 0.2 \\
 & Qwen3-VL-8B & 41.3 & 22.5 & 10.0 & -12.5 & 21.2 & 8.0 & 8.2 & 0.2 & 24.0 & 9.9 & 10.7 & 0.8 \\
 & Rzen-Embed & 41.0 & 22.9 & 5.1 & -17.8 & 21.7 & 9.0 & 7.3 & -1.7 & 24.1 & 10.6 & 9.4 & -1.2 \\
 & VLM2Vec-V2 & 32.9 & 14.9 & 4.2 & -10.7 & 18.3 & 5.5 & 5.4 & -0.1 & 20.2 & 6.6 & 7.1 & 0.5 \\
 & Gemini Emb.\ 2 & 30.7 & 12.1 & 4.9 & -7.2 & 17.5 & 4.5 & 4.8 & 0.3 & 19.2 & 5.3 & 6.2 & 0.9 \\
 & Voyage MM-3.5 & 33.8 & 15.9 & 2.9 & -13.0 & 18.2 & 5.5 & 4.7 & -0.7 & 20.3 & 6.9 & 6.6 & -0.3 \\
\midrule
\multirow{12}{*}{\textbf{FIQ}} & E5-Omni & 17.8 & 7.8 & 2.8 & -4.9 & 14.1 & 4.9 & 3.9 & -1.0 & 17.4 & 7.5 & 6.5 & -1.0 \\
 & GME-Qwen2VL & 24.8 & 14.5 & 8.3 & -6.2 & 18.4 & 9.1 & 8.0 & -1.2 & 23.8 & 13.7 & 12.6 & -1.2 \\
 & LamRA & 25.6 & 15.6 & 10.3 & -5.3 & 19.5 & 10.3 & 9.1 & -1.2 & 25.8 & 15.9 & 14.7 & -1.1 \\
 & LamRA-Qwen2.5VL & 25.1 & 15.2 & 9.3 & -5.9 & 18.9 & 9.8 & 8.3 & -1.5 & 25.2 & 15.4 & 13.9 & -1.5 \\
 & MM-Embed & 22.4 & 12.2 & 6.9 & -5.3 & 16.7 & 7.5 & 6.3 & -1.1 & 21.4 & 11.4 & 10.2 & -1.2 \\
 & Qwen3-VL-2B & 23.6 & 13.3 & 7.4 & -5.9 & 17.0 & 7.7 & 6.5 & -1.1 & 23.0 & 13.0 & 11.5 & -1.5 \\
 & Qwen3-VL-8B & 25.5 & 15.1 & 7.9 & -7.2 & 18.4 & 9.0 & 7.4 & -1.6 & 25.0 & 14.9 & 12.6 & -2.3 \\
 & Rzen-Embed & 24.7 & 14.5 & 8.0 & -6.5 & 18.5 & 9.2 & 7.8 & -1.3 & 24.2 & 14.1 & 12.5 & -1.6 \\
 & VLM2Vec-V2 & 17.8 & 7.7 & 4.8 & -2.9 & 13.4 & 4.2 & 3.8 & -0.4 & 16.2 & 6.3 & 6.1 & -0.2 \\
 & Gemini Emb.\ 2 & 21.5 & 10.8 & 4.9 & -5.9 & 16.8 & 7.1 & 5.9 & -1.2 & 21.3 & 10.9 & 9.1 & -1.8 \\
 & Voyage MM-3.5 & 19.5 & 9.5 & 5.1 & -4.4 & 14.7 & 5.5 & 4.7 & -0.8 & 18.4 & 8.5 & 7.7 & -0.8 \\
\midrule
\multirow{12}{*}{\textbf{LaSCo}} & E5-Omni & 15.5 & 1.1 & 1.2 & 0.2 & 12.0 & 0.5 & 1.8 & 1.3 & 14.4 & 1.3 & 3.3 & 2.0 \\
 & GME-Qwen2VL & 17.2 & 2.6 & 0.8 & -1.8 & 12.7 & 1.1 & 1.4 & 0.4 & 17.7 & 4.2 & 5.0 & 0.8 \\
 & LamRA & 16.8 & 2.4 & 1.7 & -0.7 & 12.6 & 1.0 & 1.7 & 0.6 & 16.5 & 3.2 & 4.8 & 1.6 \\
 & LamRA-Qwen2.5VL & 17.1 & 2.5 & 1.3 & -1.2 & 12.8 & 1.1 & 1.6 & 0.5 & 16.7 & 3.4 & 4.4 & 1.0 \\
 & MM-Embed & 19.2 & 4.8 & 3.9 & -0.9 & 13.7 & 2.2 & 3.0 & 0.7 & 20.3 & 6.8 & 8.6 & 1.7 \\
 & Qwen3-VL-2B & 16.2 & 1.7 & 0.4 & -1.3 & 12.2 & 0.6 & 1.0 & 0.4 & 16.4 & 3.0 & 4.0 & 1.0 \\
 & Qwen3-VL-8B & 17.0 & 2.2 & 0.4 & -1.8 & 12.8 & 1.0 & 1.3 & 0.4 & 17.8 & 4.1 & 4.8 & 0.7 \\
 & Rzen-Embed & 16.8 & 2.4 & -0.3 & -2.6 & 12.4 & 0.9 & 1.1 & 0.2 & 17.6 & 4.1 & 4.5 & 0.4 \\
 & VLM2Vec-V2 & 15.4 & 0.7 & 2.1 & 1.3 & 12.0 & 0.3 & 1.8 & 1.5 & 15.5 & 2.0 & 4.6 & 2.5 \\
 & Gemini Emb.\ 2 & 16.4 & 2.4 & 1.7 & -0.7 & 12.2 & 1.0 & 1.8 & 0.9 & 16.4 & 3.5 & 5.0 & 1.6 \\
 & Voyage MM-3.5 & 17.7 & 3.1 & 1.5 & -1.6 & 13.0 & 1.4 & 1.9 & 0.5 & 17.6 & 4.3 & 5.2 & 0.9 \\
\midrule
\multirow{12}{*}{\textbf{CIRCO}} & E5-Omni & 34.8 & 13.7 & 15.2 & 1.6 & 28.0 & 13.2 & 15.4 & 2.3 & 27.6 & 12.9 & 15.2 & 2.3 \\
 & GME-Qwen2VL & 46.6 & 24.8 & 24.6 & -0.2 & 40.6 & 26.3 & 27.8 & 1.6 & 39.6 & 25.1 & 27.0 & 1.9 \\
 & LamRA & 45.4 & 23.5 & 25.6 & 2.1 & 39.0 & 24.4 & 27.7 & 3.3 & 37.9 & 22.9 & 27.0 & 4.1 \\
 & LamRA-Qwen2.5VL & 46.6 & 25.2 & 27.1 & 1.9 & 39.2 & 24.6 & 27.9 & 3.4 & 38.8 & 24.0 & 28.0 & 4.0 \\
 & MM-Embed & 44.2 & 22.4 & 23.2 & 0.8 & 36.2 & 21.7 & 24.7 & 3.0 & 37.2 & 22.4 & 25.8 & 3.4 \\
 & Qwen3-VL-2B & 42.9 & 22.0 & 22.2 & 0.2 & 36.7 & 22.5 & 24.4 & 1.9 & 36.9 & 22.5 & 25.1 & 2.6 \\
 & Qwen3-VL-8B & 47.0 & 24.5 & 27.1 & 2.6 & 39.0 & 23.5 & 26.8 & 3.3 & 40.1 & 24.2 & 28.0 & 3.9 \\
 & Rzen-Embed & 48.2 & 26.2 & 24.4 & -1.8 & 42.3 & 27.3 & 28.9 & 1.6 & 40.9 & 25.7 & 27.8 & 2.1 \\
 & VLM2Vec-V2 & 27.5 & 6.4 & 12.8 & 6.4 & 19.6 & 6.0 & 9.3 & 3.2 & 20.7 & 6.8 & 10.7 & 3.9 \\
 & Gemini Emb.\ 2 & 38.9 & 17.4 & 19.3 & 1.9 & 34.6 & 19.4 & 22.5 & 3.1 & 33.0 & 17.7 & 21.2 & 3.5 \\
 & Voyage MM-3.5 & 40.9 & 18.7 & 20.7 & 1.9 & 31.6 & 17.7 & 19.4 & 1.7 & 31.9 & 17.7 & 19.6 & 1.9 \\
\bottomrule
\end{tabular}
}
\end{table*}

\begin{table*}[t]
\centering
\caption{Compact full-catalogue MRR for each retriever, computed without cutoff over the full gallery ranking. We report the original benchmark (\textbf{Full}) together with the \emph{shortcut-free} subset (\textbf{SF}) and the \emph{validated} subset (\textbf{V}). For each split we show the multimodal query score (MM) and the three signed delta columns $\Delta$MM-I, $\Delta$MM-T, and $\Delta$I-T, where I and T denote image-only and text-only queries.}
\label{tab:final_mrr}
\scriptsize
\setlength{\tabcolsep}{3pt}
\resizebox{\textwidth}{!}{
\begin{tabular}{l l cccc cccc cccc}
\toprule
 &  & \multicolumn{4}{c}{\textbf{Full}} & \multicolumn{4}{c}{\textbf{SF}} & \multicolumn{4}{c}{\textbf{V}} \\
\cmidrule(lr){3-6} \cmidrule(lr){7-10} \cmidrule(lr){11-14}
\textbf{Dataset} & \textbf{Retriever} & \textbf{MM} & \textbf{$\Delta$MM-I} & \textbf{$\Delta$MM-T} & \textbf{$\Delta$I-T} & \textbf{MM} & \textbf{$\Delta$MM-I} & \textbf{$\Delta$MM-T} & \textbf{$\Delta$I-T} & \textbf{MM} & \textbf{$\Delta$MM-I} & \textbf{$\Delta$MM-T} & \textbf{$\Delta$I-T} \\
\midrule
\multirow{12}{*}{\textbf{CIRR}} & E5-Omni & 15.9 & 11.2 & -0.8 & -12.0 & 4.0 & 3.2 & 3.0 & -0.3 & 5.2 & 4.2 & 4.1 & -0.2 \\
 & GME-Qwen2VL & 22.3 & 17.7 & 3.8 & -14.0 & 5.7 & 5.0 & 4.6 & -0.4 & 7.4 & 6.6 & 6.2 & -0.4 \\
 & LamRA & 22.9 & 18.3 & 5.4 & -12.9 & 6.3 & 5.4 & 5.1 & -0.3 & 7.9 & 6.8 & 6.7 & -0.2 \\
 & LamRA-Qwen2.5VL & 22.7 & 18.1 & 5.7 & -12.5 & 5.7 & 4.9 & 4.7 & -0.2 & 7.8 & 6.8 & 6.7 & -0.1 \\
 & MM-Embed & 21.3 & 16.7 & 4.9 & -11.8 & 5.7 & 4.9 & 4.7 & -0.1 & 7.4 & 6.4 & 6.4 & 0.0 \\
 & Qwen3-VL-2B & 23.1 & 18.2 & 5.8 & -12.4 & 5.1 & 4.2 & 4.1 & -0.1 & 6.6 & 5.5 & 5.5 & 0.0 \\
 & Qwen3-VL-8B & 25.4 & 20.4 & 9.4 & -11.0 & 6.1 & 5.2 & 5.2 & 0.1 & 8.3 & 7.2 & 7.4 & 0.3 \\
 & Rzen-Embed & 25.2 & 20.5 & 4.6 & -15.9 & 6.6 & 5.8 & 5.4 & -0.5 & 8.6 & 7.7 & 7.2 & -0.4 \\
 & VLM2Vec-V2 & 16.7 & 12.1 & 3.0 & -9.1 & 4.0 & 3.2 & 3.1 & 0.0 & 5.4 & 4.4 & 4.5 & 0.1 \\
 & Gemini Emb.\ 2 & 14.8 & 9.8 & 4.2 & -5.6 & 3.7 & 2.7 & 2.7 & 0.0 & 4.8 & 3.6 & 3.7 & 0.1 \\
 & Voyage MM-3.5 & 17.8 & 13.4 & 2.1 & -11.3 & 3.9 & 3.2 & 2.9 & -0.2 & 5.5 & 4.6 & 4.4 & -0.2 \\
\midrule
\multirow{12}{*}{\textbf{FIQ}} & E5-Omni & 5.0 & 4.2 & 1.2 & -3.0 & 2.3 & 2.1 & 2.0 & -0.1 & 4.1 & 3.8 & 3.7 & -0.1 \\
 & GME-Qwen2VL & 10.7 & 9.9 & 5.8 & -4.1 & 5.2 & 5.0 & 4.8 & -0.2 & 9.2 & 8.9 & 8.7 & -0.2 \\
 & LamRA & 11.3 & 10.7 & 7.5 & -3.2 & 5.9 & 5.7 & 5.6 & -0.2 & 10.8 & 10.5 & 10.4 & -0.2 \\
 & LamRA-Qwen2.5VL & 10.8 & 10.2 & 6.7 & -3.5 & 5.4 & 5.2 & 5.0 & -0.2 & 10.3 & 10.1 & 9.8 & -0.3 \\
 & MM-Embed & 8.7 & 7.9 & 4.6 & -3.3 & 3.9 & 3.7 & 3.5 & -0.2 & 7.2 & 6.8 & 6.6 & -0.2 \\
 & Qwen3-VL-2B & 9.8 & 9.1 & 5.2 & -3.8 & 4.1 & 3.9 & 3.8 & -0.1 & 8.5 & 8.2 & 7.9 & -0.3 \\
 & Qwen3-VL-8B & 11.4 & 10.5 & 5.8 & -4.7 & 5.0 & 4.8 & 4.6 & -0.2 & 10.0 & 9.7 & 9.1 & -0.5 \\
 & Rzen-Embed & 10.7 & 10.0 & 5.8 & -4.2 & 5.2 & 5.0 & 4.8 & -0.2 & 9.4 & 9.0 & 8.8 & -0.3 \\
 & VLM2Vec-V2 & 5.3 & 4.5 & 2.8 & -1.7 & 2.0 & 1.8 & 1.7 & 0.0 & 3.5 & 3.2 & 3.2 & 0.0 \\
 & Gemini Emb.\ 2 & 7.8 & 6.8 & 3.0 & -3.8 & 4.0 & 3.7 & 3.5 & -0.2 & 6.8 & 6.4 & 6.0 & -0.4 \\
 & Voyage MM-3.5 & 6.5 & 5.8 & 3.1 & -2.7 & 2.7 & 2.5 & 2.4 & -0.1 & 5.0 & 4.7 & 4.5 & -0.2 \\
\midrule
\multirow{12}{*}{\textbf{LaSCo}} & E5-Omni & 2.9 & 0.6 & -0.1 & -0.7 & 0.7 & 0.2 & 0.3 & 0.2 & 1.5 & 0.6 & 1.0 & 0.4 \\
 & GME-Qwen2VL & 4.0 & 1.7 & -0.1 & -1.8 & 1.0 & 0.5 & 0.6 & 0.0 & 3.8 & 2.8 & 2.9 & 0.2 \\
 & LamRA & 3.7 & 1.5 & 0.5 & -1.0 & 0.9 & 0.4 & 0.5 & 0.1 & 2.8 & 1.8 & 2.2 & 0.3 \\
 & LamRA-Qwen2.5VL & 3.8 & 1.6 & 0.2 & -1.4 & 1.0 & 0.5 & 0.5 & 0.0 & 2.8 & 1.9 & 2.1 & 0.1 \\
 & MM-Embed & 5.4 & 3.2 & 2.0 & -1.2 & 1.5 & 1.0 & 1.1 & 0.1 & 5.5 & 4.5 & 4.9 & 0.4 \\
 & Qwen3-VL-2B & 3.4 & 1.2 & -0.3 & -1.4 & 0.9 & 0.4 & 0.4 & 0.0 & 3.1 & 2.2 & 2.3 & 0.1 \\
 & Qwen3-VL-8B & 3.9 & 1.5 & -0.4 & -1.8 & 1.1 & 0.5 & 0.5 & 0.0 & 3.7 & 2.7 & 2.8 & 0.1 \\
 & Rzen-Embed & 3.8 & 1.5 & -0.9 & -2.5 & 1.0 & 0.5 & 0.4 & 0.0 & 3.6 & 2.6 & 2.6 & 0.0 \\
 & VLM2Vec-V2 & 2.8 & 0.5 & 0.7 & 0.2 & 0.8 & 0.3 & 0.5 & 0.2 & 2.4 & 1.5 & 1.9 & 0.5 \\
 & Gemini Emb.\ 2 & 3.7 & 1.5 & 0.4 & -1.2 & 1.0 & 0.5 & 0.6 & 0.1 & 3.2 & 2.2 & 2.5 & 0.3 \\
 & Voyage MM-3.5 & 4.3 & 2.0 & 0.2 & -1.7 & 1.1 & 0.6 & 0.6 & 0.0 & 3.4 & 2.5 & 2.7 & 0.2 \\
\midrule
\multirow{12}{*}{\textbf{CIRCO}} & E5-Omni & 17.5 & 11.3 & 11.1 & -0.2 & 11.3 & 9.8 & 10.5 & 0.7 & 11.0 & 9.5 & 10.2 & 0.7 \\
 & GME-Qwen2VL & 30.7 & 24.0 & 22.3 & -1.7 & 24.7 & 23.5 & 23.8 & 0.2 & 23.8 & 22.6 & 23.0 & 0.4 \\
 & LamRA & 29.5 & 22.8 & 22.7 & -0.1 & 22.9 & 21.6 & 22.5 & 0.9 & 22.0 & 20.5 & 21.7 & 1.1 \\
 & LamRA-Qwen2.5VL & 30.7 & 24.5 & 24.2 & -0.2 & 23.1 & 21.8 & 22.6 & 0.8 & 23.1 & 21.7 & 22.8 & 1.1 \\
 & MM-Embed & 27.9 & 21.2 & 19.7 & -1.4 & 19.9 & 18.7 & 19.5 & 0.8 & 21.1 & 19.8 & 20.7 & 0.9 \\
 & Qwen3-VL-2B & 26.6 & 20.8 & 19.3 & -1.5 & 20.9 & 19.7 & 20.1 & 0.4 & 21.1 & 19.8 & 20.5 & 0.7 \\
 & Qwen3-VL-8B & 31.1 & 24.0 & 24.3 & 0.2 & 22.3 & 20.7 & 21.7 & 1.0 & 23.6 & 21.8 & 23.0 & 1.3 \\
 & Rzen-Embed & 32.5 & 25.7 & 22.8 & -2.8 & 26.4 & 25.0 & 25.4 & 0.5 & 25.1 & 23.6 & 24.2 & 0.6 \\
 & VLM2Vec-V2 & 11.6 & 5.4 & 8.2 & 2.8 & 5.3 & 4.3 & 5.0 & 0.6 & 6.1 & 5.0 & 5.9 & 0.9 \\
 & Gemini Emb.\ 2 & 22.0 & 15.9 & 15.2 & -0.7 & 18.0 & 16.4 & 17.3 & 0.9 & 16.2 & 14.6 & 15.6 & 1.0 \\
 & Voyage MM-3.5 & 24.2 & 17.4 & 17.2 & -0.2 & 15.0 & 13.9 & 14.2 & 0.2 & 15.5 & 14.4 & 14.6 & 0.2 \\
\bottomrule
\end{tabular}
}
\end{table*}

\begin{table*}[t]
\centering
\caption{Compact nDCG@10 (\%) for each retriever, computed with ranking gain truncated at position 10. We report the original benchmark (\textbf{Full}) together with the \emph{shortcut-free} subset (\textbf{SF}) and the \emph{validated} subset (\textbf{V}). For each split, we show the multimodal query score (MM) and the three signed delta columns $\Delta$MM-I, $\Delta$MM-T, and $\Delta$I-T, where I and T denote image-only and text-only queries.}
\label{tab:final_ndcg_at10}
\scriptsize
\setlength{\tabcolsep}{3pt}
\resizebox{\textwidth}{!}{
\begin{tabular}{l l cccc cccc cccc}
\toprule
 &  & \multicolumn{4}{c}{\textbf{Full}} & \multicolumn{4}{c}{\textbf{SF}} & \multicolumn{4}{c}{\textbf{V}} \\
\cmidrule(lr){3-6} \cmidrule(lr){7-10} \cmidrule(lr){11-14}
\textbf{Dataset} & \textbf{Retriever} & \textbf{MM} & \textbf{$\Delta$MM-I} & \textbf{$\Delta$MM-T} & \textbf{$\Delta$I-T} & \textbf{MM} & \textbf{$\Delta$MM-I} & \textbf{$\Delta$MM-T} & \textbf{$\Delta$I-T} & \textbf{MM} & \textbf{$\Delta$MM-I} & \textbf{$\Delta$MM-T} & \textbf{$\Delta$I-T} \\
\midrule
\multirow{12}{*}{\textbf{CIRR}} & E5-Omni & 23.7 & 17.1 & 0.9 & -16.2 & 4.7 & 4.7 & 4.7 & 0.0 & 6.5 & 6.5 & 6.5 & 0.0 \\
 & GME-Qwen2VL & 31.2 & 25.1 & 6.0 & -19.0 & 6.8 & 6.8 & 6.8 & 0.0 & 9.3 & 9.3 & 9.3 & 0.0 \\
 & LamRA & 31.7 & 25.4 & 7.9 & -17.5 & 7.6 & 7.6 & 7.6 & 0.0 & 10.1 & 10.1 & 10.1 & 0.0 \\
 & LamRA-Qwen2.5VL & 31.6 & 25.4 & 8.7 & -16.7 & 6.5 & 6.5 & 6.5 & 0.0 & 9.3 & 9.3 & 9.3 & 0.0 \\
 & MM-Embed & 30.0 & 23.7 & 7.7 & -16.0 & 6.8 & 6.8 & 6.8 & 0.0 & 9.3 & 9.3 & 9.3 & 0.0 \\
 & Qwen3-VL-2B & 31.9 & 25.1 & 8.5 & -16.5 & 5.8 & 5.8 & 5.8 & 0.0 & 7.8 & 7.8 & 7.8 & 0.0 \\
 & Qwen3-VL-8B & 35.2 & 28.2 & 13.1 & -15.0 & 7.3 & 7.3 & 7.3 & 0.0 & 10.9 & 10.9 & 10.9 & 0.0 \\
 & Rzen-Embed & 34.9 & 28.4 & 6.7 & -21.6 & 7.7 & 7.7 & 7.7 & 0.0 & 10.5 & 10.5 & 10.5 & 0.0 \\
 & VLM2Vec-V2 & 24.4 & 17.9 & 5.9 & -12.0 & 4.4 & 4.4 & 4.4 & 0.0 & 6.6 & 6.6 & 6.6 & 0.0 \\
 & Gemini Emb.\ 2 & 21.6 & 14.6 & 6.0 & -8.5 & 4.6 & 4.6 & 4.6 & 0.0 & 6.3 & 6.3 & 6.3 & 0.0 \\
 & Voyage MM-3.5 & 25.1 & 19.0 & 3.6 & -15.4 & 4.1 & 4.1 & 4.1 & 0.0 & 6.4 & 6.4 & 6.4 & 0.0 \\
\midrule
\multirow{12}{*}{\textbf{FIQ}} & E5-Omni & 6.7 & 5.9 & 1.9 & -4.0 & 2.7 & 2.7 & 2.7 & 0.0 & 5.5 & 5.5 & 5.5 & 0.0 \\
 & GME-Qwen2VL & 14.4 & 13.6 & 8.1 & -5.5 & 6.5 & 6.5 & 6.5 & 0.0 & 11.9 & 11.9 & 11.9 & 0.0 \\
 & LamRA & 15.3 & 14.7 & 10.5 & -4.2 & 7.7 & 7.7 & 7.7 & 0.0 & 15.2 & 15.2 & 15.2 & 0.0 \\
 & LamRA-Qwen2.5VL & 14.7 & 14.1 & 9.4 & -4.6 & 7.0 & 7.0 & 7.0 & 0.0 & 14.4 & 14.4 & 14.4 & 0.0 \\
 & MM-Embed & 11.5 & 10.6 & 6.3 & -4.3 & 4.9 & 4.9 & 4.9 & 0.0 & 9.6 & 9.6 & 9.6 & 0.0 \\
 & Qwen3-VL-2B & 13.2 & 12.4 & 7.2 & -5.2 & 5.1 & 5.1 & 5.1 & 0.0 & 11.7 & 11.7 & 11.7 & 0.0 \\
 & Qwen3-VL-8B & 15.3 & 14.5 & 8.1 & -6.4 & 6.4 & 6.4 & 6.4 & 0.0 & 13.9 & 13.9 & 13.9 & 0.0 \\
 & Rzen-Embed & 14.4 & 13.7 & 8.0 & -5.7 & 6.6 & 6.6 & 6.6 & 0.0 & 12.6 & 12.6 & 12.6 & 0.0 \\
 & VLM2Vec-V2 & 6.9 & 6.1 & 3.8 & -2.3 & 2.2 & 2.2 & 2.2 & 0.0 & 4.2 & 4.2 & 4.2 & 0.0 \\
 & Gemini Emb.\ 2 & 10.6 & 9.6 & 4.4 & -5.3 & 4.9 & 4.9 & 4.9 & 0.0 & 8.9 & 8.9 & 8.9 & 0.0 \\
 & Voyage MM-3.5 & 8.7 & 7.9 & 4.3 & -3.5 & 3.1 & 3.1 & 3.1 & 0.0 & 6.7 & 6.7 & 6.7 & 0.0 \\
\midrule
\multirow{12}{*}{\textbf{LaSCo}} & E5-Omni & 3.5 & 0.8 & -0.2 & -1.0 & 0.2 & 0.2 & 0.2 & 0.0 & 0.8 & 0.8 & 0.8 & 0.0 \\
 & GME-Qwen2VL & 5.1 & 2.5 & 0.0 & -2.4 & 0.7 & 0.7 & 0.7 & 0.0 & 4.4 & 4.4 & 4.4 & 0.0 \\
 & LamRA & 4.7 & 2.2 & 0.8 & -1.4 & 0.5 & 0.5 & 0.5 & 0.0 & 2.8 & 2.8 & 2.8 & 0.0 \\
 & LamRA-Qwen2.5VL & 4.9 & 2.2 & 0.5 & -1.8 & 0.5 & 0.5 & 0.5 & 0.0 & 2.6 & 2.6 & 2.6 & 0.0 \\
 & MM-Embed & 7.2 & 4.6 & 3.0 & -1.7 & 1.2 & 1.2 & 1.2 & 0.0 & 7.4 & 7.4 & 7.4 & 0.0 \\
 & Qwen3-VL-2B & 4.3 & 1.7 & -0.2 & -1.9 & 0.6 & 0.6 & 0.6 & 0.0 & 3.9 & 3.9 & 3.9 & 0.0 \\
 & Qwen3-VL-8B & 5.1 & 2.3 & -0.2 & -2.5 & 0.8 & 0.8 & 0.8 & 0.0 & 5.0 & 5.0 & 5.0 & 0.0 \\
 & Rzen-Embed & 4.9 & 2.3 & -1.1 & -3.3 & 0.6 & 0.6 & 0.6 & 0.0 & 4.0 & 4.0 & 4.0 & 0.0 \\
 & VLM2Vec-V2 & 3.4 & 0.7 & 0.9 & 0.2 & 0.5 & 0.5 & 0.5 & 0.0 & 2.8 & 2.8 & 2.8 & 0.0 \\
 & Gemini Emb.\ 2 & 4.8 & 2.3 & 0.7 & -1.6 & 0.8 & 0.8 & 0.8 & 0.0 & 4.0 & 4.0 & 4.0 & 0.0 \\
 & Voyage MM-3.5 & 5.6 & 3.0 & 0.7 & -2.3 & 0.7 & 0.7 & 0.7 & 0.0 & 3.9 & 3.9 & 3.9 & 0.0 \\
\midrule
\multirow{12}{*}{\textbf{CIRCO}} & E5-Omni & 27.2 & 18.6 & 18.9 & 0.4 & 16.2 & 16.2 & 16.2 & 0.0 & 15.7 & 15.7 & 15.7 & 0.0 \\
 & GME-Qwen2VL & 43.6 & 34.1 & 32.0 & -2.1 & 34.8 & 34.8 & 34.8 & 0.0 & 33.5 & 33.5 & 33.5 & 0.0 \\
 & LamRA & 41.3 & 32.2 & 33.2 & 1.0 & 33.3 & 33.3 & 33.3 & 0.0 & 32.0 & 32.0 & 32.0 & 0.0 \\
 & LamRA-Qwen2.5VL & 62.1 & 52.4 & 53.1 & 0.7 & 41.9 & 41.9 & 41.9 & 0.0 & 41.7 & 41.7 & 41.7 & 0.0 \\
 & MM-Embed & 40.2 & 31.1 & 29.4 & -1.7 & 28.2 & 28.2 & 28.2 & 0.0 & 30.1 & 30.1 & 30.1 & 0.0 \\
 & Qwen3-VL-2B & 37.9 & 30.2 & 28.2 & -2.0 & 28.6 & 28.6 & 28.6 & 0.0 & 29.2 & 29.2 & 29.2 & 0.0 \\
 & Qwen3-VL-8B & 44.0 & 33.8 & 35.0 & 1.3 & 33.2 & 33.2 & 33.2 & 0.0 & 34.4 & 34.4 & 34.4 & 0.0 \\
 & Rzen-Embed & 67.9 & 57.2 & 51.1 & -6.1 & 48.8 & 48.8 & 48.8 & 0.0 & 46.1 & 46.1 & 46.1 & 0.0 \\
 & VLM2Vec-V2 & 17.6 & 7.8 & 13.8 & 6.0 & 6.8 & 6.8 & 6.8 & 0.0 & 8.7 & 8.7 & 8.7 & 0.0 \\
 & Gemini Emb.\ 2 & 33.5 & 24.9 & 23.6 & -1.3 & 27.8 & 27.8 & 27.8 & 0.0 & 26.1 & 26.1 & 26.1 & 0.0 \\
 & Voyage MM-3.5 & 35.9 & 26.2 & 26.5 & 0.3 & 22.5 & 22.5 & 22.5 & 0.0 & 23.2 & 23.2 & 23.2 & 0.0 \\
\bottomrule
\end{tabular}
}
\end{table*}

\clearpage
\section{MRR-Based Composition Gap}
\label{app:compgap-mrr}
We also evaluate the same normalised composition gap used in the main paper with full-catalogue MRR:
\begin{equation}
\mathrm{CompGap}_{\mathrm{MRR}} = 1 - \frac{\max(I, T)}{\mathrm{MM}},
\end{equation}
where $\mathrm{MM}$, $I$, and $T$ now denote multimodal, image-only, and
text-only full-catalogue MRR. This retains the same interpretation as in the
main text while making the comparison more sensitive to top-rank changes.
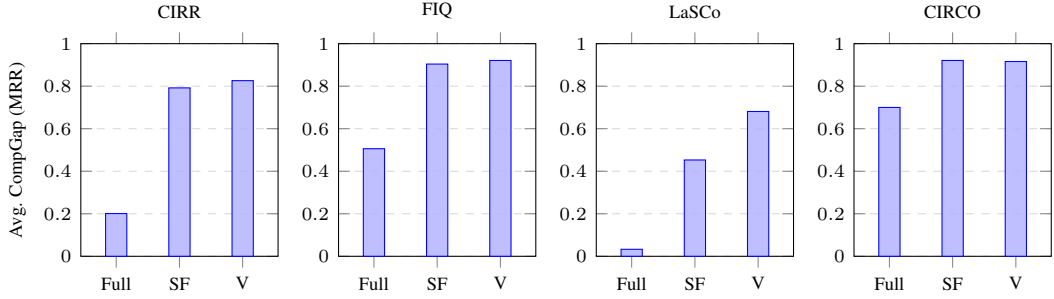
\begin{figure*}[t]
\centering
\begin{tikzpicture}
\begin{groupplot}[
  group style={group size=4 by 1, horizontal sep=0.8cm},
  width=0.3\textwidth,
  height=4.4cm,
  ymin=0.00,
  ymax=1.00,
  ybar,
  enlarge x limits=0.28,
  symbolic x coords={Full,SF,V},
  xtick=data,
  ymajorgrids,
  grid style={dashed,gray!30},
  tick label style={font=\scriptsize},
  title style={font=\scriptsize},
  ylabel style={font=\scriptsize},
  xlabel style={font=\scriptsize},
  every axis plot/.append style={draw=orange!80!black, fill=orange!65, fill opacity=0.84, /tikz/bar width=8pt},
]
\nextgroupplot[title={CIRR}, ylabel={Avg. CompGap (MRR)}, xlabel={}]
\addplot coordinates {(Full,0.201) (SF,0.792) (V,0.826)};
\nextgroupplot[title={FIQ}, ylabel={}, xlabel={}]
\addplot coordinates {(Full,0.506) (SF,0.904) (V,0.921)};
\nextgroupplot[title={LaSCo}, ylabel={}, xlabel={}]
\addplot coordinates {(Full,0.033) (SF,0.453) (V,0.681)};
\nextgroupplot[title={CIRCO}, ylabel={}, xlabel={}]
\addplot coordinates {(Full,0.700) (SF,0.921) (V,0.916)};
\end{groupplot}
\end{tikzpicture}
\caption{Retriever-averaged normalised composition gap based on full-catalogue MRR. Each panel corresponds to one dataset; the x-axis shows the original benchmark (\textbf{Full}), the \circussf (\textbf{SF}) split, and the \circusv (\textbf{V}) split.}
\label{fig:avg_compgap_mrr}
\end{figure*}
\Cref{fig:avg_compgap_mrr} supports the same conclusion as the main-paper
nDCG analysis, but now under a top-rank-sensitive metric. On CIRR, FashionIQ, and
LaSCo, the retriever-averaged MRR-based composition gap increases sharply from the
original benchmark to the filtered subsets, indicating that once shortcut-solvable
and invalid instances are removed, the earliest correct hit depends much more
strongly on combining image and text than on either modality alone. This is
especially clear for LaSCo, where the gap is nearly absent on \textbf{Full}
(0.033) but becomes substantial on \textbf{SF} (0.453) and \textbf{V} (0.681),
which is consistent with the main claim that raw benchmark success on this dataset
is heavily confounded by shortcut and validity artifacts. CIRCO again behaves
differently: its original benchmark already exhibits a high MRR-based gap (0.700),
and filtering only raises it modestly, which matches the broader picture that
CIRCO contains a comparatively clean multimodal core from the start. Overall, these results are consistent with the trends observed under nDCG, confirming that \circus{} requires multimodal composition.

\clearpage
\section{Retrieval Uncertainty Analysis}
\label{app:uncertainty}

The main paper reports point estimates for filtered-set retrieval scores and multimodal--unimodal deltas. Since these quantities are computed on finite query samples, they carry sampling uncertainty. This appendix reports 95\% confidence intervals for the retrieval analysis. We use \textbf{SF} for \circussf{} and \textbf{V} for \circusv{}. All entries are reported as \emph{estimate [95\% lower bound, 95\% upper bound]}.

\Cref{tab:uncertainty_retrieval} reports uncertainty estimates for the filtered-set retrieval analysis. The reported values are dataset-level averages of non-linear ranking metrics and paired multimodal--unimodal deltas, for which no simple closed-form interval applies. We therefore use a non-parametric query bootstrap~\citep{efron1993bootstrap}: for each (dataset, split) we resample queries with replacement, recompute the metric on the bootstrap sample, and report the 2.5\textsuperscript{th} and 97.5\textsuperscript{th} percentiles of the resulting distribution. All intervals are computed from the stored per-query ranks of the eleven retrievers used in the main paper.

For multimodal Recall@10 and multimodal nDCG, we form a per-query score for each retriever, average across retrievers, and then average over the bootstrap query sample. For the paired deltas $\Delta$MM-T and $\Delta$MM-I, we form the multimodal--unimodal difference at the per-query, per-retriever level before averaging, so that the interval reflects within-query paired variation rather than the sum of two independent uncertainties. $\Delta R@10$ vs.\ \textbf{Full} is bootstrapped analogously as a between-split difference.

The main question for \Cref{tab:uncertainty_retrieval} is whether filtering changes the evaluation in a stable direction. A $\Delta R@10$ interval that lies entirely below zero indicates that the filtered split is reliably harder than the original benchmark. A $\Delta$MM-T or $\Delta$MM-I interval that lies entirely above zero indicates that the multimodal query retains a real ranking advantage over the corresponding unimodal variant. Intervals that cross zero indicate that the sign of the effect is not resolved at this sample size.

{\color{blue}
\begin{table*}[t]
\centering
\caption{Bootstrap uncertainty for filtered-set retrieval analysis, computed from stored per-query ranks over the eleven retrievers used in the main paper. $R@10$ and MM nDCG use bootstrap intervals over provider-averaged per-query scores. MM--Text and MM--Image use paired bootstrap intervals over provider-averaged per-query nDCG differences between multimodal and unimodal queries. $\Delta R@10$ vs Full is the bootstrap interval for the difference in provider-averaged multimodal recall between the filtered split and the original benchmark.}
\label{tab:uncertainty_retrieval}
\scriptsize
\setlength{\tabcolsep}{4pt}
\resizebox{\textwidth}{!}{
\begin{tabular}{l l c c c c c c}
\toprule
\textbf{Dataset} & \textbf{Split} & \textbf{$N$} & \textbf{MM $R@10$} & \textbf{$\Delta R@10$ vs Full} & \textbf{MM nDCG} & \textbf{MM--Text nDCG} & \textbf{MM--Image nDCG} \\
\midrule
\multirow{3}{*}{\textbf{CIRR}} & Full & 4170 & 59.9 [58.7, 61.0] & -- & 36.7 [36.3, 37.2] & 5.2 [4.8, 5.6] & 18.5 [18.0, 19.0] \\
 & SF & 685 & 14.4 [12.6, 16.5] & -45.5 [-47.7, -43.4] & 19.8 [19.2, 20.5] & 6.3 [5.7, 7.0] & 7.0 [6.4, 7.5] \\
 & V & 303 & 19.8 [16.4, 23.1] & -40.1 [-43.6, -36.6] & 21.9 [20.8, 23.2] & 8.2 [7.1, 9.4] & 8.3 [7.3, 9.4] \\
\midrule
\multirow{3}{*}{\textbf{FashionIQ}} & Full & 6003 & 25.4 [24.6, 26.2] & -- & 22.6 [22.2, 22.9] & 6.9 [6.6, 7.2] & 12.4 [12.1, 12.7] \\
 & SF & 4069 & 12.0 [11.4, 12.7] & -13.3 [-14.4, -12.2] & 16.9 [16.7, 17.2] & 6.5 [6.3, 6.8] & 7.7 [7.4, 7.9] \\
 & V & 586 & 23.6 [21.2, 26.0] & -1.7 [-4.1, 0.7] & 22.0 [21.1, 22.8] & 10.7 [9.9, 11.4] & 12.0 [11.2, 12.8] \\
\midrule
\multirow{3}{*}{\textbf{LaSCo}} & Full & 30031 & 11.9 [11.7, 12.2] & -- & 16.9 [16.8, 16.9] & 1.3 [1.2, 1.4] & 2.3 [2.3, 2.4] \\
 & SF & 18418 & 1.7 [1.6, 1.8] & -10.2 [-10.5, -10.0] & 12.6 [12.5, 12.6] & 1.7 [1.6, 1.7] & 1.0 [1.0, 1.0] \\
 & V & 758 & 10.3 [9.4, 11.4] & -1.6 [-2.7, -0.5] & 17.0 [16.6, 17.4] & 4.9 [4.5, 5.3] & 3.6 [3.3, 4.0] \\
\midrule
\multirow{3}{*}{\textbf{CIRCO}} & Full & 220 & 75.2 [71.7, 78.8] & -- & 42.1 [40.6, 43.5] & 22.0 [20.2, 23.8] & 20.4 [18.8, 22.1] \\
 & SF & 56 & 58.1 [49.8, 66.4] & -17.1 [-25.9, -8.9] & 35.2 [32.2, 38.2] & 23.2 [20.1, 26.0] & 20.6 [17.8, 23.5] \\
 & V & 42 & 57.6 [47.6, 67.3] & -17.7 [-28.1, -7.6] & 35.0 [31.1, 38.5] & 23.2 [19.3, 26.9] & 20.2 [16.6, 23.8] \\
\bottomrule
\end{tabular}
}
\end{table*}
}

On CIRR, LaSCo, and CIRCO, the multimodal Recall@10 drop from \textbf{Full} to both \textbf{SF} and \textbf{V} lies clearly below zero. Thus, raw benchmark scores are inflated by shortcut-solvable or ill-posed queries to a degree that is not explained by query sampling. At the same time, the paired deltas $\Delta$MM-T and $\Delta$MM-I remain positive on the filtered splits, so the multimodal query retains a measurable ranking advantage over both unimodal variants once shortcuts and ill-posed queries are removed.

FashionIQ is informative as an edge case: its validated Recall@10 (23.6~[21.2,~26.0]) is statistically indistinguishable from the full-benchmark score (25.4~[24.6,~26.2]), with $\Delta R@10$ vs.\ \textbf{Full} of $-1.7~[-4.1,~0.7]$ crossing zero. However, the text--multimodal advantage sharpens on the validated subset, with $\Delta$MM-T rising from 6.9~[6.6,~7.2] on \textbf{Full} to 10.7~[9.9,~11.4] on \textbf{V}, while the image--multimodal advantage is essentially preserved. We read this as supportive of the validation procedure rather than as a counterexample to it: validation removes noisy items without making the benchmark uniformly harder, and the modality-composition signal is preserved on $\Delta$MM-I and sharpened on $\Delta$MM-T.

Overall, the retrieval uncertainty estimates support the main conclusions: the filtered and validated splits preserve a positive multimodal advantage over unimodal ablations, while substantially reducing inflated full-benchmark performance.

\clearpage
\section{Inter-Annotator Agreement}
\label{app:agreement}

To assess the stability of the human validation protocol, we ran a separate agreement study on a 50-query overlap subset drawn from the shortcut-free set. We evaluate agreement primarily on the binary \valid/invalid decision, because this is the quantity used to define the validated split. Since all 9 annotators labelled the same 50 queries, we report Fleiss' $\kappa$ across the 9 raters, nominal Krippendorff's $\alpha$, and mean pairwise Cohen's $\kappa$
(Table~\ref{tab:inter_annotator_agreement}). Final labels for the main audit are obtained by aggregating annotator judgments according to the decision rule described in Section~\ref{sec:human-val}; the overlap study is used only to quantify reliability and does not alter the annotation protocol.

\begin{table}[t]
\centering
\small
\caption{Inter-annotator agreement on the 50-query overlap study. Agreement is computed primarily on the binary \valid{}/invalid decision, which defines the validated split. Results are reported over all 9 available annotators.}
\label{tab:inter_annotator_agreement}
\begin{tabular}{l c}
\toprule
Metric & Value \\
\midrule
Overlap queries & 50 \\
Annotators & 9 \\
Fleiss' $\kappa$ on \valid{}/invalid & 0.456 \\
Krippendorff's $\alpha$ on \valid{}/invalid & 0.457 \\
Mean pairwise Cohen's $\kappa$ & 0.458 \\
Pairwise Cohen range & 0.280--0.756 \\
Unanimous \valid{}/invalid rate & 44.0\% \\
\broadquery Fleiss' $\kappa$ & 0.447 \\
\broadquery Krippendorff's $\alpha$ & 0.448 \\
\bottomrule
\end{tabular}
\end{table}

\begin{figure*}[t]
\centering
\includegraphics[width=\textwidth]{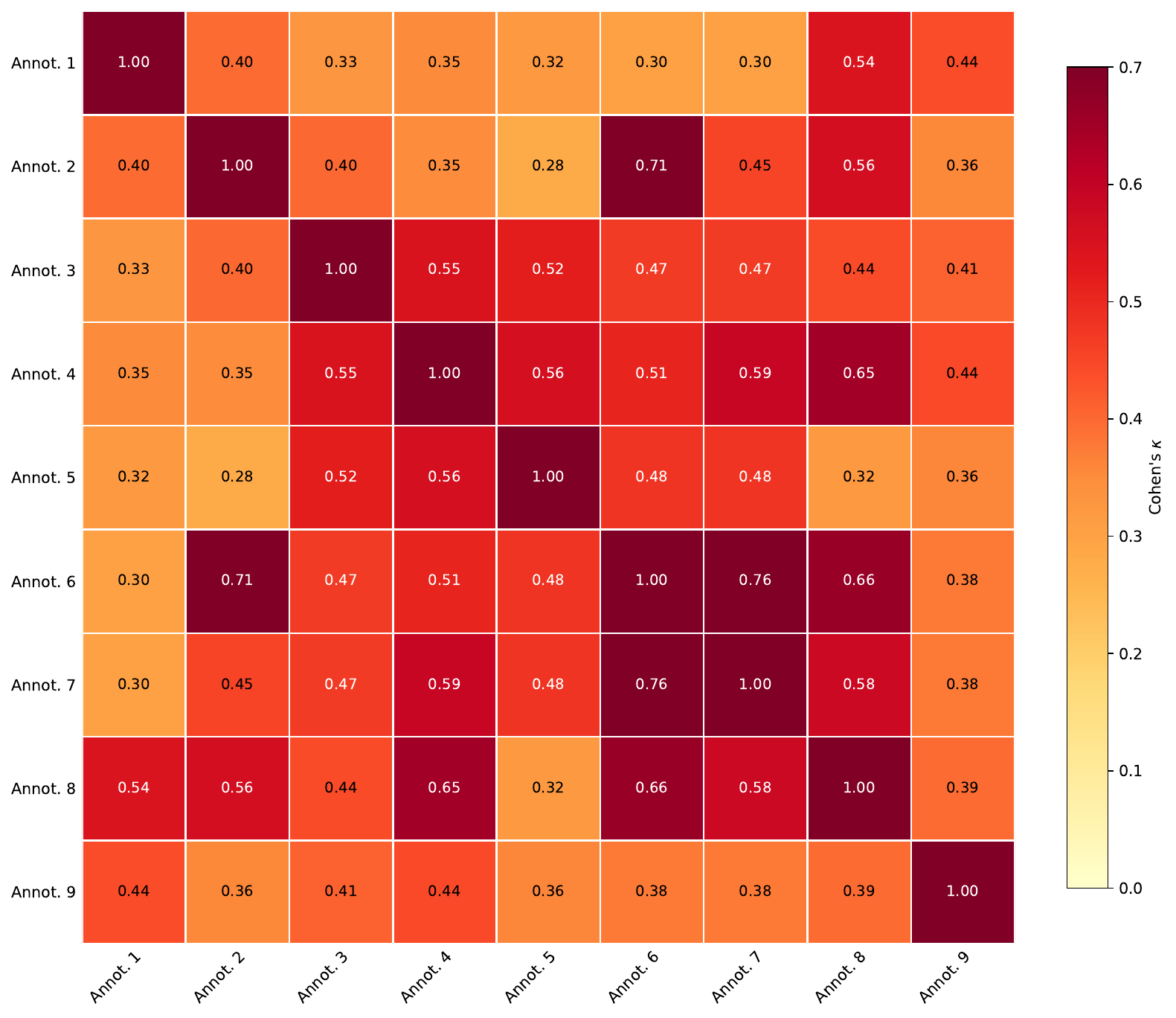}
\caption{Pairwise Cohen's $\kappa$ values on the binary \valid{}/invalid decision across the 9-annotator overlap subset, shown with anonymized annotator identities.}
\label{fig:agreement_kappa_heatmap}
\end{figure*}

\begin{figure*}[t]
\centering
\includegraphics[width=\textwidth]{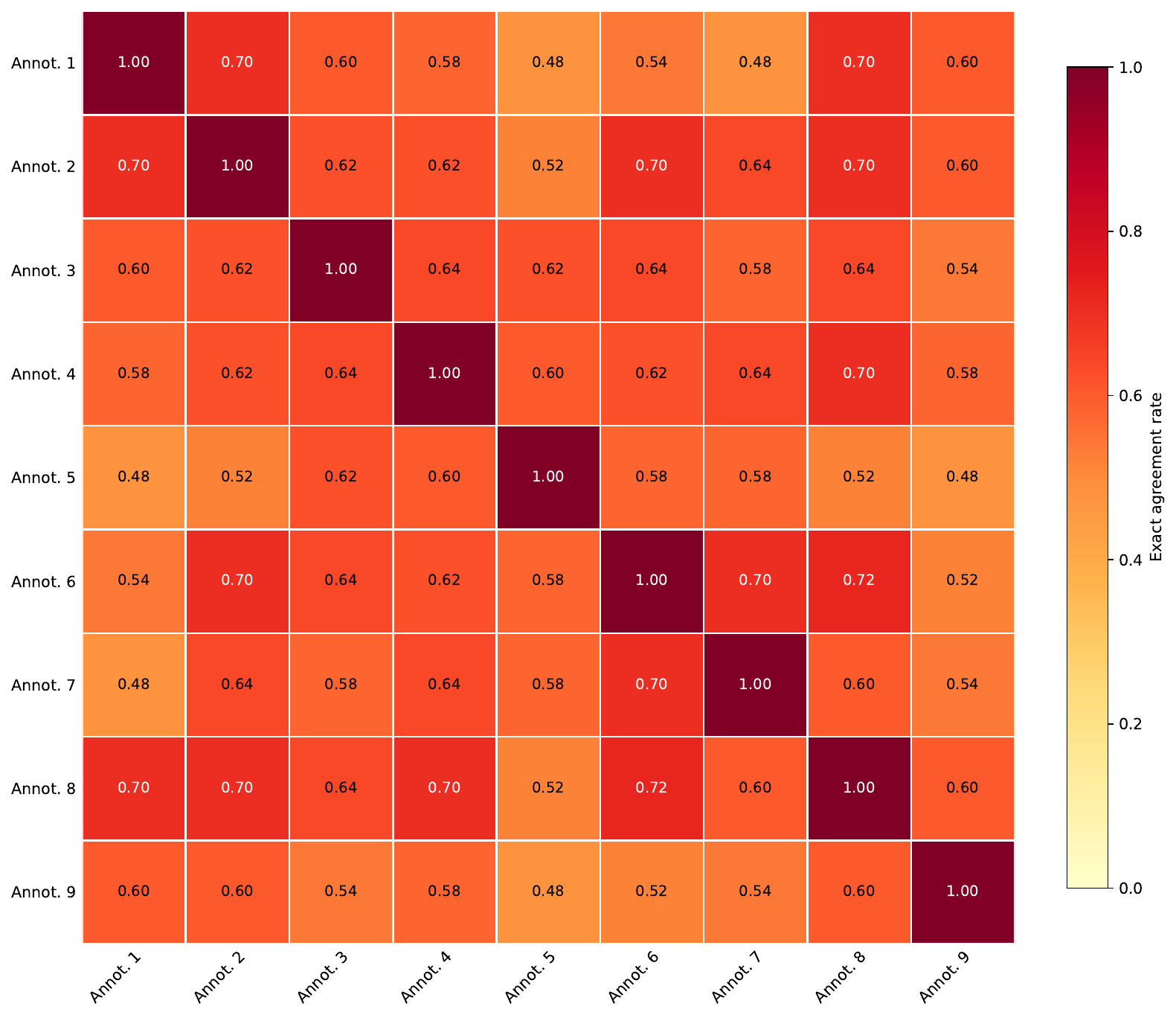}
\caption{Pairwise exact agreement rates on the full issue signature across the same 9-annotator overlap subset. This view is stricter than the binary \valid{}/invalid decision because it requires agreement on the specific failure label(s) as well.}
\label{fig:agreement_signature_heatmap}
\end{figure*}

The resulting agreement is moderate rather than near-perfect, in line with the complexity of the task and the limited sample size. Fleiss' $\kappa$ on the binary \valid{}/invalid decision is 0.456, with mean pairwise Cohen's $\kappa$ 0.458 and nominal Krippendorff's $\alpha$ 0.457. The close match between $\kappa$- and $\alpha$-based reliability is reassuring here: the conclusion does not depend on which standard multi-rater nominal agreement coefficient we use. Agreement becomes stricter when we require an exact match on the full issue signature rather than only on the binary validity decision; this is expected because annotators may agree that a query is invalid while differing on which specific failure label is most salient.

Among invalid-reason flags, \broadquery is the most consistently assigned category (Fleiss' $\kappa$ 0.447; Krippendorff's $\alpha$ 0.448), which matches the main paper's conclusion that under-specification is the dominant residual benchmark pathology. By contrast, the rarer categories such as \invalidtext and \invalidref are more prevalence-sensitive and substantially less stable in isolation, so we do not over-interpret their category-level agreement values. \Cref{fig:agreement_kappa_heatmap,fig:agreement_signature_heatmap} make the same point visually: there is no single outlier annotator pair dominating either view, but rather a broad band of moderate binary agreement with consistently lower exact taxonomy agreement.

Overall, the agreement results show that binary validity decisions are reasonably stable, while disagreement concentrates on borderline cases, reinforcing that the shortcut-free subset contains intrinsically ambiguous queries that require careful filtering.

\clearpage
\section{Shortcut Audit Robustness}
\label{app:shortcut-robustness}

\Cref{tab:shortcut_robustness} evaluates the sensitivity of the shortcut audit to the retrieval cutoff and the composition of the retriever pool. A query is classified as a shortcut at cutoff $K$ if, for at least one retriever, either its text-only or image-only rank is at most $K$. Increasing $K$ therefore relaxes the shortcut criterion, leading to a monotonic increase in shortcut rates and a corresponding decrease in the shortcut-free residue.

Despite this, the qualitative ordering of datasets is stable across cutoffs: CIRR and CIRCO remain the most shortcut-heavy, while FashionIQ remains the least. To assess sensitivity to the retriever pool, we perform a leave-one-retriever-out (LOO) analysis at $K=10$, recomputing aggregate shortcut rates while removing each retriever in turn under the same best-rank aggregation used in the main audit.

The resulting LOO ranges are narrow across all datasets (at most $\sim$1--2 percentage points), indicating that shortcut rates are not driven by any single retriever. Overall, these results show that shortcut prevalence is robust to both cutoff choice and retriever composition, and reflects a stable property of the benchmarks rather than an artifact of specific evaluation settings.

\begin{table}[t]
\centering
\caption{Robustness of aggregate shortcut rates to retrieval cutoff and retriever pool. Rates are percentages of queries classified as shortcuts. $K=10$ matches the main shortcut audit in \cref{tab:aggregated}. LOO denotes leave-one-retriever-out: we remove one retriever at a time, recompute the aggregate shortcut rate, and report the resulting min--max range.}
\label{tab:shortcut_robustness}
\small
\setlength{\tabcolsep}{6pt}
\begin{tabular}{lcccc}
\toprule
\textbf{Dataset} & \textbf{K=5} & \textbf{K=10} & \textbf{K=20} & \textbf{LOO range at K=10} \\
\midrule
CIRR & 75.4 & 83.6 & 89.7 & 82.7--83.2 \\
FashionIQ & 25.1 & 32.2 & 41.0 & 30.6--31.8 \\
LaSCo & 29.0 & 38.7 & 49.4 & 37.3--38.1 \\
CIRCO & 62.3 & 74.5 & 85.9 & 72.7--74.5 \\
\bottomrule
\end{tabular}
\end{table}

\clearpage
\section{Audit Error Distribution}
\label{app:audit-error-distribution}

This appendix breaks down the human-validation issues from \cref{sec:human-val} by dataset and by shortcut-free bucket. \Cref{tab:error_distribution_comp_nec} reports the issue distribution for invalid \emph{composition-required} queries, while \Cref{tab:error_distribution_unresolved} reports the same statistics for invalid \emph{unresolved} queries. Percentages are computed with respect to invalid audited items within each dataset and bucket, and issue labels are non-exclusive.

\begin{table*}[t]
\centering
\caption{Distribution of annotation issues among invalid \emph{composition-required} queries.
Entries are count (\% of invalid audited items within the corresponding dataset and bucket).
Issue labels are non-exclusive, so percentages do not sum to 100.
Across all datasets, three invalid composition-required items received no explicit issue label.}
\label{tab:error_distribution_comp_nec}
\small
\setlength{\tabcolsep}{5pt}
\begin{tabular}{l r r r r r}
\toprule
\textbf{Dataset} & \textbf{Invalid}
& \textbf{\makecell{\textsc{Invalid} \\ \textsc{text}}}
& \textbf{\makecell{\textsc{Invalid} \\ \textsc{reference image}}}
& \textbf{\makecell{\textsc{Invalid} \\ \textsc{target image}}}
& \textbf{\makecell{\textsc{Overly} \\ \textsc{broad query}}} \\
\midrule
CIRR       & 124 & 12 (9.7)  & 7 (5.6)   & 11 (8.9)  & 98 (79.0)  \\
FashionIQ  & 632 & 46 (7.3)  & 1 (0.2)   & 46 (7.3)  & 550 (87.0) \\
LaSCo      & 548 & 71 (13.0) & 12 (2.2)  & 94 (17.2) & 393 (71.7) \\
CIRCO      & 14  & 1 (7.1)   & 0 (0.0)   & 1 (7.1)   & 12 (85.7)  \\
\bottomrule
\end{tabular}
\end{table*}

\begin{table*}[t]
\centering
\caption{Distribution of annotation issues among invalid \emph{unresolved} queries.
Entries are count (\% of invalid audited items within the corresponding dataset and bucket).
Issue labels are non-exclusive, so percentages do not sum to 100.
Across all datasets, eight invalid unresolved items received no explicit issue label.}
\label{tab:error_distribution_unresolved}
\small
\setlength{\tabcolsep}{5pt}
\begin{tabular}{l r r r r r}
\toprule
\textbf{Dataset} & \textbf{Invalid}
& \textbf{\makecell{\textsc{Invalid} \\ \textsc{text}}}
& \textbf{\makecell{\textsc{Invalid} \\ \textsc{reference image}}}
& \textbf{\makecell{\textsc{Invalid} \\ \textsc{target image}}}
& \textbf{\makecell{\textsc{Overly} \\ \textsc{broad query}}} \\
\midrule
CIRR       & 258 & 24 (9.3)  & 14 (5.4)  & 82 (31.8)  & 155 (60.1) \\
FashionIQ  & 782 & 55 (7.0)  & 6 (0.8)   & 172 (22.0) & 572 (73.1) \\
LaSCo      & 694 & 109 (15.7) & 29 (4.2)  & 205 (29.5) & 413 (59.5) \\
CIRCO      & 0   & --        & --        & --         & --         \\
\bottomrule
\end{tabular}
\end{table*}

These results provide further evidence that the shortcut-free residue is often driven by under-specification rather than genuine compositional difficulty. Across datasets and buckets, \broadquery is the dominant issue category, accounting for the majority of invalid queries in all settings, which indicates that many queries lack sufficient specificity to define a well-posed retrieval task.

A clearer distinction emerges between the two buckets. In the \emph{composition-required} subset, invalid queries are overwhelmingly dominated by \broadquery (e.g., 71--87\% across datasets), suggesting that many examples classified as composition-requiring are in fact ill-posed rather than intrinsically complex. In contrast, the \emph{unresolved} subset shows a relatively higher prevalence of \invalidtarget cases (e.g., up to 30\%), indicating that a non-trivial fraction of unresolved queries are difficult because of annotation errors in the benchmark instance itself.

Taken together, these results clarify the role of the validated split. While shortcut filtering removes queries that admit unimodal solutions, it does not eliminate under-specified or corrupted examples. The additional human validation step is therefore necessary to isolate queries that are both shortcut-free and well-posed, enabling a more reliable interpretation of model behaviour.

\clearpage
\section{Additional Implementation and Evaluation Details}
\label{app:checklist-details}
This appendix provides additional details on broader impacts, dataset usage and licensing, and the human validation study.

\subsection{Broader impacts}

Our work aims to improve the reliability of composed image retrieval evaluation by identifying and isolating benchmark artifacts such as unimodal shortcuts and under-specified queries. A positive outcome is that future models can be evaluated on cleaner and more diagnostic subsets, reducing inflated claims of multimodal reasoning that in practice rely on spurious cues or ill-posed instances. The release of filtered splits and audit metadata also improves transparency by making it easier to disentangle benchmark quality from model capability.

To ensure transparency and reproducibility, we retain full provenance to the original benchmarks, report results on both the original and filtered subsets, and release only derived artifacts (e.g., audit metadata and filtered query identifiers) rather than new standalone datasets. This enables others to reproduce our analysis, inspect validation decisions, and separate benchmark properties from model performance.

\subsection{Licenses and redistribution policy}

This work builds on four publicly available CIR benchmarks: CIRR, FashionIQ, CIRCO, and LaSCo. These datasets are distributed under a mix of licenses and usage terms, including MIT-style licenses for annotations (CIRR, LaSCo), the
Community Data License Agreement (FashionIQ), and CC BY-NC 4.0 for CIRCO, with underlying images sourced from datasets such as NLVR2 and COCO that remain subject to their respective terms.

Our work does not modify or redistribute the original datasets. Instead, we release only derived artifacts, including audit metadata, filtered query identifiers, and evaluation code. We do not re-host or distribute any third-party image data, and users are expected to obtain the underlying datasets through the
official sources and comply with their respective licenses.

\subsection{Additional Human Validation Details}
\label{app:human-validation-materials}

The human validation study was conducted by nine annotators following a shared protocol. Annotators were co-authors and collaborating researchers familiar with composed image retrieval, and were assigned fixed batches of queries.

\begin{figure}[t]
\centering
\includegraphics[width=\linewidth]{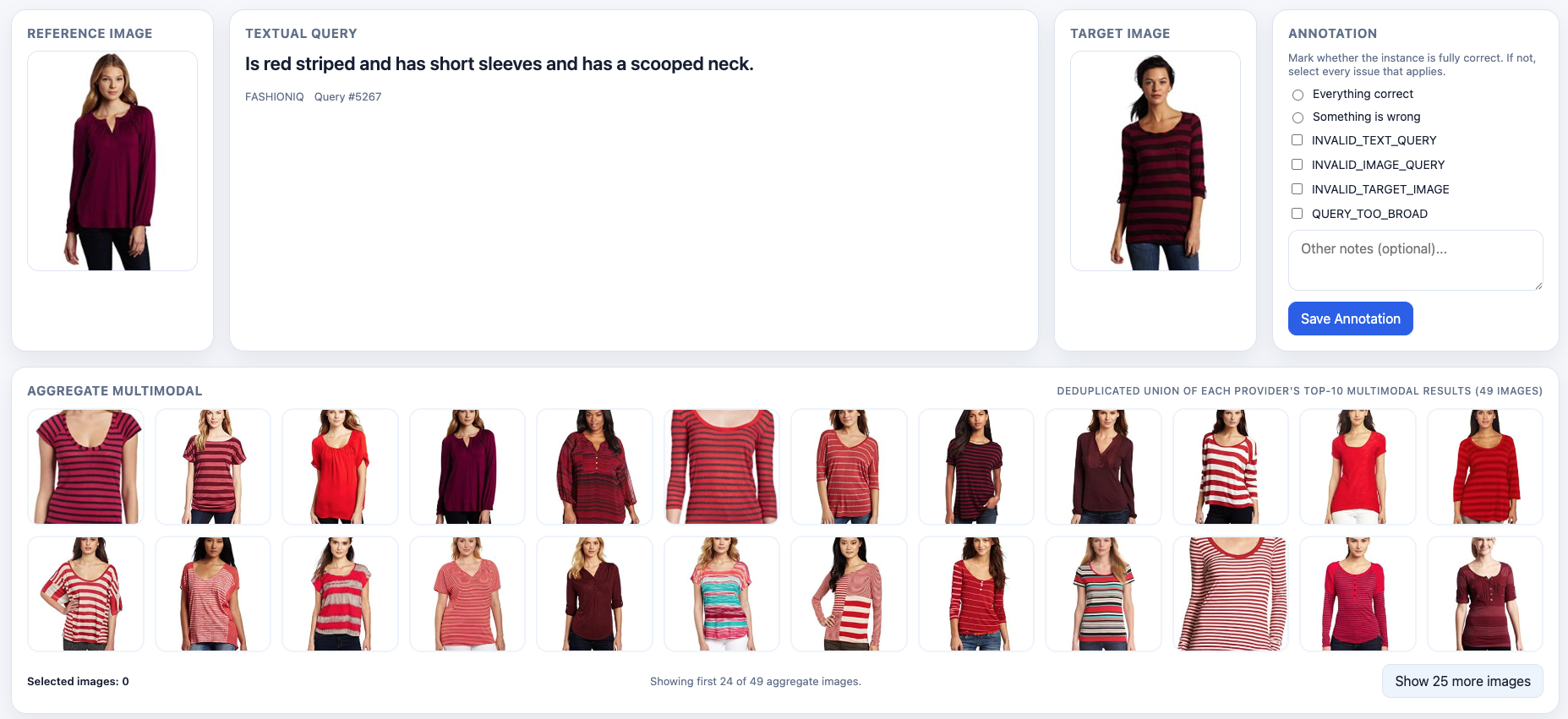}
\caption{Annotation interface used in the human validation study.}
\label{fig:annotation-interface}
\end{figure}

The annotation interface (Figure~\ref{fig:annotation-interface}) displays the reference image, textual edit, target image, and an aggregate multimodal retrieval panel. The panel consists of the deduplicated union of top-$K$ multimodal retrieval results across all retrievers and is provided as a diagnostic aid for assessing query specificity.

To make the protocol concrete, the appendix includes a small representative gallery with one retained example and one example for each issue label, using the same nomenclature as \Cref{tab:validity_audit}. The main-text figure \Cref{fig:validity_issue_examples} focuses on failure modes only, while the appendix gallery also includes a retained \valid{} example.

Annotators followed a structured decision process. They first evaluated the textual edit, then the reference image, then the target image, and finally the specificity of the composed query. Multiple issue labels could be assigned when
appropriate.

The retrieval panel was used only to support judgments about query specificity, e.g., by revealing whether multiple plausible matches exist in the dataset. Annotators were instructed to base their decisions on the coherence of the reference--edit--target triplet, rather than on whether a model succeeds or fails on the example.

The study operated exclusively on publicly available benchmark data and involved only the annotation of dataset quality. No personal or sensitive data was collected.

\subsection{Compute resources}

All experiments were run on a single local machine with one NVIDIA H100 NVL GPU, an AMD EPYC Genoa CPU, and 256\,GB of RAM. The code was implemented using PyTorch as the main framework, with HuggingFace Transformers and vLLM for retriever inference. Retrievers required approximately 16 to 86\,GiB of GPU memory at load time.

The full-benchmark retrieval experiments were the most computationally intensive stage. Across all runs, total wall-clock time was approximately 239.3 hours, with about 161.8 hours corresponding to local (non-API) runs on a single GPU. These measurements provide an approximate upper bound on compute cost, as they include startup overheads and exclude additional exploratory runs.

\clearpage
\subsection{Representative Annotation Gallery}
\label{app:annotation-gallery}

This gallery shows one retained \valid{} query and one representative query for each issue label. Each example uses the same reference--edit--target presentation inspected by annotators during the validation study.

\newcommand{\annotationpanel}[7]{
\begingroup
\setlength{\fboxsep}{7pt}
\fcolorbox[RGB]{214,208,198}{250,248,244}{
\begin{minipage}[t]{#1}
\centering
{\small\textbf{#2}}\\[4pt]
\begin{minipage}[t]{0.485\linewidth}
\centering
{\scriptsize\textbf{Ref}}\\[3pt]
\includegraphics[width=\linewidth,height=#7,keepaspectratio]{#3}
\end{minipage}
\hfill
\begin{minipage}[t]{0.485\linewidth}
\centering
{\scriptsize\textbf{Tgt}}\\[3pt]
\includegraphics[width=\linewidth,height=#7,keepaspectratio]{#4}
\end{minipage}\\[6pt]
{\footnotesize\textbf{Query:} ``#5''\par}
\vspace{3pt}
{\footnotesize\textbf{Audit note:} #6\par}
\end{minipage}}
\endgroup}

\begin{figure*}[p]
\centering
\annotationpanel{0.9\textwidth}{\valid{} -- FashionIQ q5382}{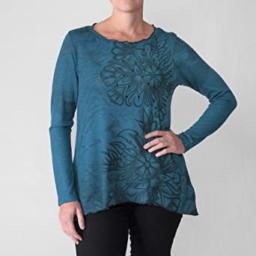}{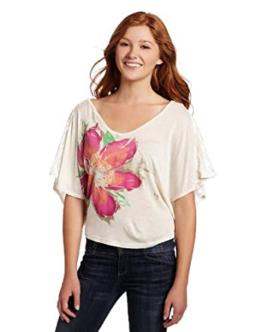}{Is white with shorter sleeves and a flower and is shorter sleeved.}{Retained in the validated split; the multimodal query succeeds while both unimodal variants fail.}{0.19\textheight}

\vspace{8pt}

\annotationpanel{0.9\textwidth}{\invalidtext{} -- FashionIQ q841}{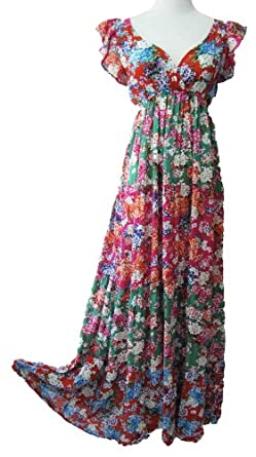}{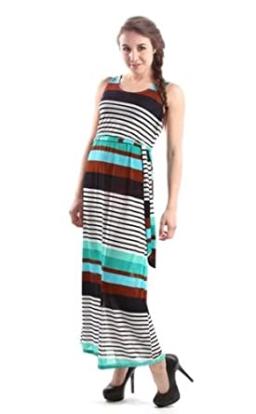}{Has stripes and is without sleeves and is plain black.}{``has stripes'' and ``is plain black'' are contradictory.}{0.19\textheight}

\vspace{8pt}

\annotationpanel{0.9\textwidth}{\invalidref{} -- FashionIQ q573}{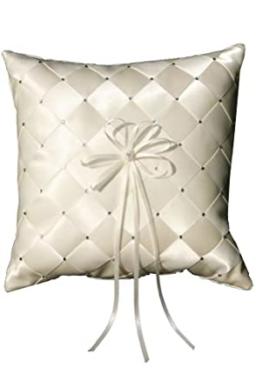}{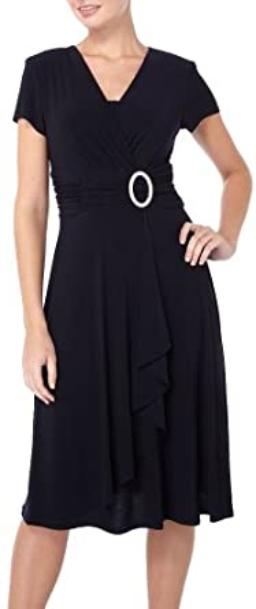}{Is darker and has longer sleeves and is solid black.}{The reference image is ``a pillow, not a dress.''}{0.19\textheight}
\caption{Representative FashionIQ annotation outcomes. Each page now shows three stacked panels using the same reference--edit--target triplet inspected in the annotation platform, with the audit text embedded directly in the figure.}
\label{fig:annotation-gallery-fashioniq-a}
\end{figure*}

\begin{figure*}[p]
\centering
\annotationpanel{0.9\textwidth}{\invalidtarget{} -- FashionIQ q378}{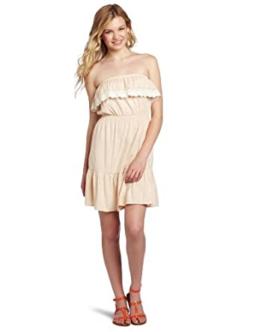}{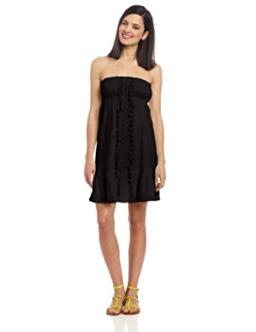}{Is more black and simple and darker and has green.}{``The target has no green.''}{0.19\textheight}

\vspace{8pt}

\annotationpanel{0.9\textwidth}{\broadquery{} -- FashionIQ q568}{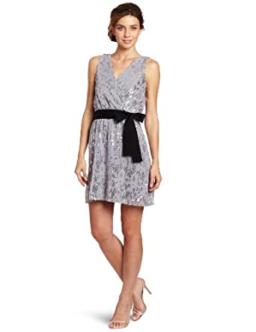}{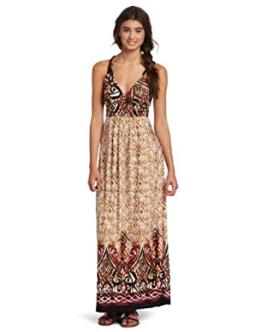}{Is longer and darker and is longer and more patterned.}{``I could find many acceptable candidates for the query.''}{0.19\textheight}

\vspace{8pt}

\annotationpanel{0.9\textwidth}{\valid{} -- CIRR q1231}{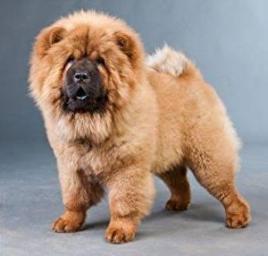}{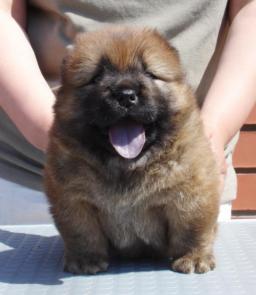}{Person in gray shirt stands behind dog.}{Retained in the validated split; the query remains composition-required under the aggregate shortcut audit.}{0.19\textheight}
\caption{Additional FashionIQ and CIRR examples, including the remaining FashionIQ \broadquery{} case and a retained CIRR example.}
\label{fig:annotation-gallery-fashioniq-b}
\end{figure*}

\begin{figure*}[p]
\centering
\annotationpanel{0.9\textwidth}{\invalidtarget{} -- CIRR q2820}{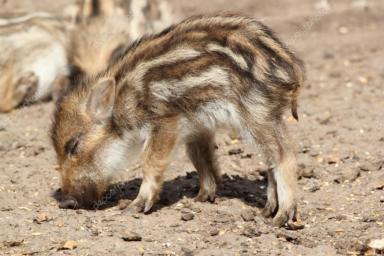}{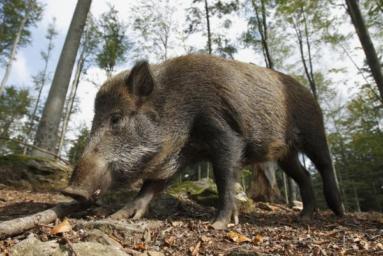}{Pig darker and seen from below.}{The target looks like a mature pig rather than a darker piglet viewed from below.}{0.19\textheight}

\vspace{8pt}

\annotationpanel{0.9\textwidth}{\broadquery{} -- CIRR q2800}{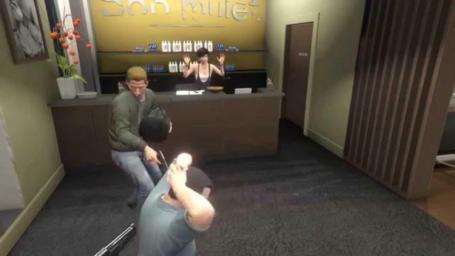}{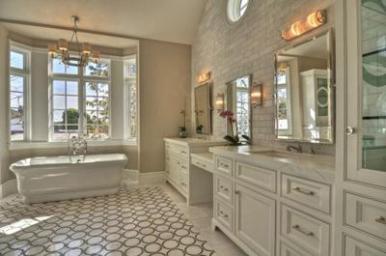}{Remove the people and switch to a patterned floor.}{``So many patterned floors without people in this dataset.''}{0.19\textheight}

\vspace{8pt}

\annotationpanel{0.9\textwidth}{\invalidtext{} -- CIRR q3127}{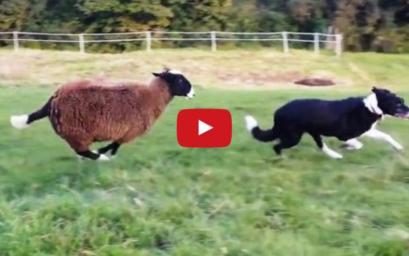}{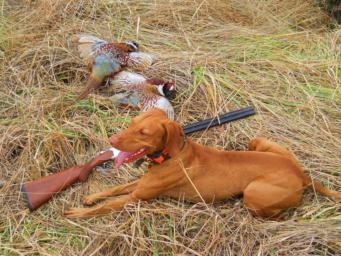}{A dog laying down on the ground with a gum on its side.}{Annotator note: ``gum'' should be ``gun.''}{0.19\textheight}
\caption{Representative CIRR outcomes showing non-unique targets, broad queries, and ambiguous text.}
\label{fig:annotation-gallery-cirr}
\end{figure*}

\begin{figure*}[p]
\centering
\annotationpanel{0.9\textwidth}{\valid{} -- LaSCo q573}{}{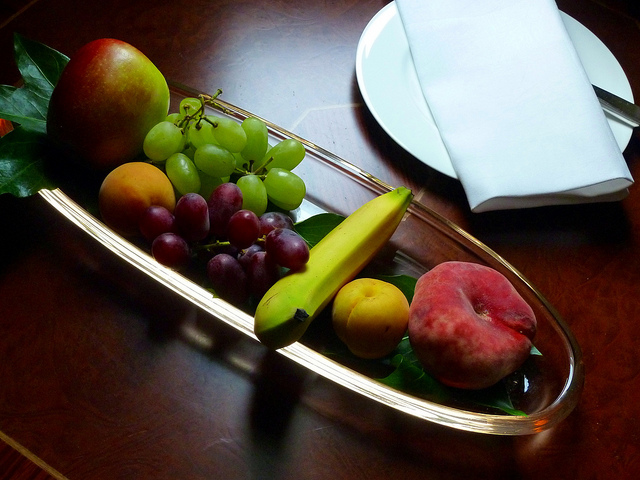}{change the number of different fruits to 7}{Retained in the validated split; multimodal retrieval succeeds while both unimodal conditions fail under the aggregate audit.}{0.19\textheight}

\vspace{8pt}

\annotationpanel{0.9\textwidth}{\invalidtext{} -- LaSCo q3190}{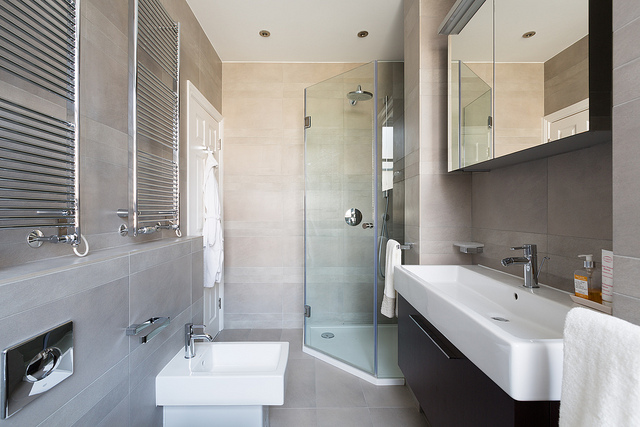}{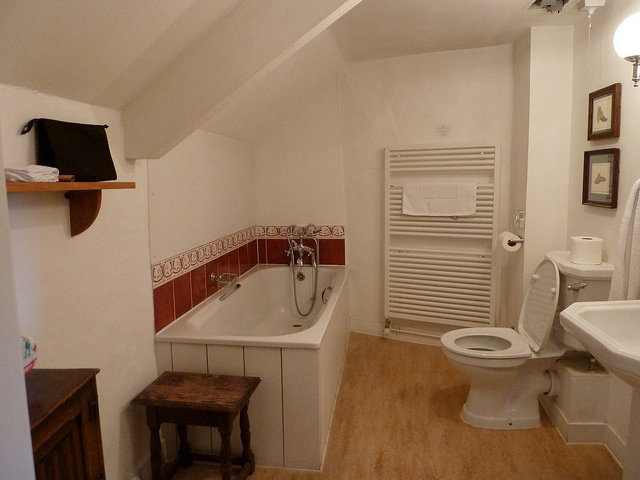}{Do we still have kleenex?}{``I can't understand the retrieval instruction here.''}{0.19\textheight}

\vspace{8pt}

\annotationpanel{0.9\textwidth}{\broadquery{} -- LaSCo q8908}{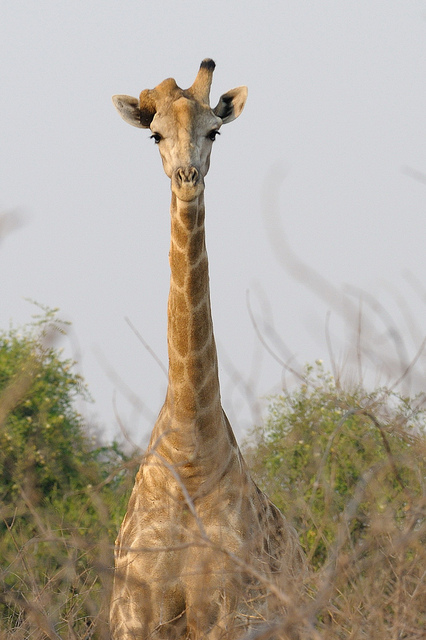}{}{The giraffe has two horns}{``The retrieved images are correct too.''}{0.19\textheight}
\caption{Representative LaSCo outcomes showing retained composition-required queries, problematic text, and non-unique targets.}
\label{fig:annotation-gallery-lasco}
\end{figure*}

\begin{figure*}[p]
\centering
\annotationpanel{0.9\textwidth}{\valid{} -- CIRCO q31}{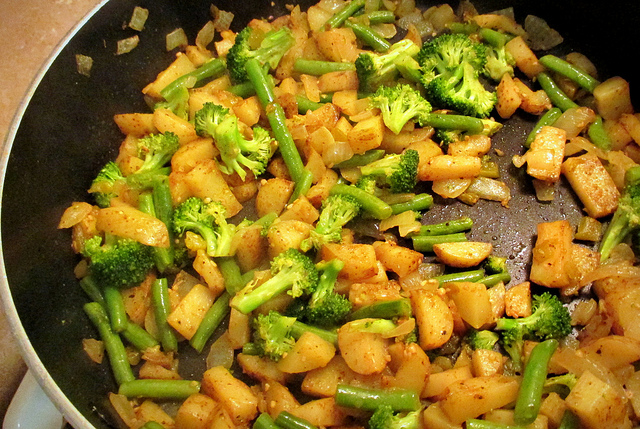}{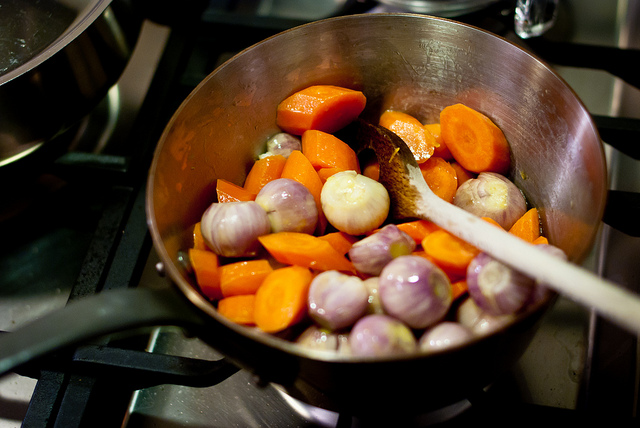}{is smaller and there are some sliced carrots}{Retained in the validated split; the query remains composition-required after the shortcut audit.}{0.19\textheight}

\vspace{8pt}

\annotationpanel{0.9\textwidth}{\invalidtext{} -- CIRCO q199}{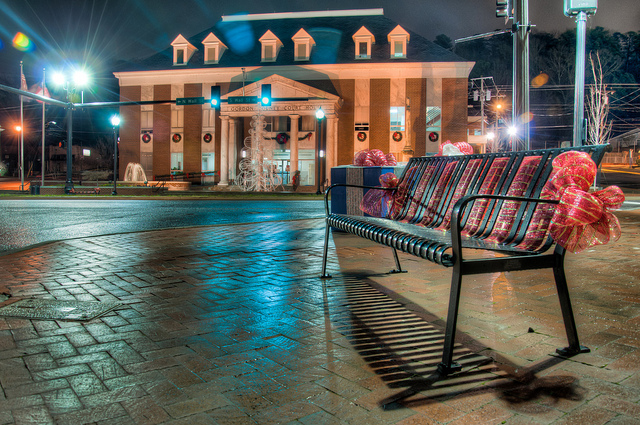}{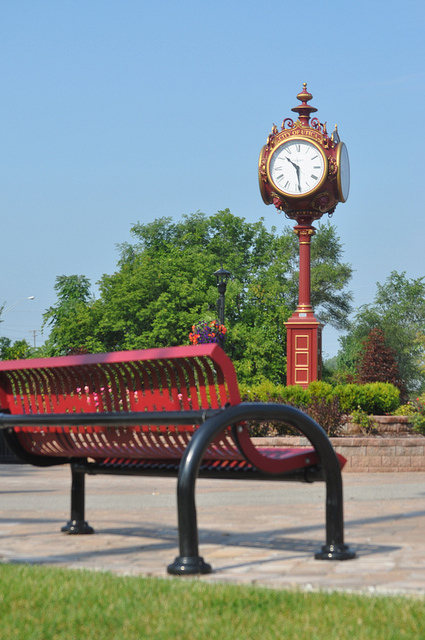}{is seen from behind and has a big clock in the background}{Annotated as problematic text; the stored audit note says the query is ``a bit underspecified.''}{0.19\textheight}
\caption{Representative CIRCO outcomes. Even in the smallest benchmark, the audit still surfaces both clearly retained queries and instructions whose semantics remain questionable.}
\label{fig:annotation-gallery-circo}
\end{figure*}

\clearpage
\section{Full Qualitative Gallery}
\label{app:qualitative-gallery}
The appendix retains the full qualitative gallery used throughout the paper.
For each dataset, we include several representative figures for each query category: a text-only shortcut, an image-only shortcut, a both-conditions shortcut, a composition-required case, and an unresolved case.

\begin{figure*}[p]
\centering
\includegraphics[width=\textwidth]{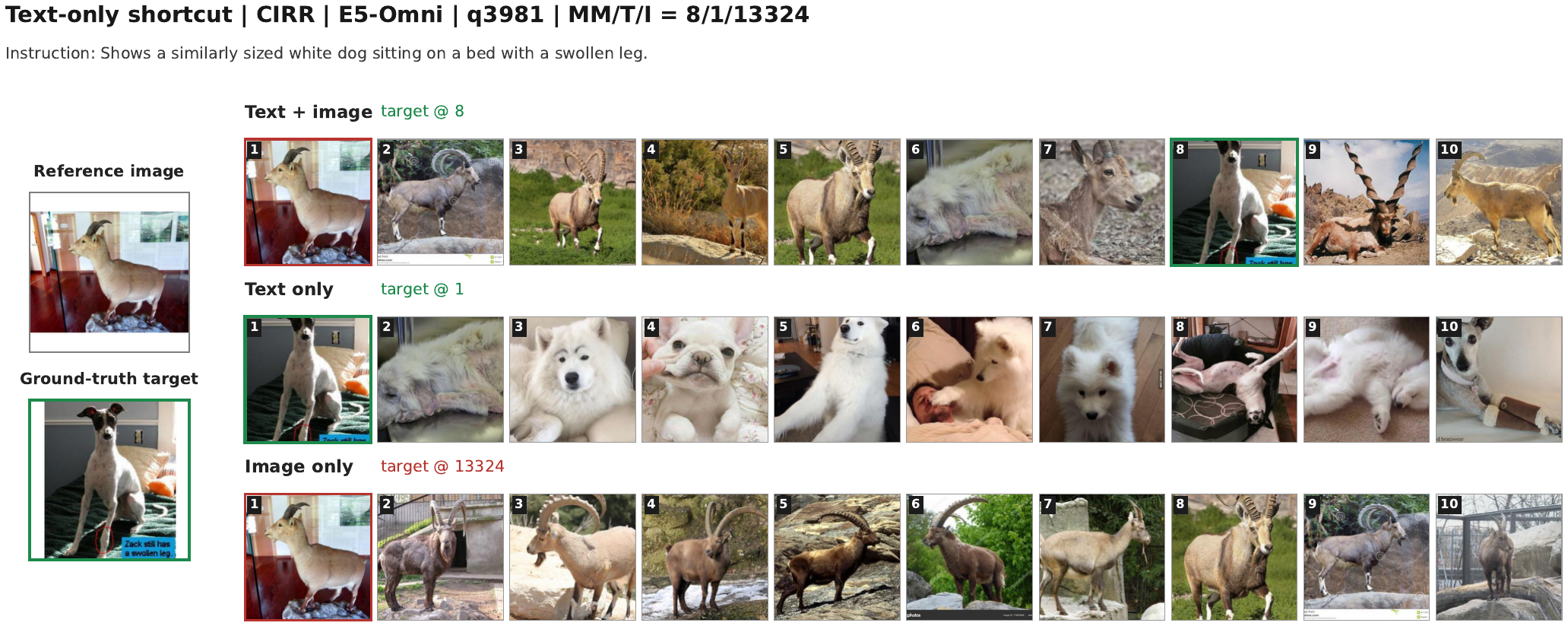}
\caption{Text-only shortcut on CIRR for E5-Omni (q3981). Text-only finds the target at rank 1 while image-only misses it (rank 13324), so the instruction already isolates the target.}
\label{fig:qual-cirr-text-only-e5omni}
\end{figure*}

\begin{figure*}[p]
\centering
\includegraphics[width=\textwidth]{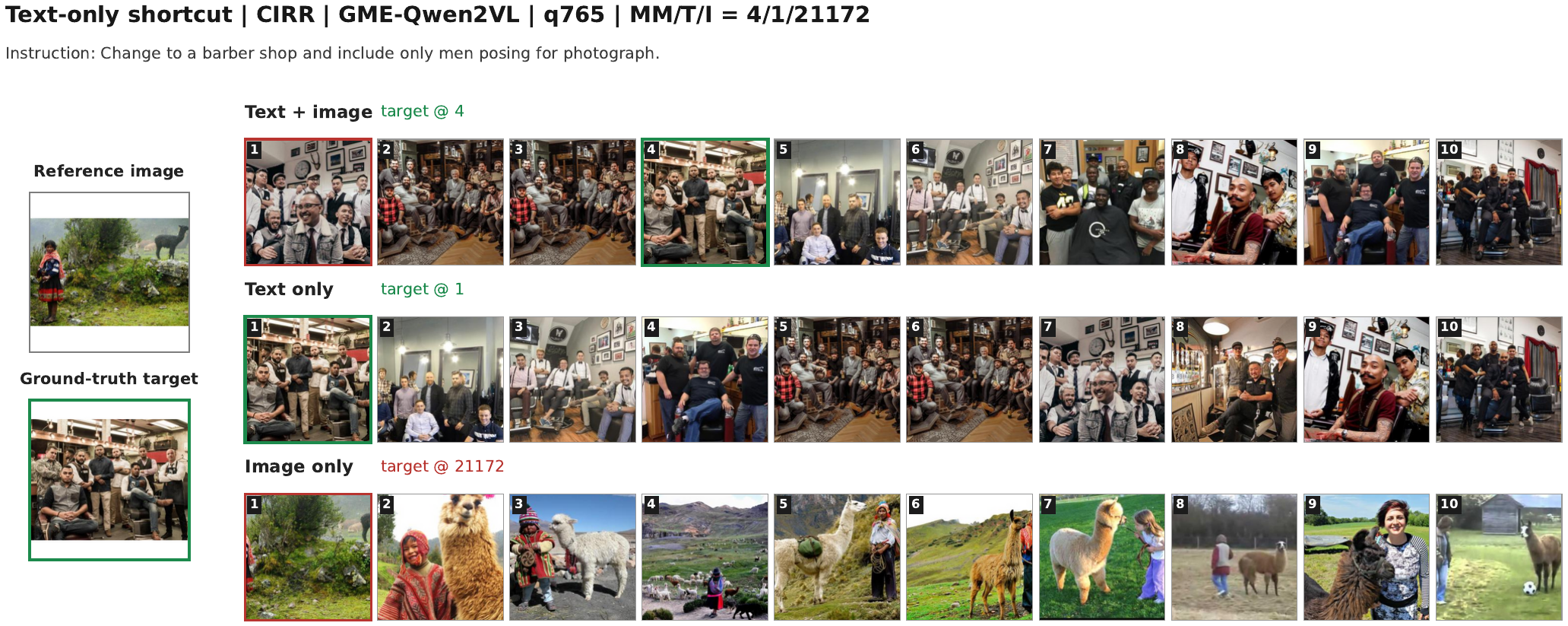}
\caption{Text-only shortcut on CIRR for GME-Qwen2VL (q765). Text-only finds the target at rank 1 while image-only misses it (rank 21172), so the instruction already isolates the target.}
\label{fig:qual-cirr-text-only-gmeqwen2vl}
\end{figure*}

\begin{figure*}[p]
\centering
\includegraphics[width=\textwidth]{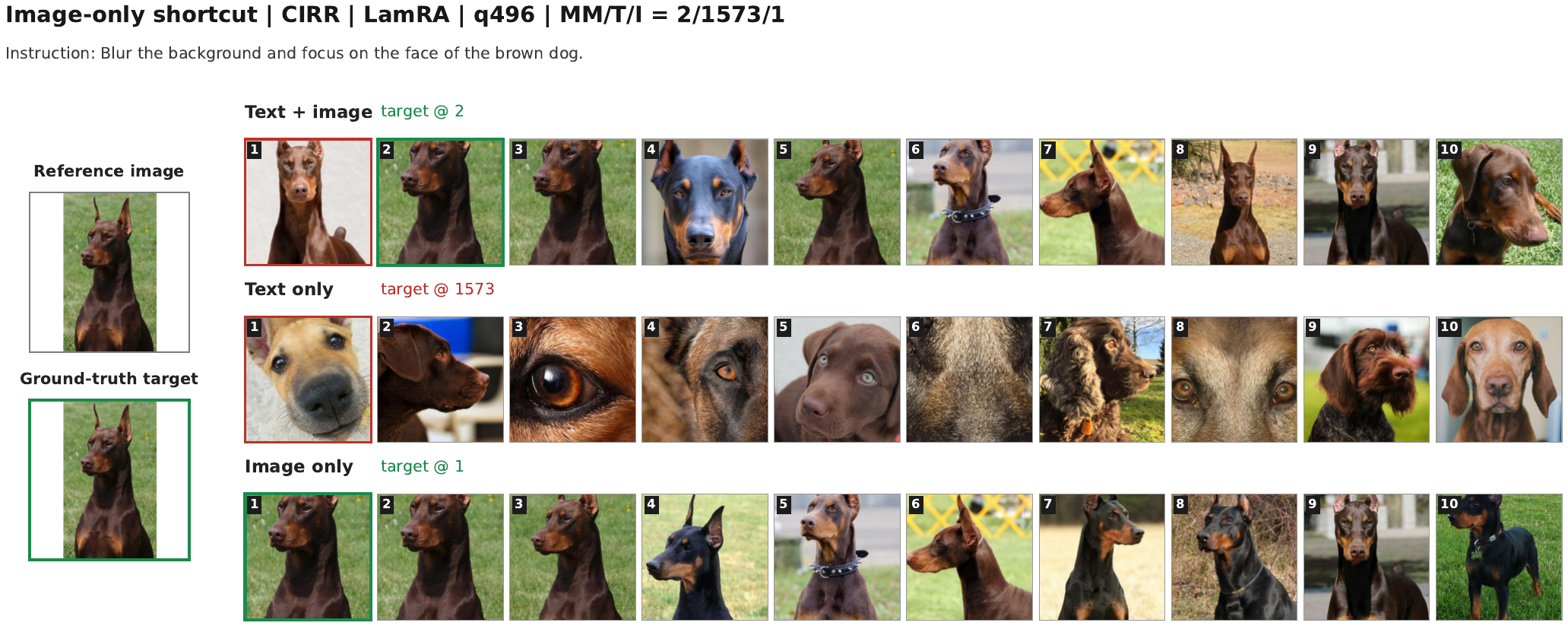}
\caption{Image-only shortcut on CIRR for LamRA (q496). Image-only finds the target at rank 1 while text-only misses it (rank 1573), so the reference image is doing most of the work.}
\label{fig:qual-cirr-image-only-lamra}
\end{figure*}

\begin{figure*}[p]
\centering
\includegraphics[width=\textwidth]{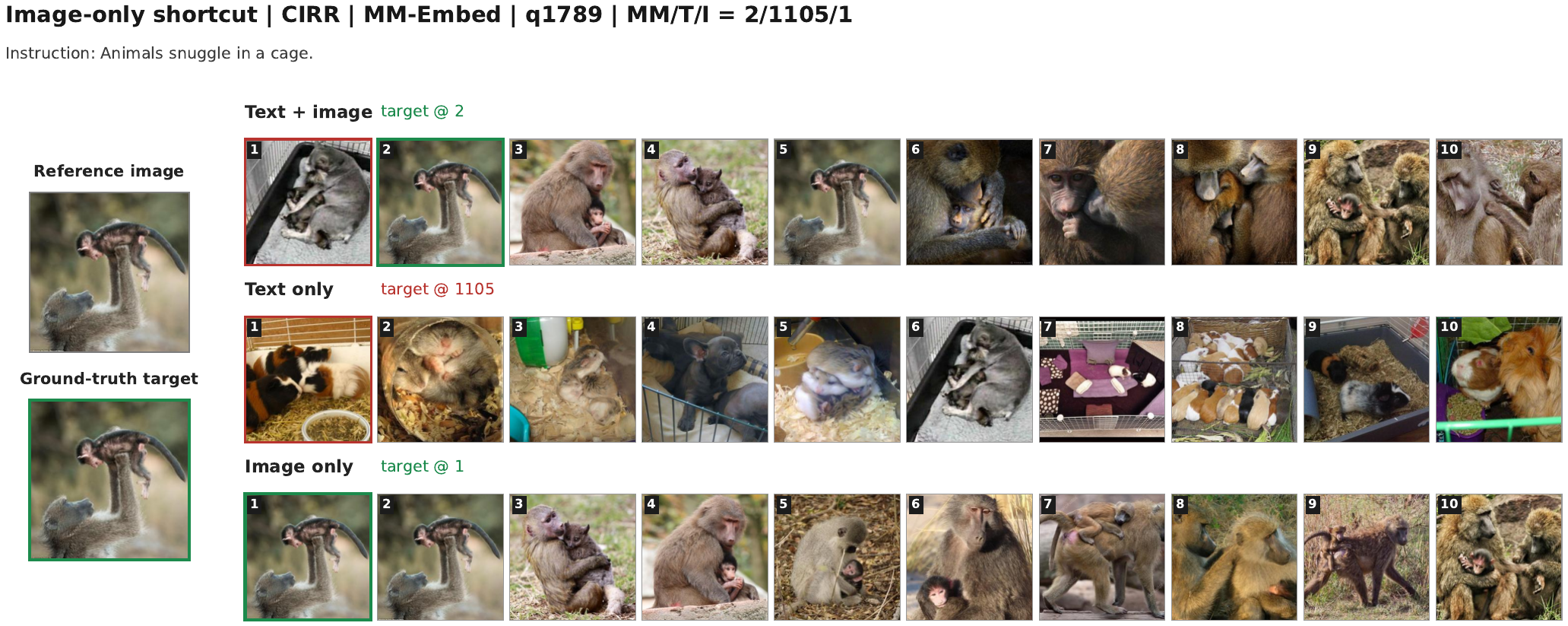}
\caption{Image-only shortcut on CIRR for MM-Embed (q1789). Image-only finds the target at rank 1 while text-only misses it (rank 1105), so the reference image is doing most of the work.}
\label{fig:qual-cirr-image-only-mmembed}
\end{figure*}

\begin{figure*}[p]
\centering
\includegraphics[width=\textwidth]{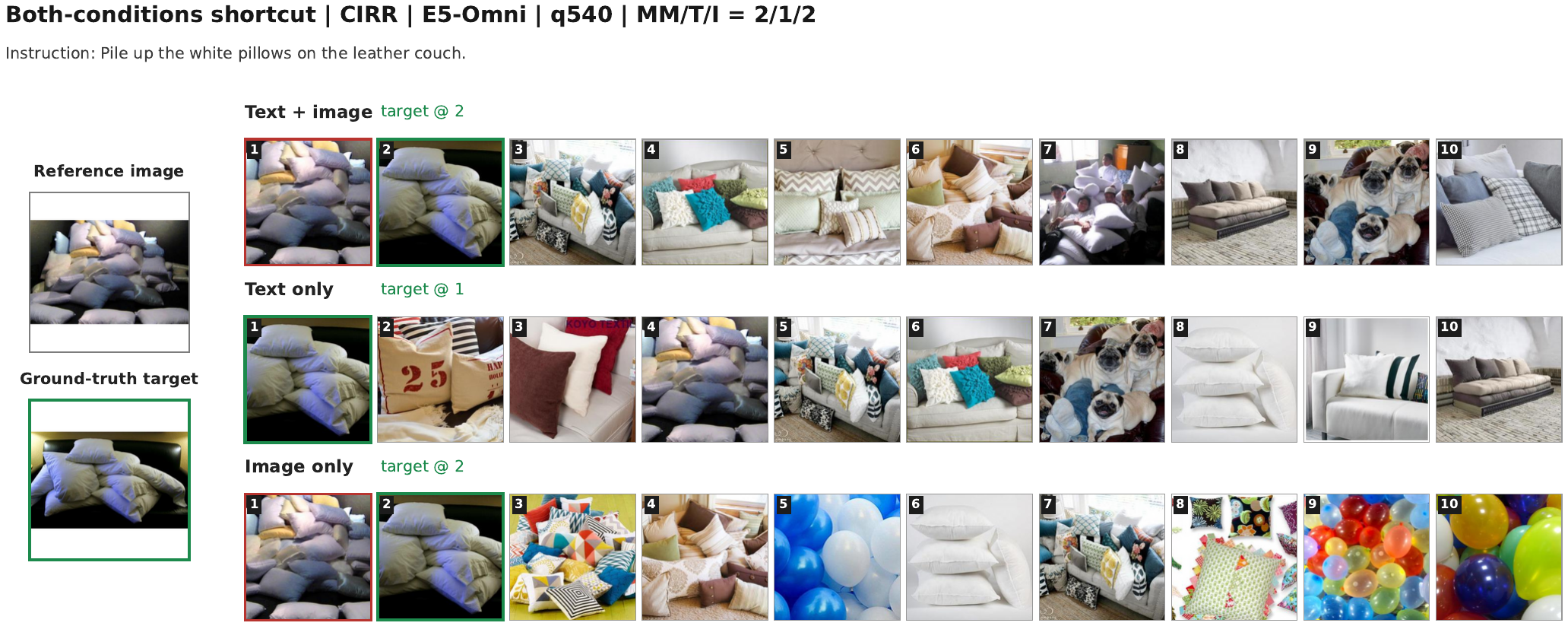}
\caption{Both-conditions shortcut on CIRR for E5-Omni (q540). Both text-only and image-only retrieve the target in the top-10 (ranks 1 and 2), placing the query in the both-conditions shortcut bucket.}
\label{fig:qual-cirr-both-e5omni}
\end{figure*}

\begin{figure*}[p]
\centering
\includegraphics[width=\textwidth]{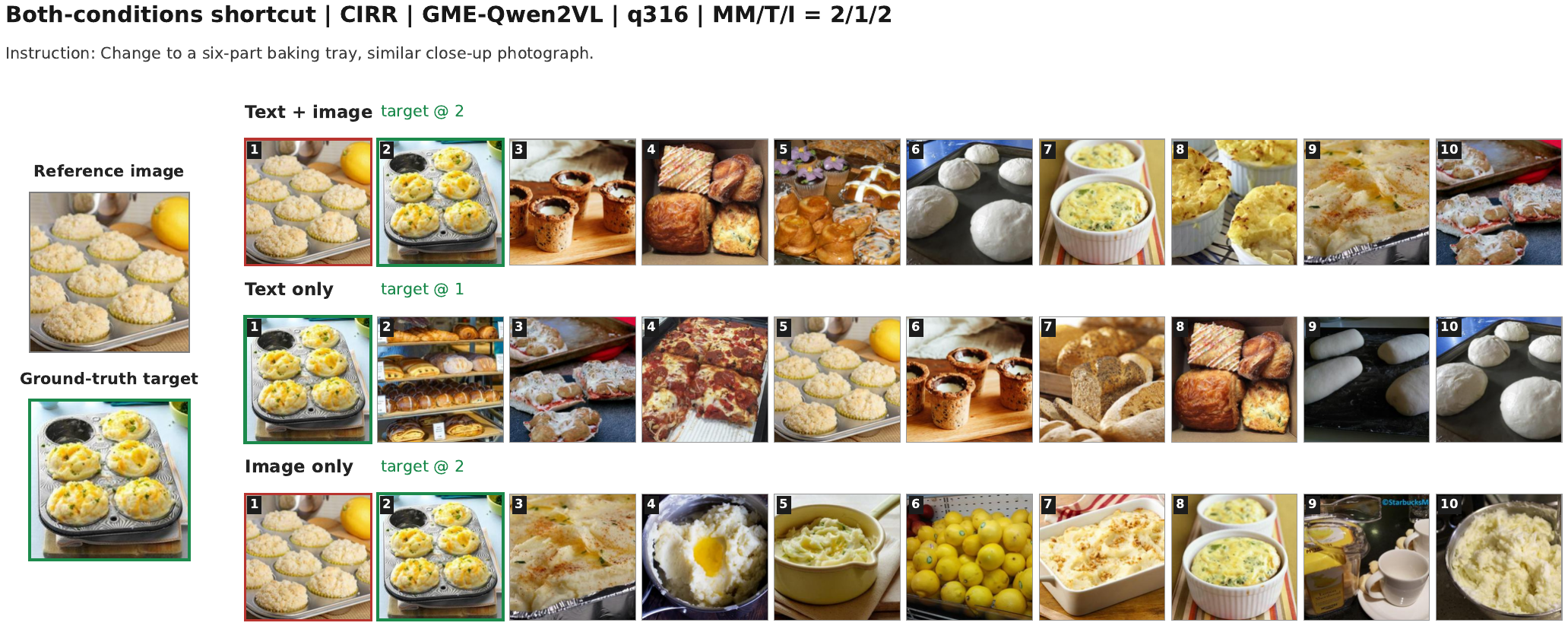}
\caption{Both-conditions shortcut on CIRR for GME-Qwen2VL (q316). Both text-only and image-only retrieve the target in the top-10 (ranks 1 and 2), placing the query in the both-conditions shortcut bucket.}
\label{fig:qual-cirr-both-gmeqwen2vl}
\end{figure*}

\begin{figure*}[p]
\centering
\includegraphics[width=\textwidth]{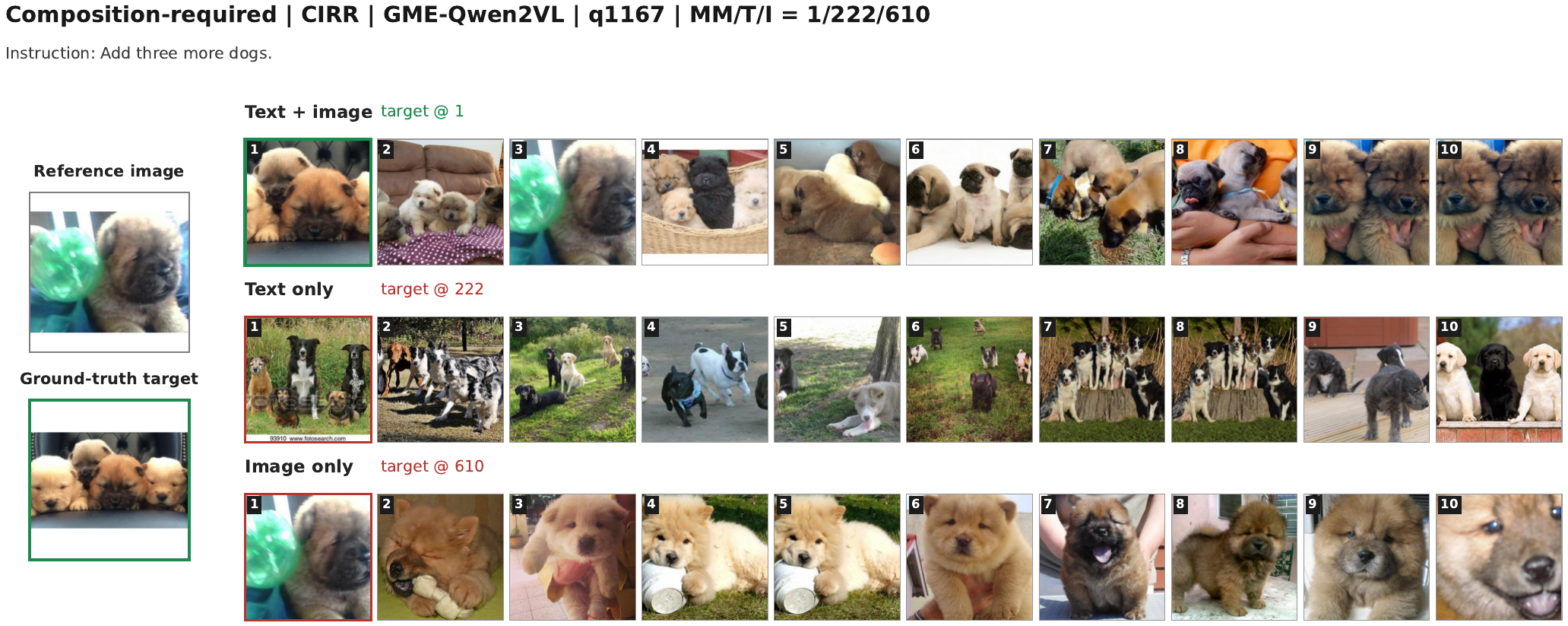}
\caption{Composition-required on CIRR for GME-Qwen2VL (q1167). Only multimodal retrieval places the target in the top-10 (MM/T/I = 1/222/610), so the edit must be grounded in the reference image.}
\label{fig:qual-cirr-composition-required-gmeqwen2vl}
\end{figure*}

\begin{figure*}[p]
\centering
\includegraphics[width=\textwidth]{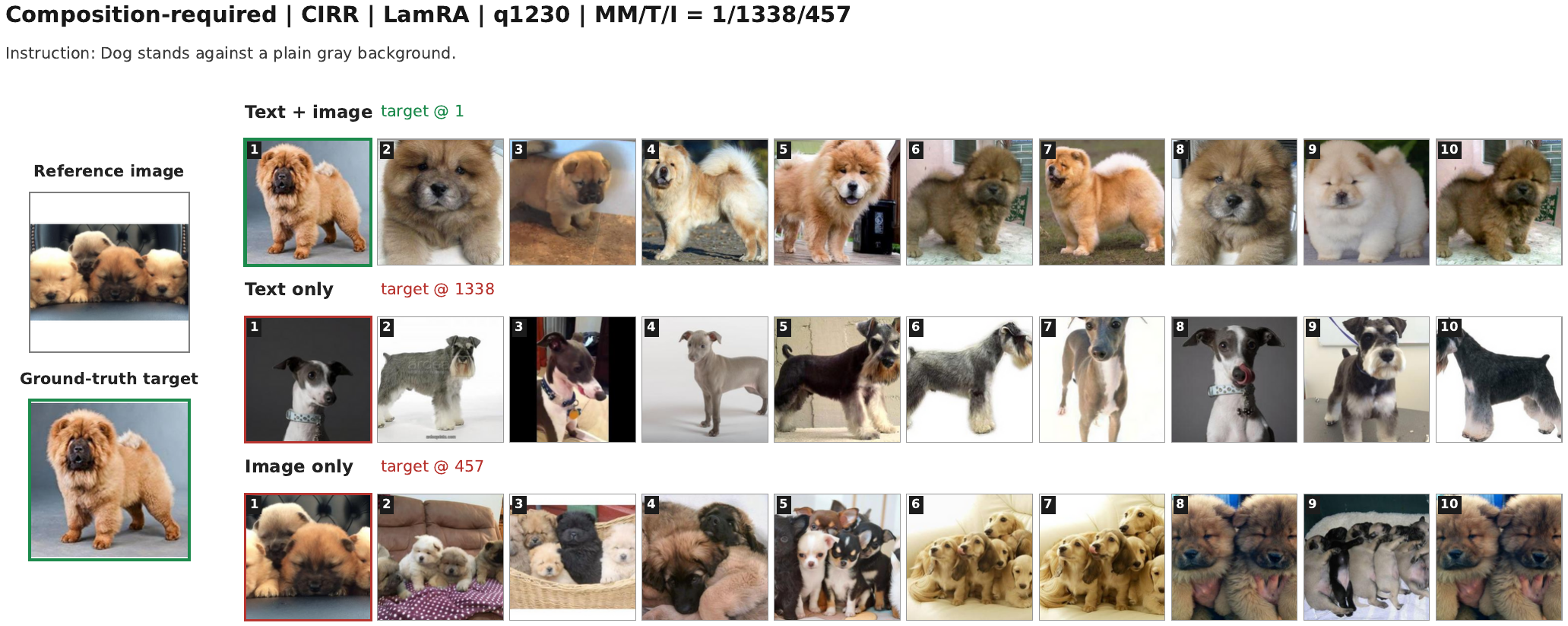}
\caption{Composition-required on CIRR for LamRA (q1230). Only multimodal retrieval places the target in the top-10 (MM/T/I = 1/1338/457), so the edit must be grounded in the reference image.}
\label{fig:qual-cirr-composition-required-lamra}
\end{figure*}

\begin{figure*}[p]
\centering
\includegraphics[width=\textwidth]{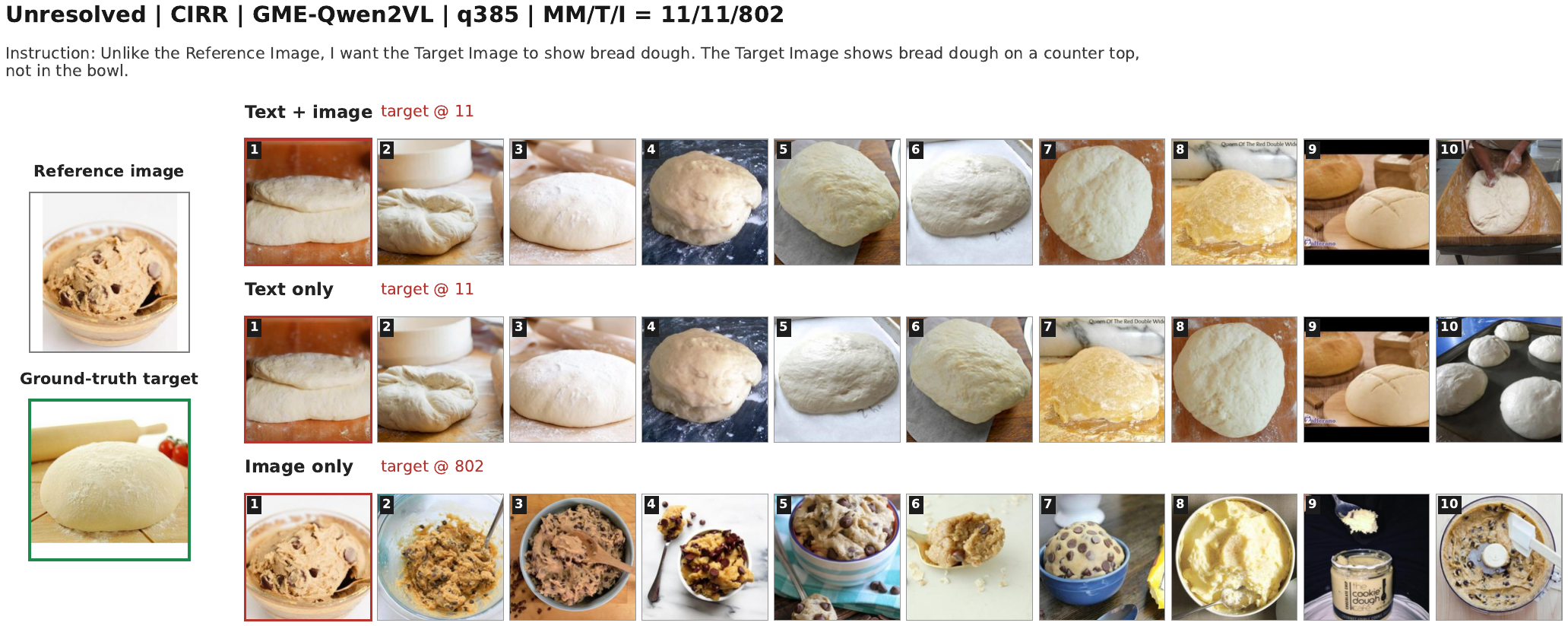}
\caption{Unresolved on CIRR for GME-Qwen2VL (q385). All three variants miss the target (MM/T/I = 11/11/802), placing the query in the unresolved set pending human validation.}
\label{fig:qual-cirr-unresolved-gmeqwen2vl}
\end{figure*}

\begin{figure*}[p]
\centering
\includegraphics[width=\textwidth]{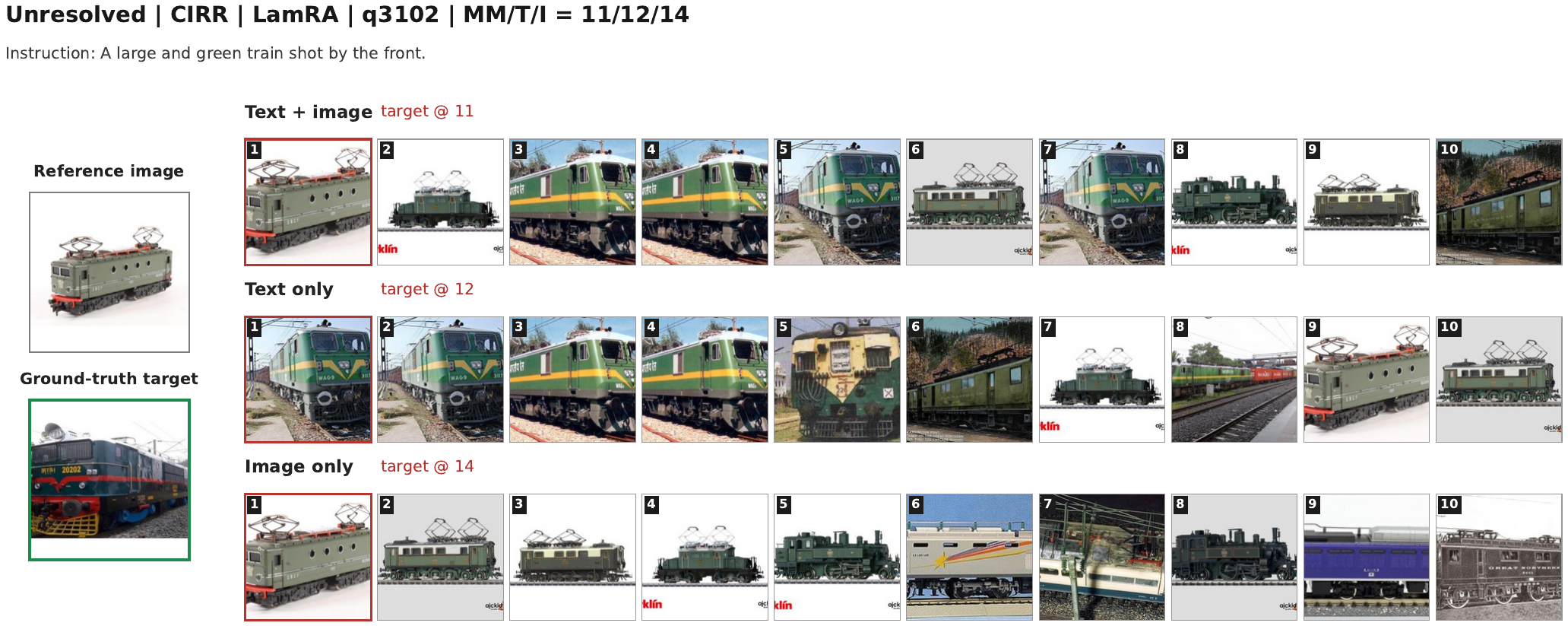}
\caption{Unresolved on CIRR for LamRA (q3102). All three variants miss the target (MM/T/I = 11/12/14), placing the query in the unresolved set pending human validation.}
\label{fig:qual-cirr-unresolved-lamra}
\end{figure*}

\begin{figure*}[p]
\centering
\includegraphics[width=\textwidth]{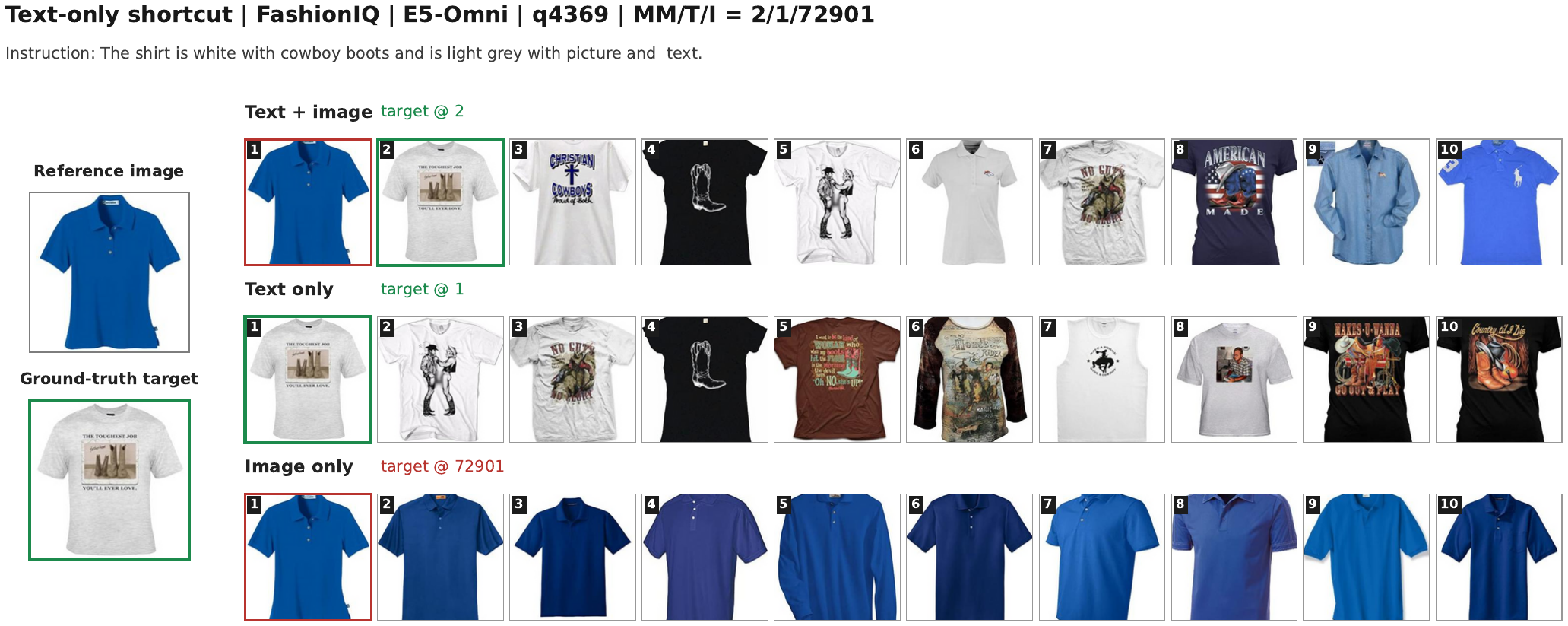}
\caption{Text-only shortcut on FashionIQ for E5-Omni (q4369). Text-only finds the target at rank 1 while image-only misses it (rank 72901), so the instruction already isolates the target.}
\label{fig:qual-fashioniq-text-only-e5omni}
\end{figure*}

\begin{figure*}[p]
\centering
\includegraphics[width=\textwidth]{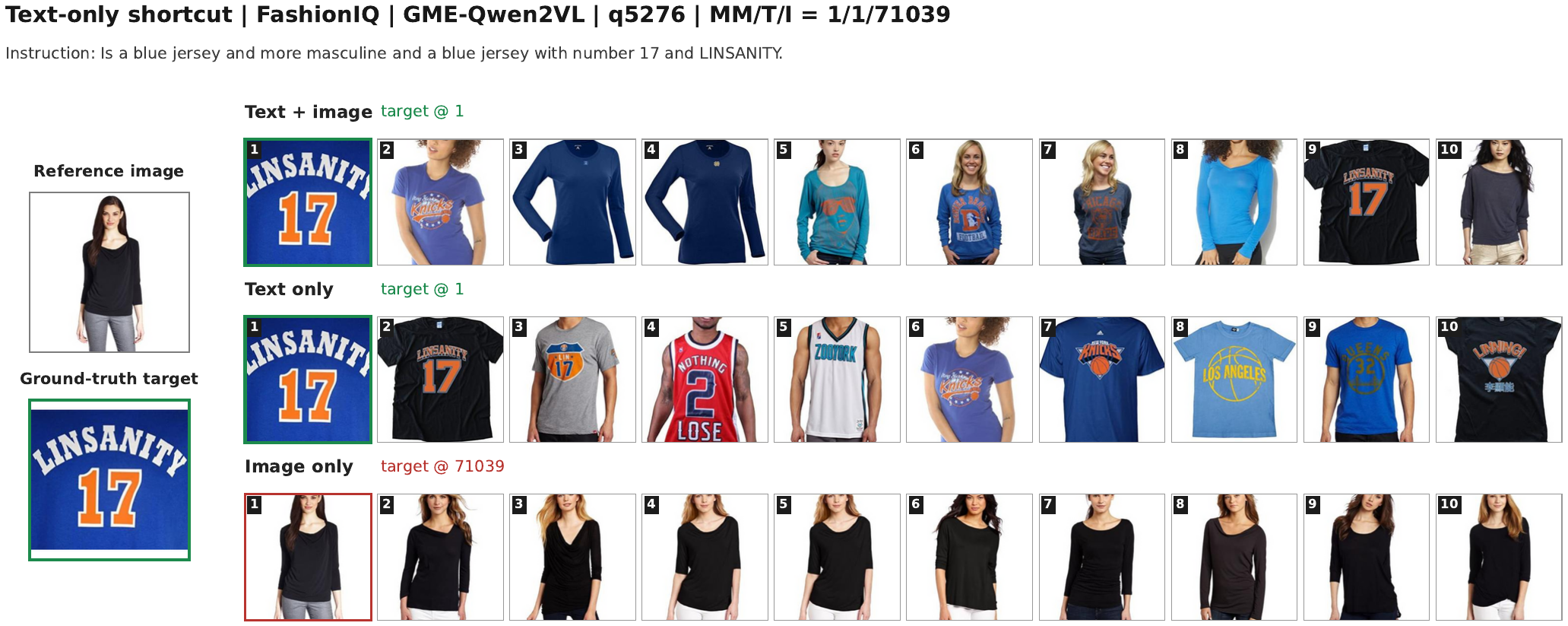}
\caption{Text-only shortcut on FashionIQ for GME-Qwen2VL (q5276). Text-only finds the target at rank 1 while image-only misses it (rank 71039), so the instruction already isolates the target.}
\label{fig:qual-fashioniq-text-only-gmeqwen2vl}
\end{figure*}

\begin{figure*}[p]
\centering
\includegraphics[width=\textwidth]{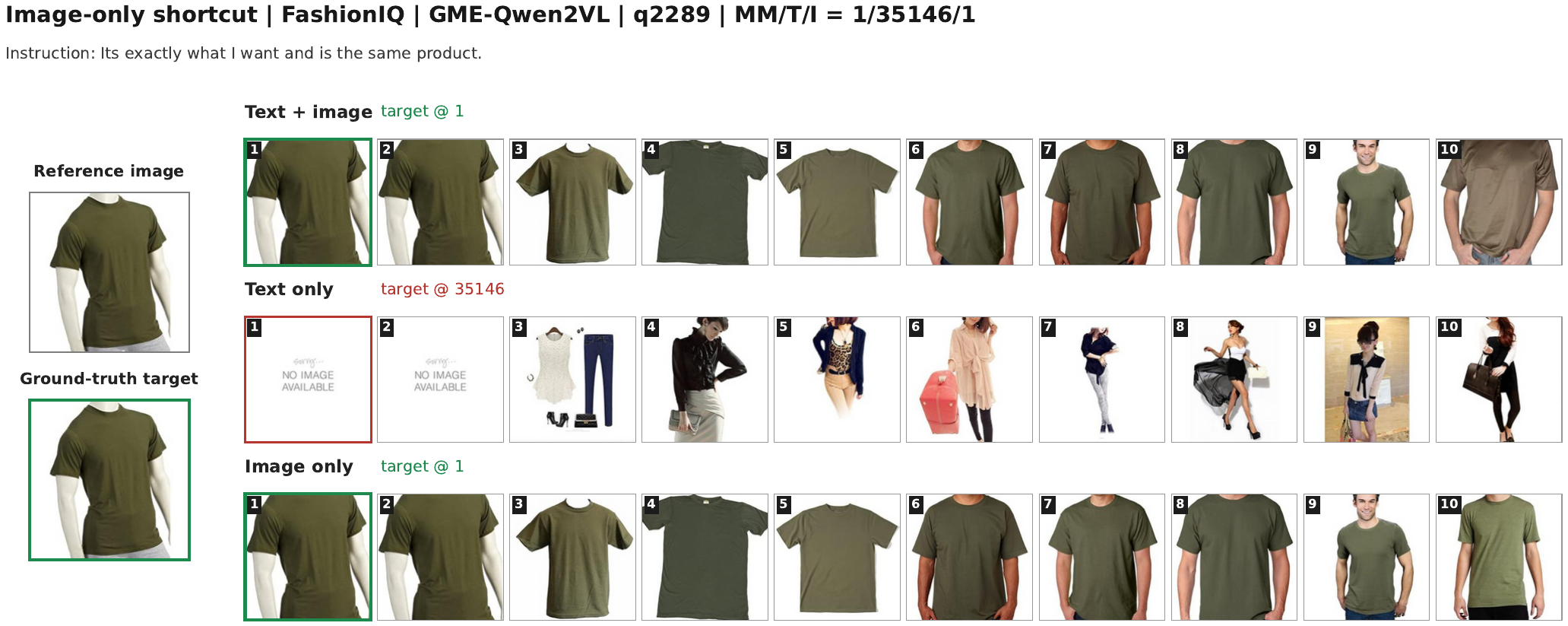}
\caption{Image-only shortcut on FashionIQ for GME-Qwen2VL (q2289). Image-only finds the target at rank 1 while text-only misses it (rank 35146), so the reference image is doing most of the work.}
\label{fig:qual-fashioniq-image-only-gmeqwen2vl}
\end{figure*}

\begin{figure*}[p]
\centering
\includegraphics[width=\textwidth]{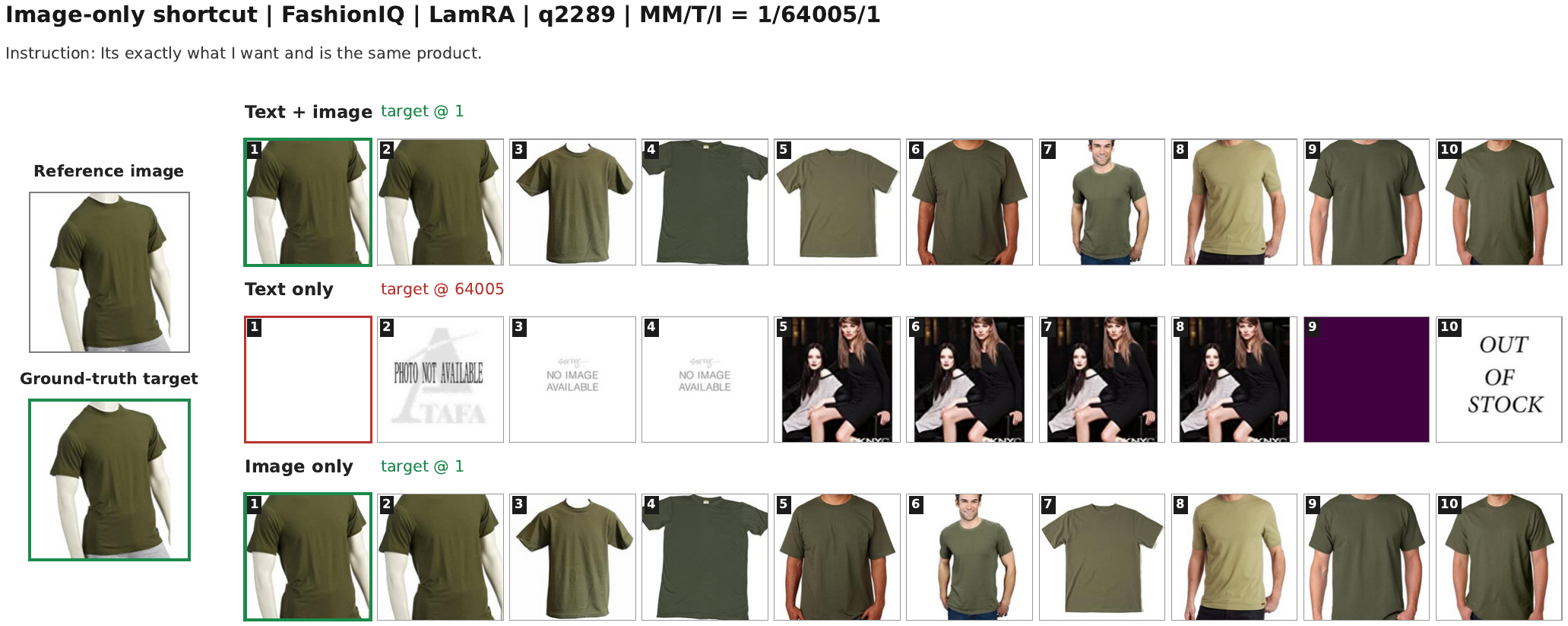}
\caption{Image-only shortcut on FashionIQ for LamRA (q2289). Image-only finds the target at rank 1 while text-only misses it (rank 64005), so the reference image is doing most of the work.}
\label{fig:qual-fashioniq-image-only-lamra}
\end{figure*}

\begin{figure*}[p]
\centering
\includegraphics[width=\textwidth]{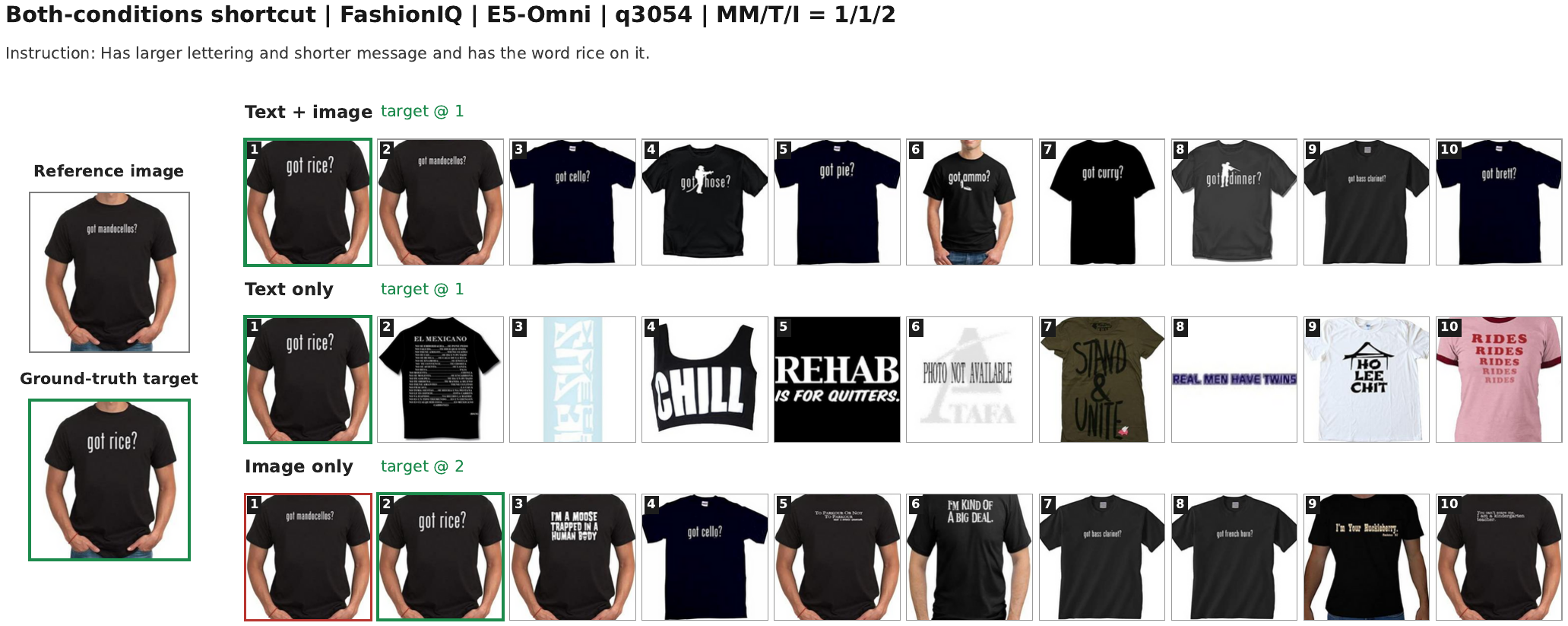}
\caption{Both-conditions shortcut on FashionIQ for E5-Omni (q3054). Both text-only and image-only retrieve the target in the top-10 (ranks 1 and 2), placing the query in the both-conditions shortcut bucket.}
\label{fig:qual-fashioniq-both-e5omni}
\end{figure*}

\begin{figure*}[p]
\centering
\includegraphics[width=\textwidth]{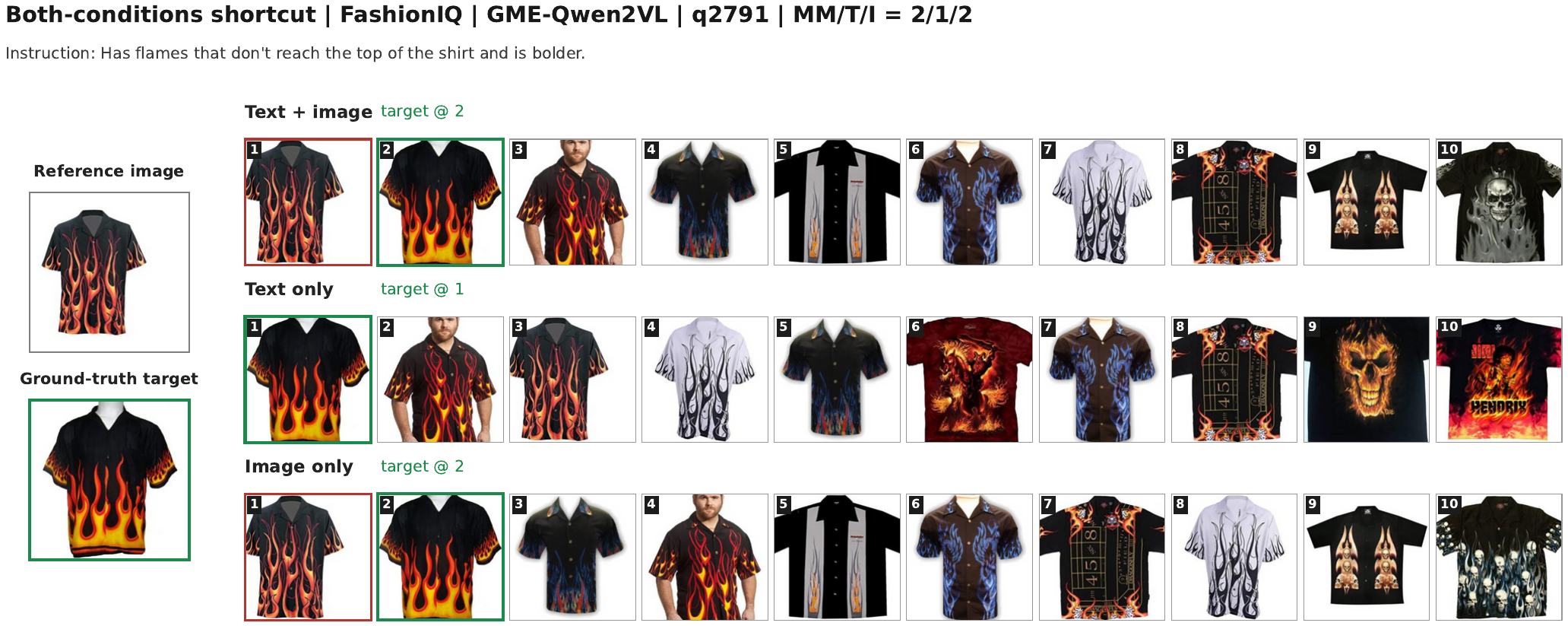}
\caption{Both-conditions shortcut on FashionIQ for GME-Qwen2VL (q2791). Both text-only and image-only retrieve the target in the top-10 (ranks 1 and 2), placing the query in the both-conditions shortcut bucket.}
\label{fig:qual-fashioniq-both-gmeqwen2vl}
\end{figure*}

\begin{figure*}[p]
\centering
\includegraphics[width=\textwidth]{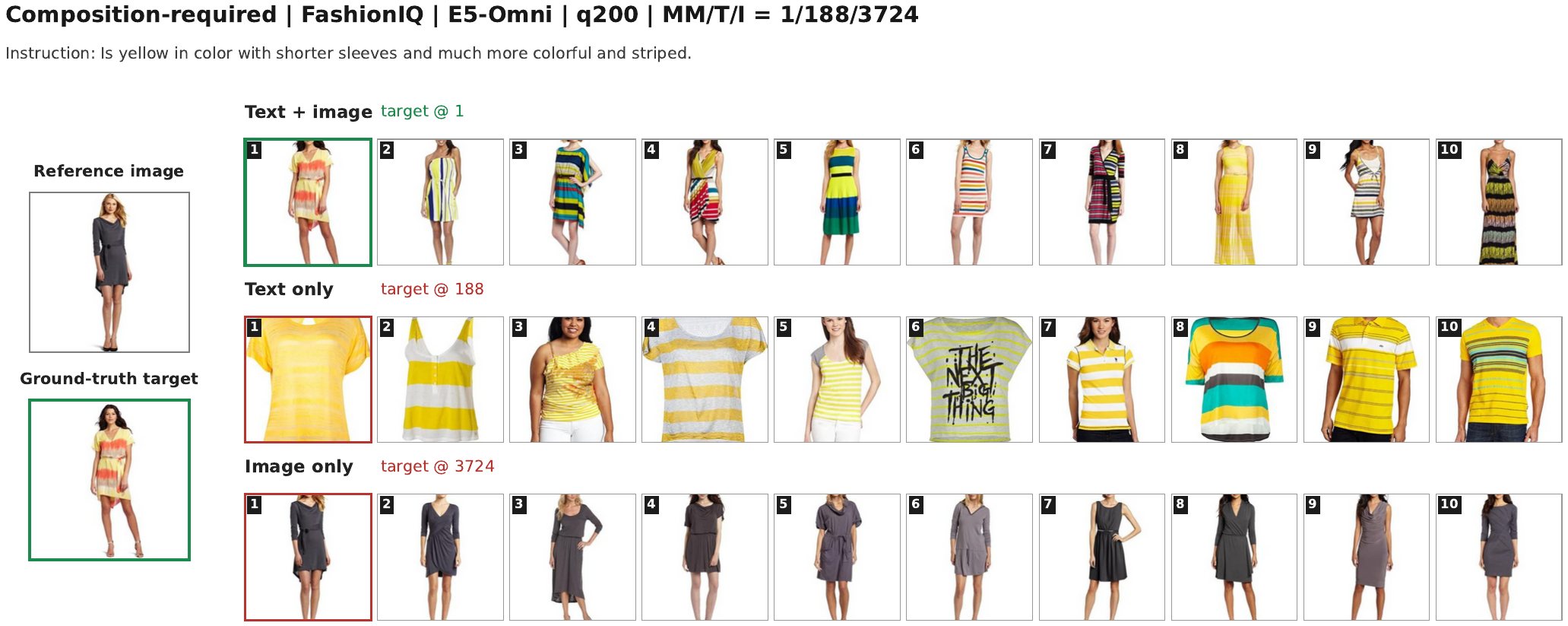}
\caption{Composition-required on FashionIQ for E5-Omni (q200). Only multimodal retrieval places the target in the top-10 (MM/T/I = 1/188/3724), so the edit must be grounded in the reference image.}
\label{fig:qual-fashioniq-composition-required-e5omni}
\end{figure*}

\begin{figure*}[p]
\centering
\includegraphics[width=\textwidth]{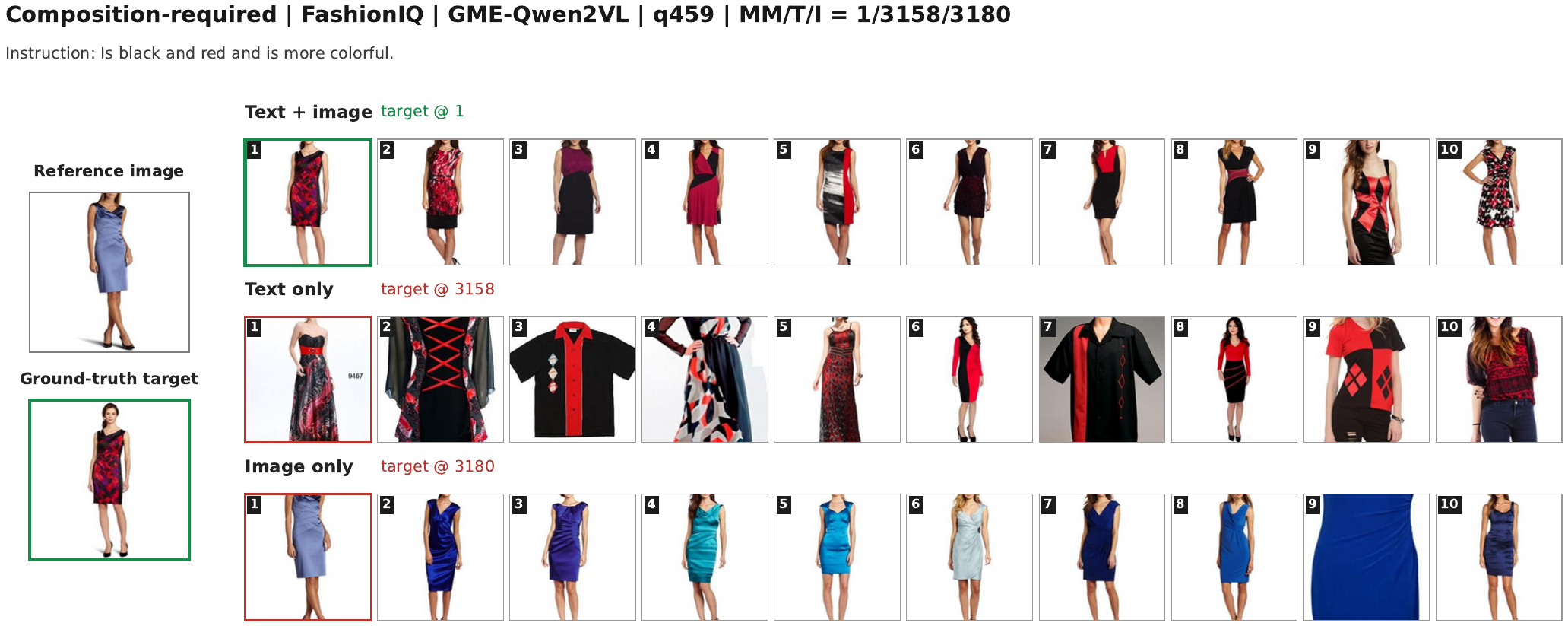}
\caption{Composition-required on FashionIQ for GME-Qwen2VL (q459). Only multimodal retrieval places the target in the top-10 (MM/T/I = 1/3158/3180), so the edit must be grounded in the reference image.}
\label{fig:qual-fashioniq-composition-required-gmeqwen2vl}
\end{figure*}

\begin{figure*}[p]
\centering
\includegraphics[width=\textwidth]{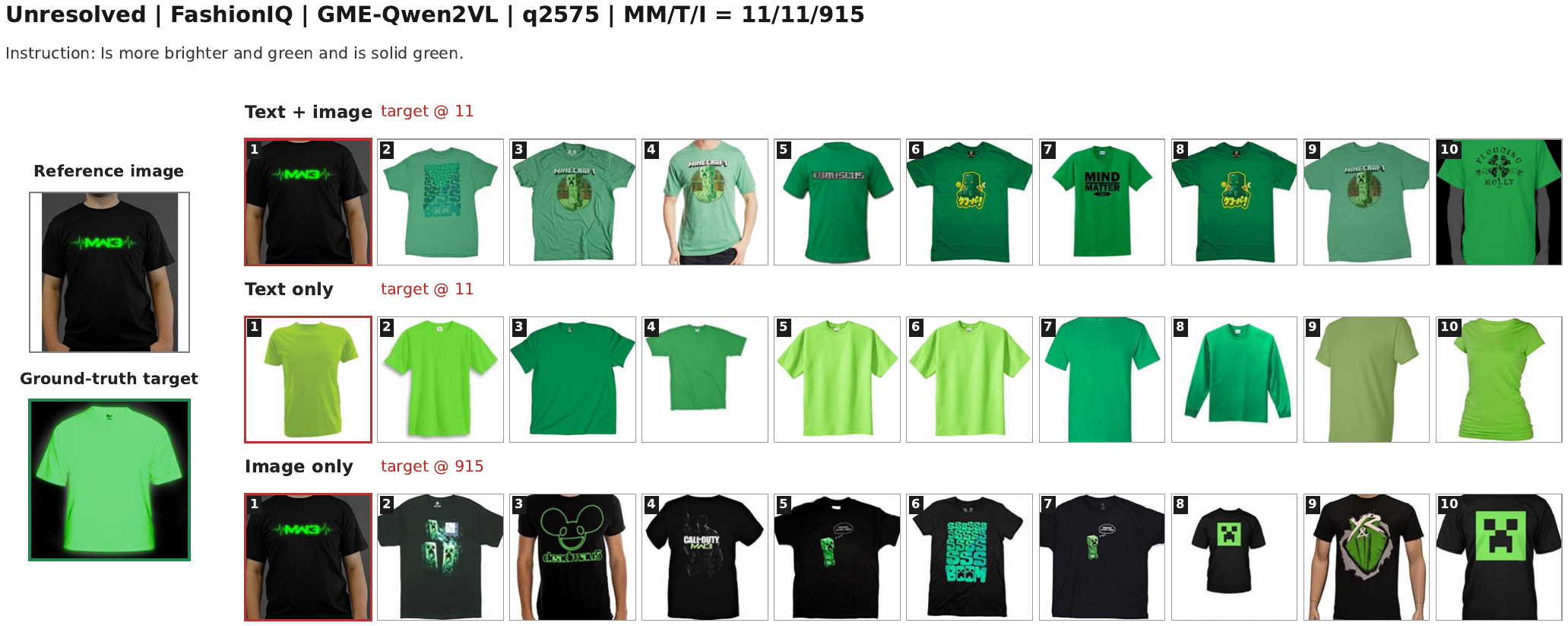}
\caption{Unresolved on FashionIQ for GME-Qwen2VL (q2575). All three variants miss the target (MM/T/I = 11/11/915), placing the query in the unresolved set pending human validation.}
\label{fig:qual-fashioniq-unresolved-gmeqwen2vl}
\end{figure*}

\begin{figure*}[p]
\centering
\includegraphics[width=\textwidth]{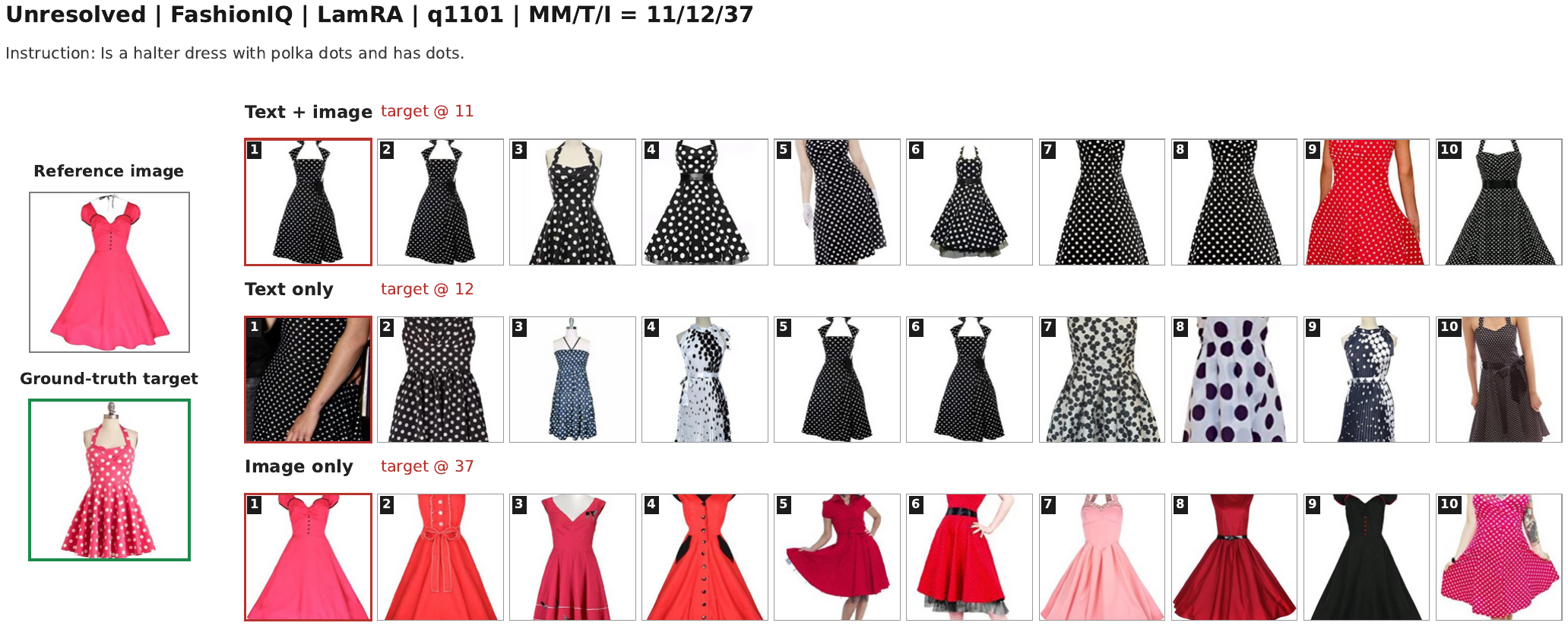}
\caption{Unresolved on FashionIQ for LamRA (q1101). All three variants miss the target (MM/T/I = 11/12/37), placing the query in the unresolved set pending human validation.}
\label{fig:qual-fashioniq-unresolved-lamra}
\end{figure*}

\begin{figure*}[p]
\centering
\includegraphics[width=\textwidth]{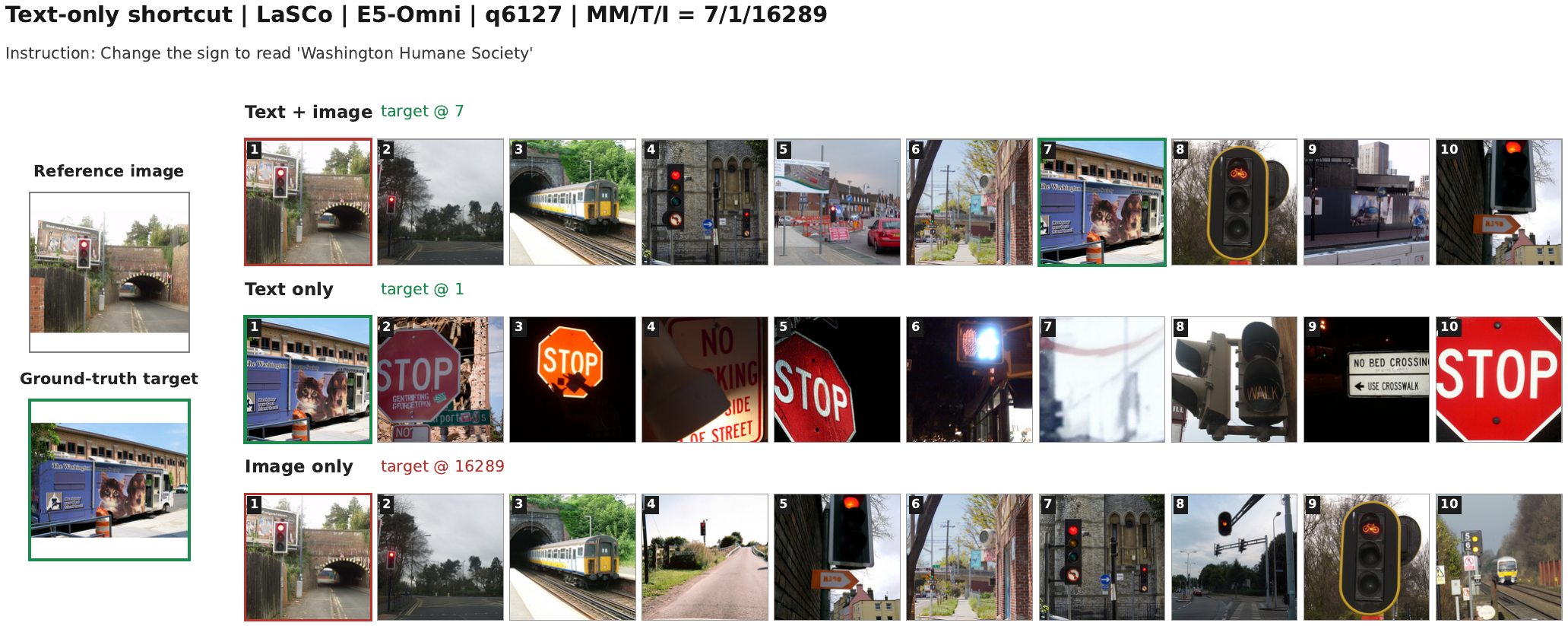}
\caption{Text-only shortcut on LaSCo for E5-Omni (q6127). Text-only finds the target at rank 1 while image-only misses it (rank 16289), so the instruction already isolates the target.}
\label{fig:qual-lasco-text-only-e5omni}
\end{figure*}

\begin{figure*}[p]
\centering
\includegraphics[width=\textwidth]{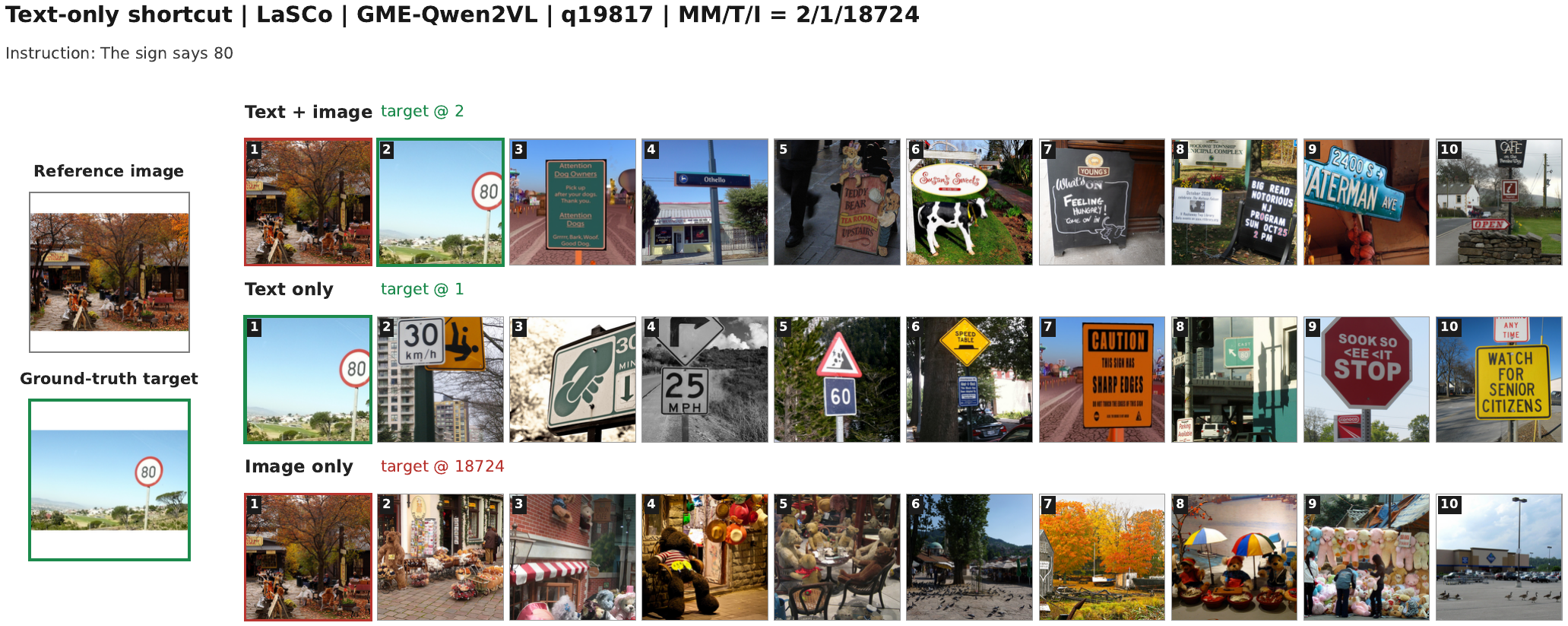}
\caption{Text-only shortcut on LaSCo for GME-Qwen2VL (q19817). Text-only finds the target at rank 1 while image-only misses it (rank 18724), so the instruction already isolates the target.}
\label{fig:qual-lasco-text-only-gmeqwen2vl}
\end{figure*}

\begin{figure*}[p]
\centering
\includegraphics[width=\textwidth]{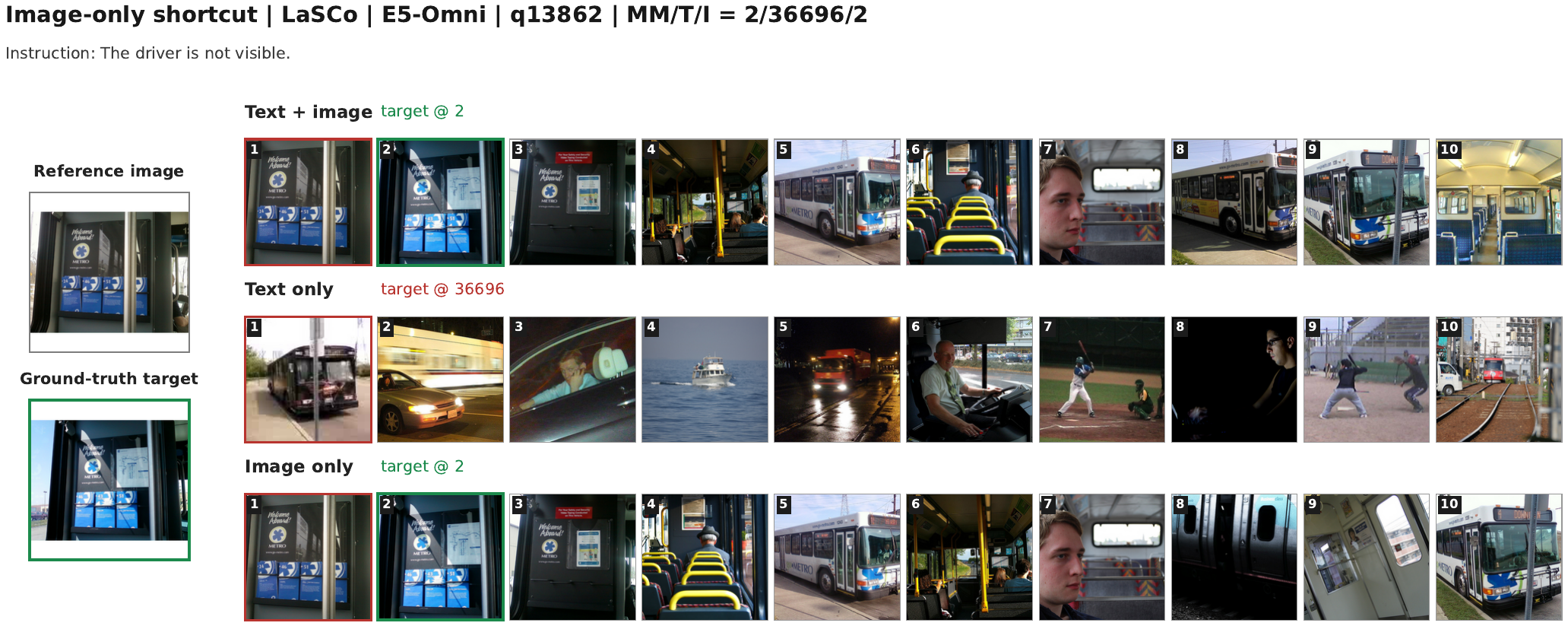}
\caption{Image-only shortcut on LaSCo for E5-Omni (q13862). Image-only finds the target at rank 2 while text-only misses it (rank 36696), so the reference image is doing most of the work.}
\label{fig:qual-lasco-image-only-e5omni}
\end{figure*}

\begin{figure*}[p]
\centering
\includegraphics[width=\textwidth]{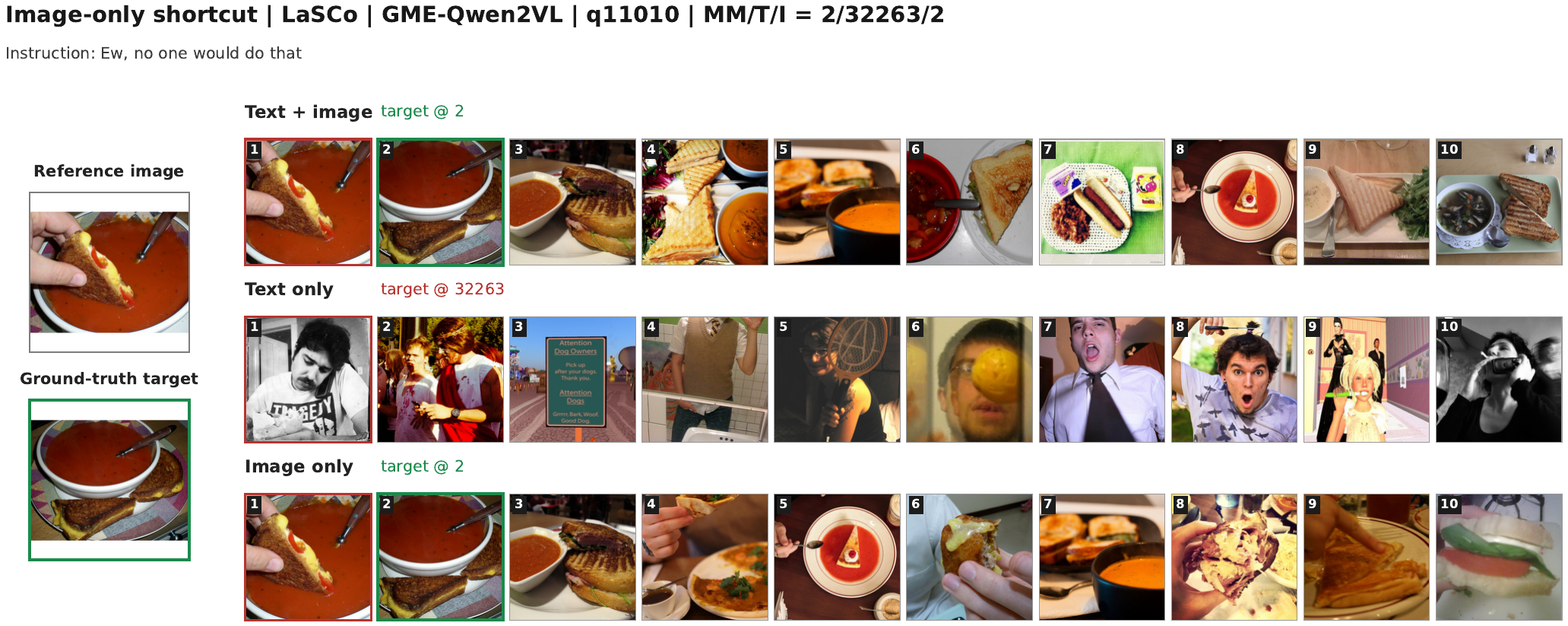}
\caption{Image-only shortcut on LaSCo for GME-Qwen2VL (q11010). Image-only finds the target at rank 2 while text-only misses it (rank 32263), so the reference image is doing most of the work.}
\label{fig:qual-lasco-image-only-gmeqwen2vl}
\end{figure*}

\begin{figure*}[p]
\centering
\includegraphics[width=\textwidth]{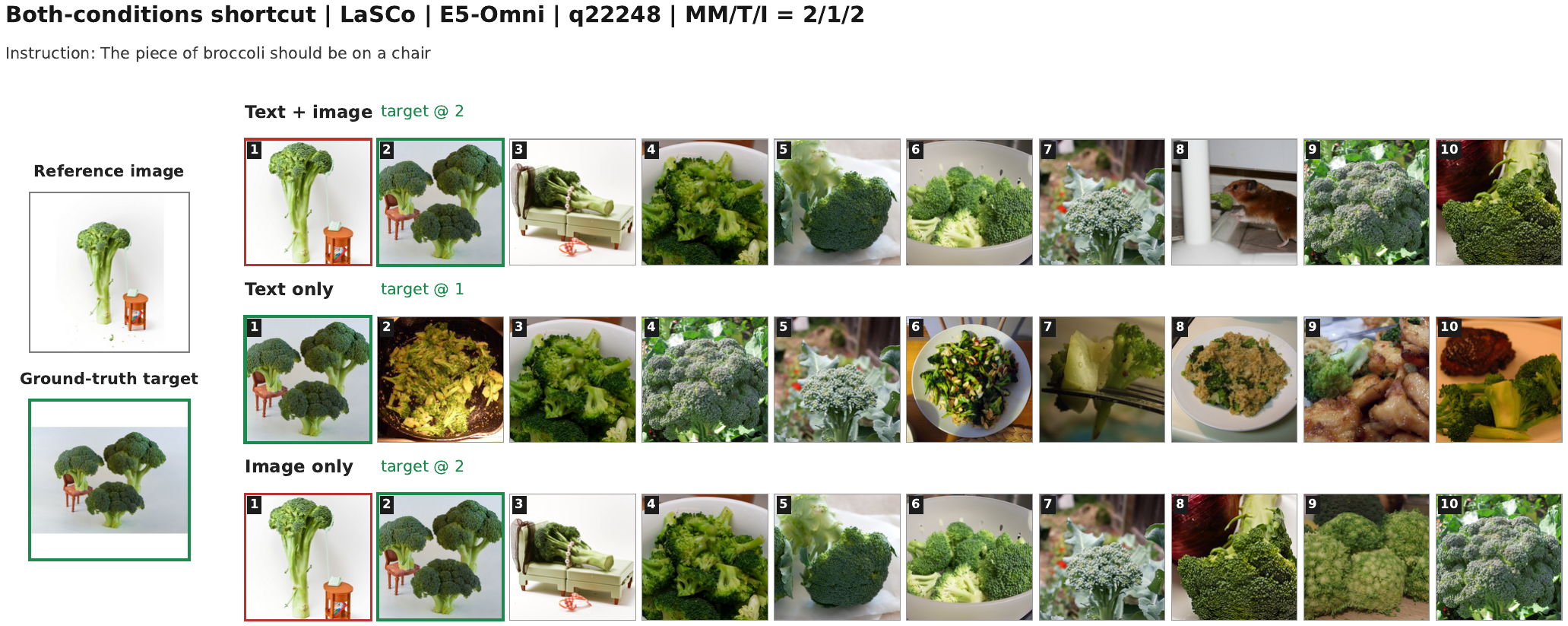}
\caption{Both-conditions shortcut on LaSCo for E5-Omni (q22248). Both text-only and image-only retrieve the target in the top-10 (ranks 1 and 2), placing the query in the both-conditions shortcut bucket.}
\label{fig:qual-lasco-both-e5omni}
\end{figure*}

\begin{figure*}[p]
\centering
\includegraphics[width=\textwidth]{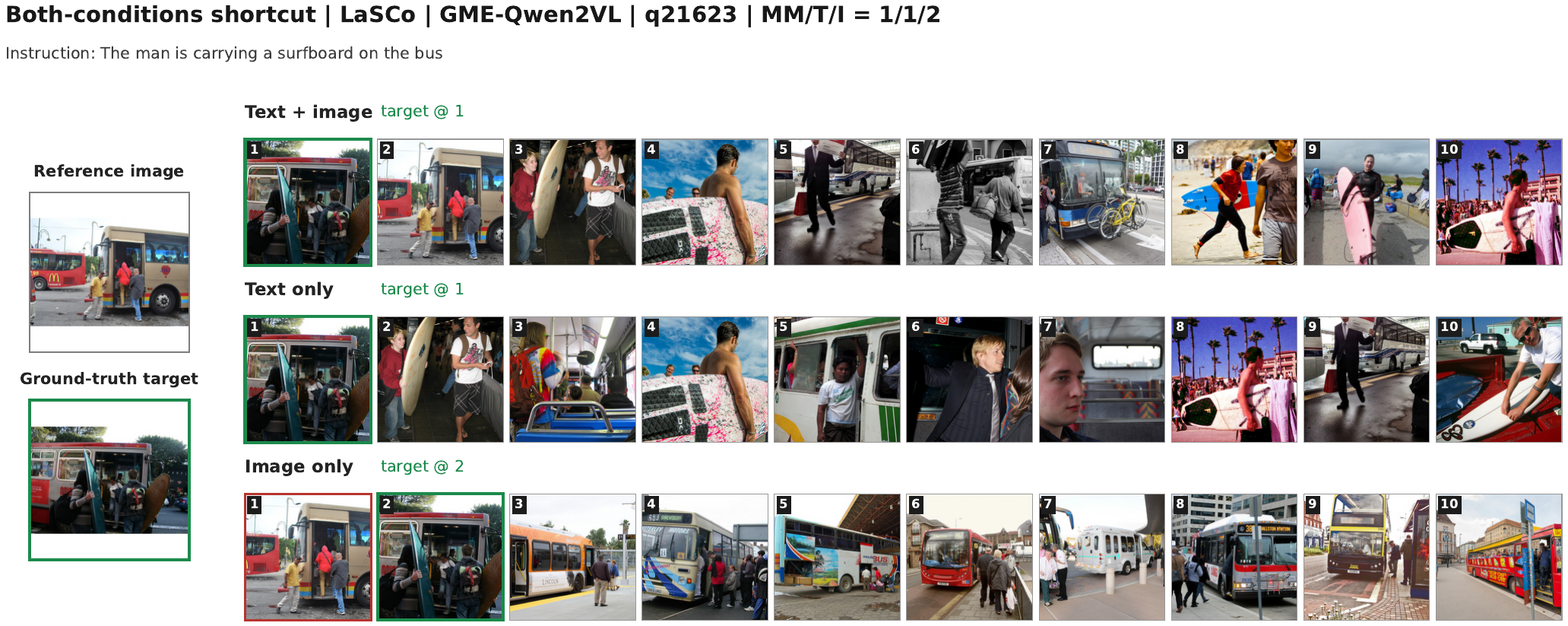}
\caption{Both-conditions shortcut on LaSCo for GME-Qwen2VL (q21623). Both text-only and image-only retrieve the target in the top-10 (ranks 1 and 2), placing the query in the both-conditions shortcut bucket.}
\label{fig:qual-lasco-both-gmeqwen2vl}
\end{figure*}

\begin{figure*}[p]
\centering
\includegraphics[width=\textwidth]{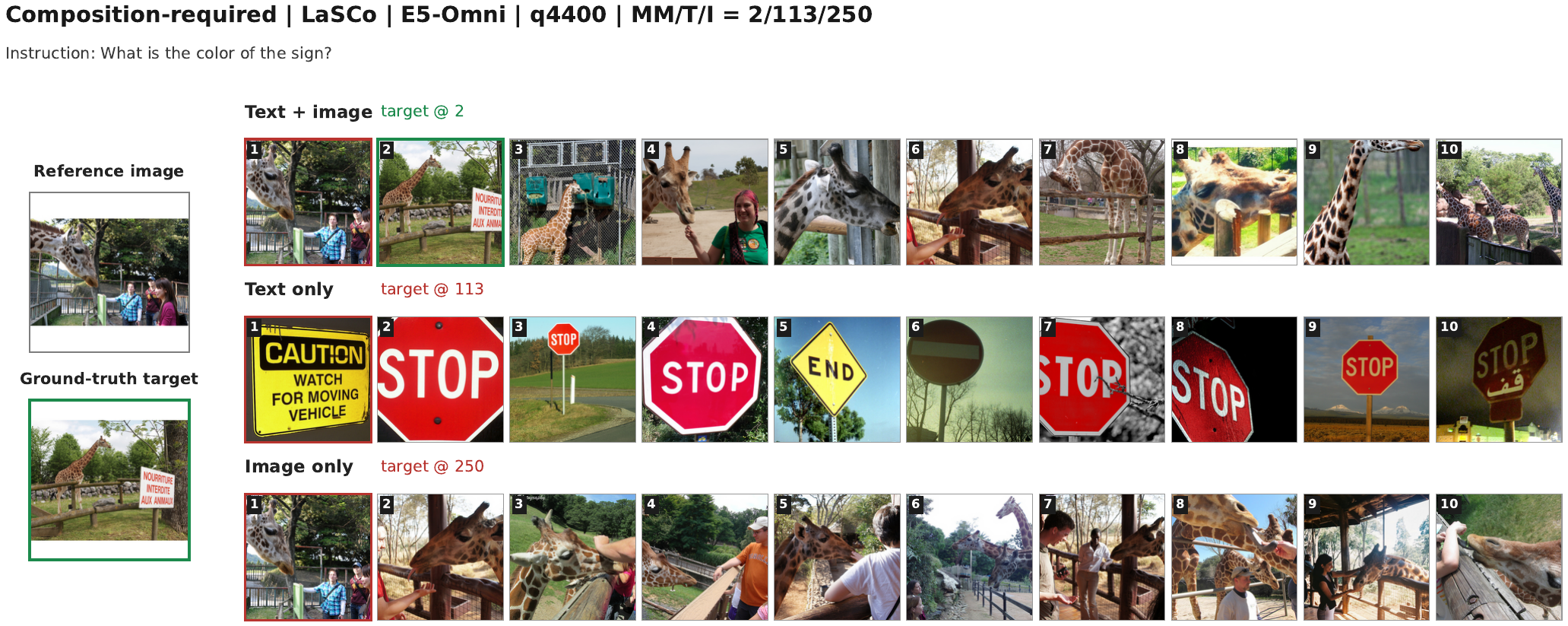}
\caption{Composition-required on LaSCo for E5-Omni (q4400). Only multimodal retrieval places the target in the top-10 (MM/T/I = 2/113/250), so the edit must be grounded in the reference image.}
\label{fig:qual-lasco-composition-required-e5omni}
\end{figure*}

\begin{figure*}[p]
\centering
\includegraphics[width=\textwidth]{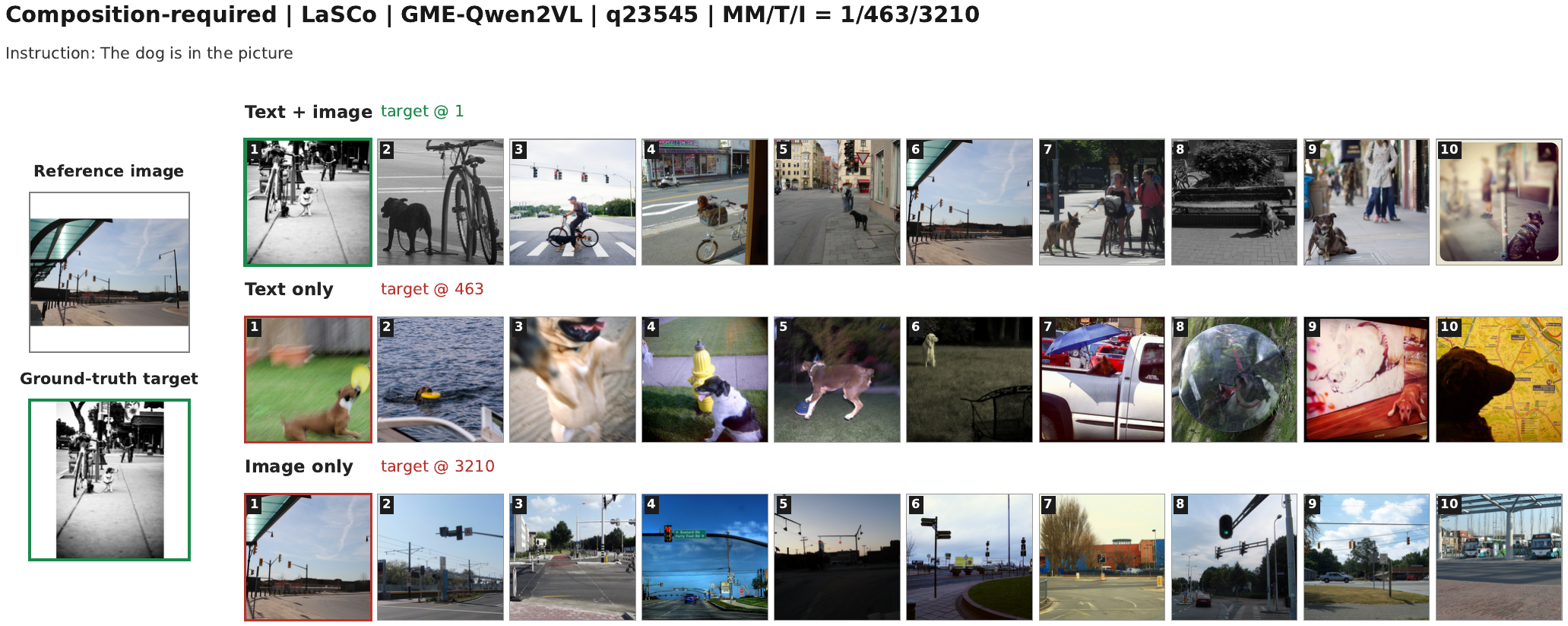}
\caption{Composition-required on LaSCo for GME-Qwen2VL (q23545). Only multimodal retrieval places the target in the top-10 (MM/T/I = 1/463/3210), so the edit must be grounded in the reference image.}
\label{fig:qual-lasco-composition-required-gmeqwen2vl}
\end{figure*}

\begin{figure*}[p]
\centering
\includegraphics[width=\textwidth]{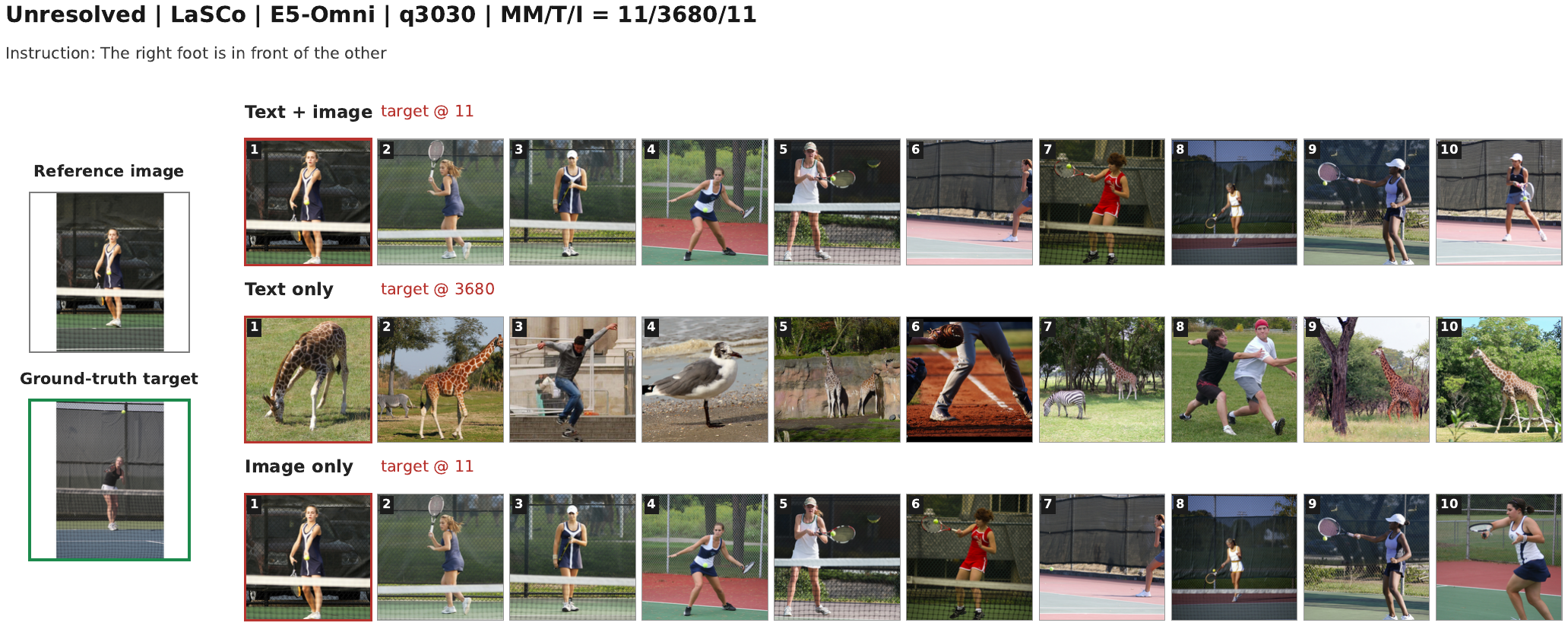}
\caption{Unresolved on LaSCo for E5-Omni (q3030). All three variants miss the target (MM/T/I = 11/3680/11), placing the query in the unresolved set pending human validation.}
\label{fig:qual-lasco-unresolved-e5omni}
\end{figure*}

\begin{figure*}[p]
\centering
\includegraphics[width=\textwidth]{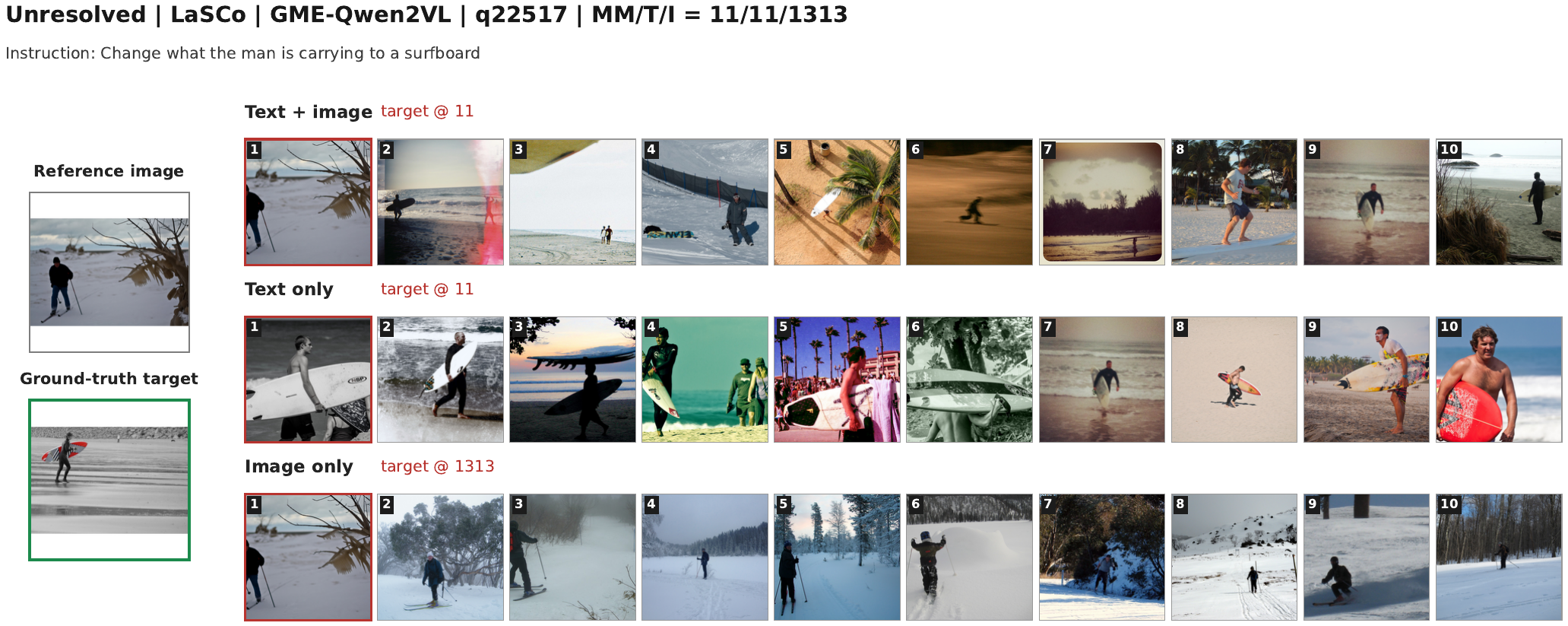}
\caption{Unresolved on LaSCo for GME-Qwen2VL (q22517). All three variants miss the target (MM/T/I = 11/11/1313), placing the query in the unresolved set pending human validation.}
\label{fig:qual-lasco-unresolved-gmeqwen2vl}
\end{figure*}

\end{document}